\documentclass[lettersize,journal]{IEEEtran}

\usepackage{amsfonts}
\usepackage{algorithmicx}
\usepackage{algorithm}
\usepackage{array}
\usepackage[caption=false,font=normalsize,labelfont=sf,textfont=sf]{subfig}
\usepackage{textcomp}
\PassOptionsToPackage{hyphens}{url}
\usepackage{hyperref}

\usepackage{stfloats}
\usepackage{url}
\usepackage{verbatim}
\usepackage{graphicx}
\usepackage{cite}
\usepackage{float}
\usepackage{epsfig}
\usepackage{amsmath}
\usepackage{amssymb}

\usepackage{bm}
\usepackage{booktabs}
\usepackage{longtable}
\usepackage{tabularray}
\usepackage{siunitx}
\usepackage[accsupp]{axessibility}  
\usepackage{xspace}
\usepackage[usenames,dvipsnames]{xcolor}
\usepackage{comment}
\usepackage{makecell}
\usepackage{pifont}
\usepackage{multirow}
\usepackage{flushend}
\usepackage{colortbl}
\usepackage{enumitem}
\usepackage{comment}
\usepackage[normalem]{ulem}

\usepackage{sidecap}

\usepackage{nicefrac}
\newcolumntype{R}[2]{%
    >{\adjustbox{angle=#1,lap=1.3\width-(#2)}\bgroup}%
    l%
    <{\egroup}%
}

\usepackage{pgfkeys}
\usepackage[]{pgfplots,pgfplotstable}
\usepackage{algpseudocode}
\usepackage{pgfplots}
\usepackage{threeparttable}
\usepackage[export]{adjustbox}

\usepackage{forest}
\useforestlibrary{edges}
\forestset{multiple directions/.style={for tree={#1}, phantom, for relative level=1{no edge, delay={!c.content/.pgfmath=content("!u")}, before computing xy={l=0,s=0}}},
    multiple directions/.default={},
    grow subtree/.style={for tree={grow=#1}}, 
    grow' subtree/.style={for tree={grow'=#1}}}

\usepackage{flexisym}

\usepackage[]{tikz}
\usetikzlibrary{arrows.meta,shapes,positioning,shadows,trees,math, fit, backgrounds}

\tikzset{
    my node/.style={
        draw=gray,
        inner color=gray!5,
        outer color=gray!10,
        thick,
        minimum width=1cm,
        text height=1.5ex,
        text depth=0ex,
        font=\sffamily,
    }
}

\definecolor{mh1}{RGB}{13,8,135}
\definecolor{mh2}{RGB}{84,2,163}
\definecolor{mh3}{RGB}{139,10,165}
\definecolor{mh4}{RGB}{185,50,137}
\definecolor{mh5}{RGB}{219,92,104}
\definecolor{mh6}{RGB}{244,136,73}
\definecolor{mh7}{RGB}{254,188,43}
\definecolor{mh8}{RGB}{240,249,33}

\usepackage{tkz-kiviat,numprint,fullpage} 

\usetikzlibrary{arrows}
\usepackage[nomessages]{fp}

\usetikzlibrary{arrows}

\pgfplotsset{
    x tick label style={/pgf/number format/.cd, set thousands separator={}}
}
\pgfplotsset{compat=1.11,
    /pgfplots/ybar legend/.style={
    /pgfplots/legend image code/.code={%
       \draw[##1,/tikz/.cd,yshift=-0.25em]
        (0cm,0cm) rectangle (3pt,0.8em);},
   },
}

\definecolor{myblue}{rgb}{0.19, 0.55, 0.91}
\definecolor{myred}{rgb}{0.82, 0.1, 0.26}
\definecolor{MyGreen}{RGB}{0, 180, 0}
\definecolor{MyRed}{RGB}{180, 0, 0}
\definecolor{MyYellow}{RGB}{180, 180, 0}
\definecolor{chpe}{RGB}{254,178,76}   
\definecolor{chmr}{RGB}{243,112,88}   
\newcommand{\cmark}{{\textcolor{MyGreen}{\ding{51}}}}%
\newcommand{\xmark}{{\textcolor{MyRed}{\ding{55}}}}%
\newcommand\na{\mbox{\scriptsize\color{gray} N/A}}

\makeatletter
\renewcommand\paragraph{\@startsection{paragraph}{4}{\z@}%
    {0.8ex \@plus 1ex \@minus .1ex}%
    {-1em}%
    {\normalfont \normalsize \bfseries}}
\makeatother

\def\addlegendimage{\csname pgfplots@addlegendimage\endcsname}

\newcommand\waymo{Waymo \xspace}

\newcommand\sloper{SLOPER4D\xspace}
\newcommand\hpe{3D Human Pose Estimation\xspace}
\newcommand\hmr{Human Mesh Recovery\xspace}

\newcommand\hpes{3D HPE\xspace}
\newcommand\hmrs{HMR\xspace}

\setlist[itemize]{leftmargin=*,topsep=2pt,itemsep=2pt}

\definecolor{shadeSuperT}{RGB}{251,233,231}
\colorlet{shadeSuper}{shadeSuperT!100!white} 
\definecolor{shadeSuperT2}{RGB}{255,243,224}
\colorlet{shadeSuper2}{shadeSuperT2!100!white} 
\definecolor{shadeWeakT}{RGB}{251,233,231}
\colorlet{shadeWeak}{shadeWeakT!70!white} 
\definecolor{shadeWeakT2}{RGB}{255,243,224}
\colorlet{shadeWeak2}{shadeWeakT2!70!white} 
\definecolor{shadeUnT}{RGB}{251,233,231}
\colorlet{shadeUn}{shadeUnT!40!white}

\begin{document}

\title{3D Human Pose and Shape Estimation from LiDAR Point Clouds: A Review}

\author{
    \IEEEauthorblockN{Salma Galaaoui\IEEEauthorrefmark{1}\IEEEauthorrefmark{2}, Eduardo Valle\IEEEauthorrefmark{1}, David Picard\IEEEauthorrefmark{2}, Nermin Samet\IEEEauthorrefmark{1}}
    \thanks{\IEEEauthorrefmark{1}Valeo.ai, Paris, France}
    \thanks{\IEEEauthorrefmark{2}LIGM, École Nationale des Ponts et Chaussées, IP Paris, Univ Gustave Eiffel, CNRS, France}
    \thanks{corresponding author: salma.galaaoui@valeo.com}
}

\maketitle

\begin{abstract}
In this paper, we present a comprehensive review of 3D human pose estimation and human mesh recovery from in-the-wild LiDAR point clouds.
We compare existing approaches across several key dimensions, and propose a structured taxonomy to classify these methods. Following this taxonomy, we analyze each method's strengths, limitations, and design choices. In addition, (i) we perform a quantitative comparison of the three most widely used datasets, detailing their characteristics; (ii) we compile unified definitions of all evaluation metrics; and (iii) we establish benchmark tables for both tasks on these datasets to enable fair comparisons and promote progress in the field. We also outline open challenges and research directions critical for advancing LiDAR-based 3D human understanding. Moreover, we maintain an accompanying webpage that organizes papers according to our taxonomy and continuously update it with new studies: \url{https://github.com/valeoai/3D-Human-Pose-Shape-Estimation-from-LiDAR}
\end{abstract}

\begin{IEEEkeywords}
3D human mesh recovery, 
3D human pose estimation, 
3D human pose estimation from LiDAR, 
3D human recovery from LiDAR,
human pose estimation benchmarks, 
human pose estimation metrics, 
human mesh recovery benchmarks, 
LiDAR point clouds, 
LiDAR human pose estimation datasets, 
point cloud-based human modeling, 
review, 
survey. 

\end{IEEEkeywords}

\section{Introduction}
\hpe and \hmr are crucial tasks in computer vision, with a critical role in understanding and modeling human motion and appearance. Indeed, the accurate 3D representation of human posture and shapes is fundamental for AI applications such as virtual and augmented reality~\cite{nguyen2005real}, human-computer interaction~\cite{rossol2015multisensor}, sports analytics~\cite{sharma2022pilot}, healthcare~\cite{chua2021telehealth}, safe autonomous driving~\cite{zanfirHUM3DILSemisupervisedMultimodal2022}, and behavior recognition in urban scenes~\cite{hu2004survey}. 

\hpe (\hpes) predicts, from inputs such as images, videos, depth maps, and point clouds, the 3D coordinates of key human body joints. Unlike its 2D counterpart, which only tracks the joint positions across the plane, 3D HPE provides a deeper understanding of spatial relationships and motion dynamics in the 3D space. By capturing depth and perspective information, 3D HPE overcomes the ambiguities caused by limb occlusions, making it essential for advancing AI-driven human-centric technologies.

\hmr (\hmrs), meanwhile, generates 3D representations of the human body beyond its skeletal structure, comprising detailed surface geometry, shape, and, sometimes, texture. 
\begin{figure}[t!]
\centering

\resizebox{1.0\linewidth}{!}{ %
\setlength{\tabcolsep}{8pt} 
\renewcommand{\arraystretch}{1.2} 
\begin{tabular}{>{\centering\arraybackslash}m{0.25\textwidth} 
                >{\centering\arraybackslash}m{0.25\textwidth} 
                >{\centering\arraybackslash}m{0.25\textwidth}}
    \includegraphics[width=0.27\textwidth]{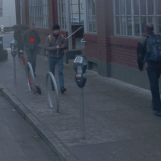} &
    \includegraphics[width=0.27\textwidth]{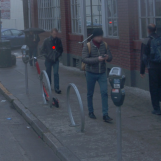} &
    \includegraphics[width=0.27\textwidth]{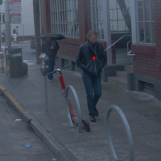} \\

    \includegraphics[width=0.22\textwidth]{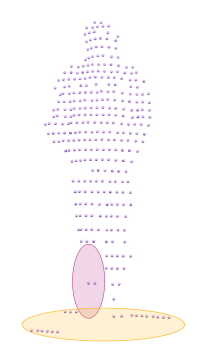} &
    \includegraphics[width=0.22\textwidth]{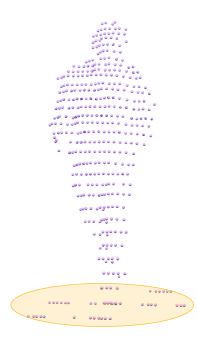} &
    \includegraphics[width=0.22\textwidth]{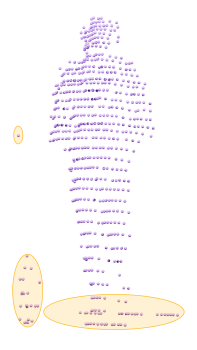} \\
    \Large{314 pts @ 16.7 m} & \Large{416 pts @ 13.9 m} & \Large{617 pts @ 10.8 m} \\[2ex]
\end{tabular}
}
\caption{\textbf{LiDAR human point cloud challenges} include noise, occlusions, and significant variations in point cloud density. The top rows show raw images and their corresponding human point clouds extracted from Waymo Open Dataset~\cite{sunScalabilityPerceptionAutonomous2020}. The subject of interest is marked with a red dot on the chest. The bottom row indicates the distance to the sensor (in meters) and the corresponding number of LiDAR points (pts). \textcolor{mh4}{\textbf{Occlusion}} of the right shin in the first frame; \textcolor{mh7}{\textbf{persistent noise}} near the feet; and point density is observed to increase as the subject moves closer to the sensor.}
\label{fig:lidar_hpc_challenges}
\end{figure}

In autonomous driving, point clouds captured by LiDAR sensors are crucial for detecting and analyzing human poses in dynamic environments, enabling accurate prediction of pedestrian behavior and enhancing safety. Extracting detailed skeletons and shapes from sparse LiDAR point clouds is an emerging frontier of \hpes and \hmrs that trades the density of images and videos for the geometric precision of point clouds. Historically, due to the lack of adequately annotated LiDAR datasets, RGB-D cameras were used more often to provide depth information for pose estimation tasks~\cite{ganapathi2012real,  girshick2011efficient, v2v, shotton2011real}. However, recent datasets such as \waymo~\cite{sunScalabilityPerceptionAutonomous2020} and \sloper~\cite{daiSLOPER4DSceneAwareDataset2023} provide point cloud data with annotated 2D and 3D human poses, enabling researchers to train and validate their models and raising the interest in LiDAR-based \hpes and \hmrs. 
Nevertheless, data diversity is still a challenge, which researchers address by leveraging additional modalities, such as images and depth maps, and exploiting unlabeled data, with unsupervised and weakly supervised training.

LiDAR point clouds offer unique advantages, such as intrinsic 3D geometric information, accurate distance measurement, robustness to lighting conditions, and inherent privacy preservation. They also pose important challenges, as shown in Figure~\ref{fig:lidar_hpc_challenges}: point clouds are sparse and irregularly sampled, often resulting in incomplete scans of the human body; scan patterns depend on the LiDAR technology employed, sensor-dependent noise may be significant; and point clouds are sensitive to occlusions. Extracting accurate human poses or reconstructing detailed meshes requires addressing those challenges.

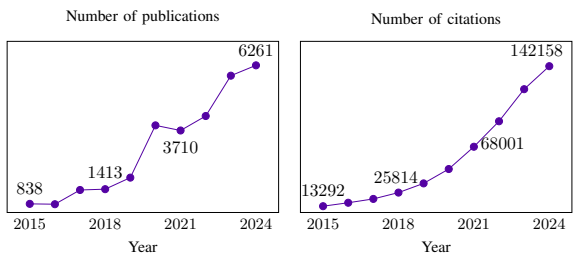
\begin{figure}[ht]
    \resizebox{\columnwidth}{!}{
\begin{tabular}{cc}
    \centering

    \begin{tikzpicture}
        \begin{axis}[
            width=7cm,
            height=5cm,
            xlabel={Year},
            xtick={2015, 2018,  2021,  2024},
            xtick style={draw=none},
            ymin=500, ymax=7200,
            grid=none,
            ymajorticks=false,
            tick label style={font=\small},
            label style={font=\small},
            title style={font=\small},
            title={Number of publications},
            yticklabels={},
        ]
            \addplot[color=mh2, mark=*] coordinates {
(2015,838)  (2016,824)  (2017,1381)  (2018,1413)  (2019,1865)  
(2020,3911)  (2021,3710)  (2022,4277)  (2023,5856)  (2024,6261)  
            };
            \node [above] at (axis cs:  2015,1000) {$838$};
            \node [above] at (axis cs:  2018,1600) {$1413$};
            \node [above] at (axis cs:  2021,2600) {$3710$};
            \node [above] at (axis cs:  2024,6400) {$6261$};
        \end{axis}
    \end{tikzpicture} &
\begin{tikzpicture}
        \begin{axis}[
            width=7cm,
            height=5cm,
            xlabel={Year},
            xtick={2015, 2018,  2021,  2024},
            xtick style={draw=none},
            ymin=7500, ymax=165000,
            grid=none,
            ymajorticks=false,
            tick label style={font=\small},
            label style={font=\small},
            title style={font=\small},
            title={Number of citations},
            yticklabels={},
        ]
            \addplot[color=mh2, mark=*] coordinates {
(2015,13292)  (2016,16464)  (2017,19989)  (2018,25814)  (2019,34279)  
(2020,47492)  (2021,68001)  (2022,91450)  (2023,120983)  (2024,142158)  
            };
            \node [above] at (axis cs:  2015,16900) {$13292$};
            \node [above] at (axis cs:  2017.9,29000) {$25814$};
            \node [above] at (axis cs:  2022.15,60000) {$68001$};
            \node [above] at (axis cs:  2023.5,145900) {$142158$}; 
        \end{axis}
    \end{tikzpicture}
    \vspace{-10pt}
    \\

\end{tabular}

}

														

    \caption{\textbf{The growing interest in 3D human pose estimation from LiDAR point clouds.} (a) The number of publications per year. (b) The yearly count of citations. This data was retrieved from www.dimensions.ai using the query: \textit{3D human pose estimation from LiDAR point clouds}.}   
    \label{fig:interest}
\end{figure}

\subsection{Scope, Aim, and Reading Guide}

Recently, the release of annotated LiDAR point cloud datasets for 3D HPE and HMR has fueled growing research and application interest in performing these tasks directly from LiDAR data (Figure~\ref{fig:interest}). 

In light of this trend, we review deep learning models for \hpes and \hmrs from outdoor LiDAR point clouds, with a particular focus on autonomous driving and human behavior analysis in urban environments. 
Our main goal in this paper is to provide a comprehensive analysis and discussion of existing methods in this domain. To achieve this,

\begin{enumerate}
    \item We categorize sensors used in \hpes/\hmrs tasks, proposing a taxonomy that highlights the strengths and limitations of cameras, LiDARs, and radars. Given our survey’s focus on LiDAR, we detail its types and underlying technologies (Section \ref{sec:background}). 

    \item We present a comparative summary of existing methods, analyze their network architectures, and propose a taxonomy for their systematic categorization (Section~\ref{sec:overview}).

    \item We conduct an extensive review of 32 studies published between 2019 and 2025. This review offers detailed analyses of methods, highlighting their design choices, strengths, and limitations on 3D HPE and HMR from LiDAR point clouds (Section~\ref{sec:hpe} and Section~\ref{sec:hmr}).

    \item We describe the three most widely used datasets, outline their key statistics and characteristics, and briefly summarize the remaining public datasets for completeness (Section~\ref{sec:datasets}). 

    \item We compile and formalize definitions of all metrics used to evaluate 3D HPE and HMR methods from LiDAR point clouds (Section~\ref{sec:benchmark}).
    
    \item We establish benchmarks on the three most widely used datasets, providing a comprehensive assessment of the current state-of-the-art (Section~\ref{sec:benchmark}).

    \item We discuss the key challenges and unresolved issues in \hpes and \hmrs from LiDAR point clouds, offering insights and directions for future research (Section \ref{sec:future}).

    \item Finally, we provide an accompanying repository \footnote{\fontsize{6pt}{8pt}\selectfont\url{https://github.com/valeoai/3D-Human-Pose-Shape-Estimation-from-LiDAR}} of studies on \hpes and \hmrs from LiDAR point clouds. This resource will be continuously updated to include new research developments.
\end{enumerate}

Each section in this survey has been designed to be self-contained for ease of reading.
For readers already familiar with LiDAR point clouds and 3D HPE/HMR tasks, we recommend proceeding directly to Section~\ref{sec:overview} and using Figure~\ref{fig:taxonomy} as a guide to navigate the methods categorized by our taxonomy. Readers less familiar with these topics are encouraged to begin with Section~\ref{sec:background}; For those seeking more comprehensive background information, we refer to the in-depth reviews cited in Section~\ref{sec:background}.

\subsection{Previous surveys}
\label{subsec:pre_surveys}
Recent surveys on \hpes \cite{zheng2023deepsurvey, chen2020monocularsurvey, neupane2024survey, knap2024survey, nogueira2024markerlesssurvey, azam2024survey, zhou2023humansurvey, peng2024emergingsurvey, liu2024deepsurvey} primarily focus on deep learning applied to images or videos, in monocular or multi-view setups. The reader will find taxonomies for \hpes and comprehensive analyses of key aspects such as human representations\cite{knap2024survey}, learning strategies~\cite{neupane2024survey}, pose data augmentation~\cite{neupane2024survey, lan2022vision}, role of \hpes within tracking and action recognition tasks~\cite{zhou2023humansurvey}, and the adoption of tools such as vision transformers, large language models, and multi-modality fusion~\cite{peng2024emergingsurvey}. Those references also offer detailed discussions of popular datasets, evaluation metrics, and applications within the domain.

\hmrs may cursorily appear \hpes-focused surveys~\cite{zheng2023deepsurvey, knap2024survey}, but the reader will find more detail in dedicated surveys: Tian et al.\cite{tian2023recoveringsurvey}, a thorough review with a focus on explicit models, while Chen et al.\cite{chen2021towardsoverview} focuses on implicit rendering techniques. Liu et al.~\cite{liu2024deepsurvey} reviews in detail both parametric and non-parametric \hmrs. 

Machine-learning methods prior to the deep learning era are reviewed in~\cite{liu2015survey, gong2016humansurvey, sarafianos20163dsurvey}.

LiDAR sensors are notably absent in existing surveys. A few briefly and broadly address sensors beyond cameras. \hpes with depth sensors, Inertial Measurement Units (IMUs), and radio frequency devices are summarized in~\cite{zheng2023deepsurvey} and \cite{lan2022vision}, while ~\cite{nogueira2024markerlesssurvey} outlines alternative cameras such as RGB-D and ToF and wireless sensors such as RF and Wi-Fi. 

In contrast, our work deep-dives on \hpes and \hmrs from LiDAR point clouds, targeting real-world applications for outdoor scenes, using "in-the-wild" datasets.

\section{Background}
\label{sec:background}

\subsection{3D Human Pose and Shape Estimation from Images and Videos}

\textbf{\hpe} for images and videos was the target of intense research and remains at the forefront of the field, while also influencing \hpes for LiDAR point cloud data.

Monocular \hpe approaches are broadly categorized in two groups: \textit{deterministic} and \textit{generative} methods. Early deterministic approaches directly predicted 3D keypoints from images in an end-to-end manner~\cite{pavlakos2017coarse, moreno20173d, mehta2017vnect, sun2017compositional}. Subsequently, two-stage approaches gained popularity, where 2D keypoints were first detected and then "lifted" into the 3D space using specialized neural networks~\cite{sb}.

To leverage temporal information, video-based methods were explored utilizing temporal dependencies between consecutive frames. Most video-based approaches~\cite{cai2019exploiting, chen2021anatomy, foo2023unified, hu2021conditional, videopose, shan2022p, xu2020deep, poseformer} adopt a two-stage pipeline: 2D poses are extracted from video frames using 2D pose detectors, and spatio-temporal information from these pose sequences is then utilized for 3D estimation. Methods for encoding spatio-temporal information range from Convolutional Neural Networks (CNNs)~\cite{videopose} to Graph Convolutional Networks (GCNs)\cite{zhao2019semantic, liu2020comprehensive, cai2019exploiting, zou2021modulated, xu2021graph}. More recently, attention-based spatio-temporal transformer architectures have gained attention~\cite{mixste, poseformer, shan2022p}.

A core challenge in lifting 2D poses to 3D space lies in the inherent ambiguity, as multiple 3D poses can correspond to the same 2D input. Deterministic approaches often struggle to address this ambiguity effectively. To overcome this limitation, researchers have focused on generative methods with multiple hypotheses. Approaches include variational autoencoders (VAEs)\cite{sharma2019monocular}, normalizing flows\cite{kolotouros2021probabilistic}, and, very recently, diffusion models~\cite{diffpose, ddp, diffhpe, h3wb, manipose}.

State-of-the-art methods now integrate both the temporal and generative paradigms~\cite{ddp, diffpose, diffhpe, manipose, pafuse}. For instance, D3DP\cite{ddp} utilizes the MixSTE backbone as a denoiser and conditions on raw 2D keypoints. DiffPose\cite{diffpose} employs a diffusion approach based on Gaussian mixture models trained on 2D heatmaps. 

Despite these advancements, 3D HPE from images and videos inherits limitations associated with passive sensors. For example, absolute scale estimation remains challenging and precise distance measurements are not feasible without multiple camera setups.

\textbf{3D Human Mesh Recovery} aims to reconstruct detailed 3D body manifolds, typically either adopting a parametric approach that predicts latent parameters to be mapped to a mesh by a pretrained model or a non-parametric approach that directly infers the dense mesh from data.

Parametric models, particularly SMPL (Skinned Multi-Person Linear) \cite{loper2023smpl}, have become the foundation of human mesh recovery. SMPL represents the human body as a learnable skinned mesh model, where pose parameters ($\theta$) define joint angles and global posture, and shape parameters ($\beta$) capture body morphology. Building upon this framework, several extensions have been proposed to enhance its representational power. MANO \cite{mano} focuses on detailed hand modeling, FLAME  \cite{flame} targets facial representation, and SMPL-X \cite{pavlakos2019expressive} unifies body, face, and hands into a single model. To enable accurate whole-body mesh recovery from 2D observations, SMPLify-X \cite{pavlakos2019expressive} iteratively fits SMPL-X to body, face, and hand keypoints.

A range of methods has focused on SMPL-based mesh recovery, exploring approaches such as direct parameter regression \cite{tan2017indirect, tung2017self, pavlakos2018learning, omran2018neural}, mesh vertex prediction with Graph-CNNs \cite{kolotouros2019convolutional}, and self-supervised contrastive learning frameworks \cite{xu20203d}. Temporal consistency has been addressed with recurrent models like TCMR \cite{choi2021beyond}, while probabilistic formulations and lightweight transformer-based designs further enhance accuracy and efficiency \cite{zheng2022lightweight, choi2020pose2mesh, bregier2025condimen, baradel2024multi}.

Beyond parametric models, non-parametric approaches have emerged for flexibility. Voxel-based methods such as BodyNet \cite{varol2018bodynet} predict volumetric human shape from single images. Implicit surface representations, like TetraTSDF \cite{onizuka2020tetratsdf}, address reconstruction under loose clothing, while PaMIR \cite{zheng2021pamir} leverages image features and SMPL priors for high-quality reconstructions.

For those interested in a deeper dive into 3D HPE and HMR, we recommend referring to the surveys \cite{zheng2023deepsurvey, chen2020monocularsurvey, neupane2024survey, knap2024survey, nogueira2024markerlesssurvey, azam2024survey, zhou2023humansurvey, peng2024emergingsurvey, liu2024deepsurvey}.
\subsection{Sensors and Data Acquisition}

Embodied AI systems require a detailed understanding of the environment and all its actors, in our case, humans, in order to interact effectively and safely. These systems depend on input signals to support decision-making and planning. Often, exteroception refers to the perception of the external environment through appropriate sensors, enabling a comprehensive and accurate understanding of the surroundings. This perceptual ability serves as a cornerstone of embodied AI.

A conventional camera sensor is often insufficient to extract the scene's spatial information, since its projection on 2D loses the 3D depth information. While powerful deep learning algorithms empower cameras for affordably detecting objects' position and pose, 2D sensors' intrinsic lack of depth measurements makes them an easy target for physical adversarial attacks \cite{guesmiNavigatingThreatsSurvey2024}. Technologies such as depth cameras, RADAR, and LiDAR allow bypassing those difficulties.

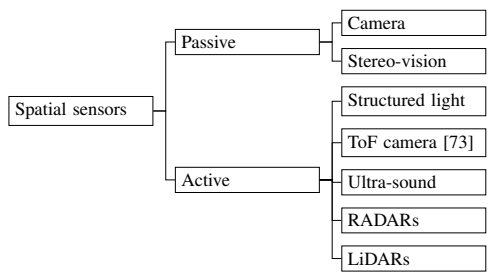
\begin{figure}[ht]
    \centering

    \resizebox{0.8\linewidth}{!}{\begin{forest}
  for tree={
    child anchor=west,
    parent anchor=east,
    grow'=east,
    text width=2.5cm,%
    font=\rmfamily,
    draw,
    anchor=west,
    edge path={
      \noexpand\path[\forestoption{edge}]
        (.child anchor) -| +(-5pt,0) -- +(-5pt,0) |-
        (!u.parent anchor)\forestoption{edge label};
    },
  }
    [Spatial sensors
    [Passive [Camera, tier=actpasstypes][Stereo-vision, tier=actpasstypes]]
    [Active[Structured light, tier=actpasstypes][ToF camera \cite{hansardTimeofFlightCamerasPrinciples2013}, tier=actpasstypes][Ultra-sound, tier=actpasstypes]
    [RADARs, tier=actpasstypes]
    [LiDARs, tier=actpasstypes]
    ]
    ]
\end{forest}}
    
    \caption{\textbf{Taxonomy of spatial sensors.} ToF stands for time-of-flight.} 
    \label{fig:spatial_sensor_taxonomy}
\end{figure}

Sensors may be active or passive (Figure \ref{fig:spatial_sensor_taxonomy}). We briefly introduce most of them, with a focus on those most relevant to outdoor applications, such as autonomous driving.

\begin{figure}[ht]

\tikzset{
basic/.style  = {draw, text width=3cm, font=\large\rmfamily, rectangle},
root/.style   = {basic, thin, align=center, text width=3cm},
cat1/.style = {basic, thin, align=center, text width=3cm,},
cat2/.style = {basic, thin, align=center, text width=3cm,},
cat3/.style = {basic, thin, align=center, text width=4cm},
edge from parent/.style={draw=black, edge from parent fork right}
}

\centering
\resizebox{\linewidth}{!}{
\begin{forest}
  for tree={
    child anchor=west,
    parent anchor=east,
    grow'=east,
    text width=2.3cm,%
    font=\rmfamily,
    draw,
    anchor=west,
    edge path={
      \noexpand\path[\forestoption{edge}]
        (.child anchor) -| +(-5pt,0) -- +(-5pt,0) |-
        (!u.parent anchor)\forestoption{edge label};
    },
  }
[LiDAR \\ Categories
    [Scanning Pattern
        [Non-Repetitive Scanning,cat3, tier=last]
        [Rotating Multi-Beam Scanning,cat3, tier=last]
        [No Scanning (Flash LiDAR),cat3, tier=last]
    ,cat1]
    [Beam Steering Technology
        [Mechanical
            [{ Rotating/Oscillating Mirror/Polygon},cat3, tier=last]
            [{ Rotating LiDAR Head},cat3, tier=last]
          ,cat2]
        [Quasi-Solid-State
            [Micro-Electro-Mechanical Systems (MEMS),cat3, tier=last] 
            [Wedge (Risley) Prism,cat3, tier=last]
          ,cat2]
        [Solid-State
            [Optical Phased Array,cat3, tier=last] 
          ,cat2]
    ,cat1]
    [Sensing Technology
        [Pulse (ToF),cat3, tier=last]
        [Amplitude-Modulated Continuous Wave,cat3, tier=last] 
        [Frequency-Modulated Continuous Wave,cat3, tier=last] 
    ,cat1]
]
\end{forest}
}
    \caption{\textbf{Taxonomy of LiDAR sensors.} ToF stands for time-of-flight.}
    \label{fig:lidar_sensor_taxonomy}
\end{figure}
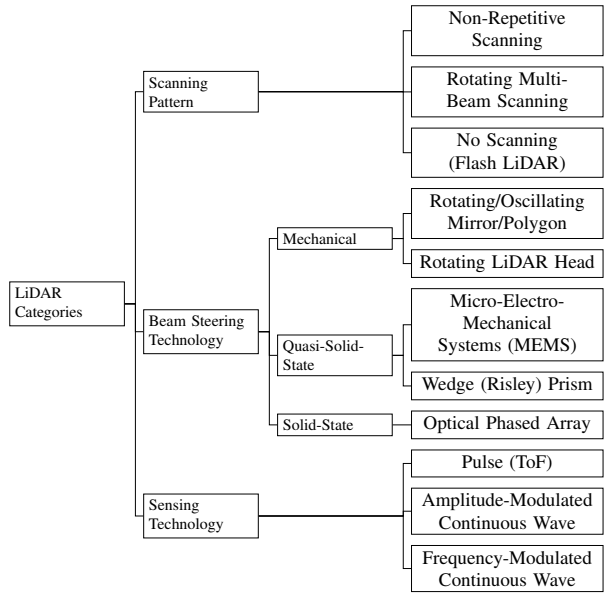

\paragraph{\textbf{Depth cameras}}

Depth cameras come in three different technologies: passive stereo-vision, active structured light, and active time-of-flight. Stereo-vision relies on two well-calibrated (for intrinsics and extrinsics) camera sensors, using the apparent displacement of visual features between the two images to compute depth~\cite{martiReviewSensorTechnologies2019}. Structured light combines a monocular camera with an infrared source that projects a known pattern onto the scene, whose deformation, once reflected by objects, allows inferring depth. Finally, time-of-flight computes depth by measuring the round-trip time of modulated infrared light projected onto the scene.

\paragraph{\textbf{Radar Sensors}}
RADAR determines object distance by emitting high-frequency electromagnetic waves and analyzing their reflections, while the Doppler effect provides measurements of relative velocity. 
RADAR is widely employed in military systems and is extensively integrated into Autonomous Vehicles (AV) and Advanced Driver Assistance Systems (ADAS), supporting functions such as parking assistance and adaptive cruise control.

\paragraph{\textbf{LiDAR Sensors}}
LiDAR (Light Detection And Ranging) uses laser pulses to measure distances to objects, relying on the time-of-flight principle. Categorizing LiDARs is challenging, but three (non-orthogonal) axes of analysis are prominent: scanning pattern, beam steering technology, and sensing technology (Figure \ref{fig:lidar_sensor_taxonomy}). Sensor surveys \cite{liLidarAutonomousDriving2020}-II.B, \cite{martiReviewSensorTechnologies2019}, and \cite{liProgressReviewSolidState2022} provide a detailed categorization for solid-state solutions, considering laser sources, wavelengths, photo-detectors, and scanning systems. We follow much of their terminology.

\begin{figure}[t]
    \centering
    \includegraphics[width=1.0\linewidth]{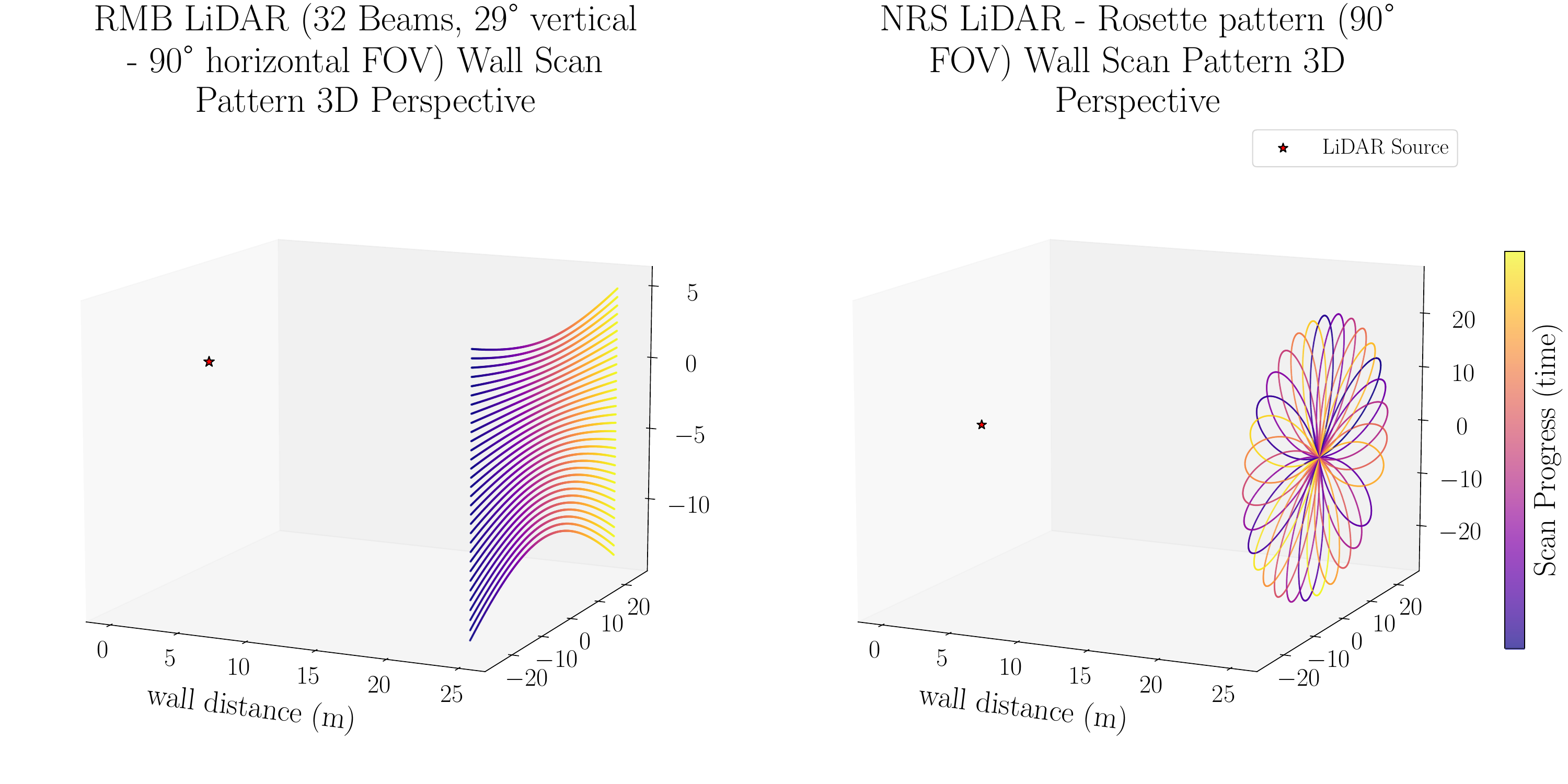}
    \caption{\textbf{Comparison of scanning patterns:} RMB rotating pattern and NRS rosette pattern, simulated at 25 m from an infinite wall.} 
    \label{fig:rmb_vs_nrs_pattern}
\end{figure}

\textit{Scanning pattern} is the most important aspect for us. NRS (Non-Repetitive Scanning) progressively covers the field of view by casting a non-repetitive pattern. RMB (Rotating Multi-Beam) scans the field of view (FoV) in a uniform direction. In contrast to these two types of scanning patterns, flash LiDAR is a type of sensor that involves no scanning, it makes use of an optical diffuser to illuminate the scene and a photodetector array capturing the entire target scene within a single shot. Two examples of the most common NRS and RMB patterns are illustrated in Figure~\ref{fig:rmb_vs_nrs_pattern}.

\textit{Beam steering} technologies represent how scanning is achieved, \textit{i.e.,} methods to control the path of the laser beam as it scans the field of view. 
Mechanical steering relies on an oscillating mirror, a prism, or, as in rotating LiDARs, the whole sensor. 
Quasi-solid-state steering relies on minimally moving mirrors or prisms controlled by tiny (electro-)mechanical devices. Finally, Optical Phased Arrays move the beam with solid-state phase modulators.

\textit{Sensing technology} refers to the acquisition of distance information, with three main sub-categories. 
Pulse LiDARs and Amplitude-Modulated Continuous Wave both rely on time-of-flight 
\cite{hansardTimeofFlightCamerasPrinciples2013}. 
The former uses the turnaround delay of an optical pulse reflected by the scene, while the latter uses the phase shift between the emitted and received amplitude-modulated signal. 
Finally, Frequency-Modulated Continuous Wave LiDARs frequency-modulate the emitted signal and derive both distance and velocity from the target return.

LiDAR has emerged as a cornerstone technology, enabling applications from outdoor environments to indoor spaces, where accurate perception is important for the deployment of smart autonomous systems. 

\textit{Outdoor and large space LiDAR applications} mainly use mid- and long-range sensors for AV/ADAS, traffic management, surveillance, and even tasks like gardening.
\textsl{RoboSense} \cite{robosenseRoboSenseRoboSense} provides mid- (30–75 m) and long-range (up to 200 m) LiDARs using Micro Electro Mechanical Systems (MEMS), Flash, and mechanical spinning technologies for AV/ADAS, security, and robotics. Its sensors are deployed in Geely and Toyota vehicles \cite{robosenseRoboSenseRanked} and support Outsight’s Smart City solutions \cite{outsightShiftPerception, outsightOutsightRobosense}.
\textsl{Luminar} \cite{luminartechLuminarsTechnologies} provides long-range MEMS LiDARs, including the IRIS (up to 300 m) and the Sentinel platform, used in AV/ADAS systems by Toyota \cite{thevergeFederalGovernment}, Volvo \cite{volvoES90pressrelease}, and Mercedes \cite{luminartechLuminarMercedesBenz}.
\textsl{Continental} \cite{continentalHomeContinental} offers mid-range (50 m) Flash (HFL110 \cite{continentalaftermarketContinentalAftermarket}) and long-range MEMS (HRL131 \cite{continentalautomotiveContinentalAutomotive}) LiDARs for AV/ADAS. The company is also testing its sensors in airport operations, with plans to equip passenger buses, fuel trucks, luggage vehicles, and even aircraft for enhanced adaptive cruise control and collision avoidance \cite{continentalInnovationsFrom}.
\textsl{Waymo} \cite{waymoWaymoSelfDriving} previously partnered with Velodyne (now Ouster) for spinning long-range LiDAR, but now develops its own mid- and long-range sensors for its new vehicle generation.
\textsl{Ouster} \cite{ousterDigitalLidar} offers the mechanically spinning OS1 (mid-range) \cite{ousterOS1HighRes} and OS2 (long-range) \cite{ousterOusterOS2}, using a rotating multi-beam pattern for security, traffic management, and AV/ADAS applications. 
\textsl{Livox} offers the MID Series \cite{livoxtechMid40Lidar} LiDARs, with the MID-40 reaching 260 m range and employing a proprietary non-repetitive Rosette-like scanning pattern via rotating Risley prisms.

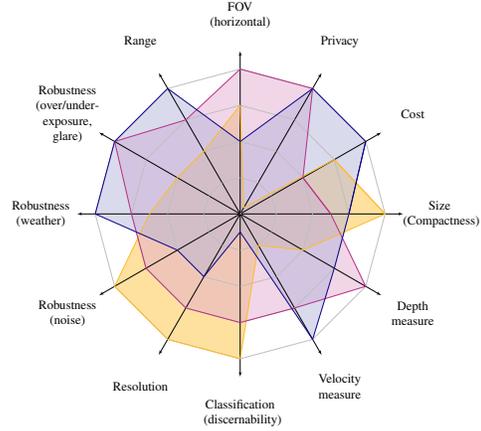
\begin{figure}[t]
    \centering
    \resizebox{0.8\linewidth}{!}{\begin{tikzpicture}

\tkzKiviatDiagramFromFile[
        scale=1,
        label distance=.5cm,
        gap     = 1,label space=1.5,  
        lattice = 4]{Figures/tableae.dat}
\tkzKiviatLineFromFile[thick,
                       color      = mh7,
                       mark size  = 4pt,
                       fill       = mh7!90]{Figures/tableae.dat}{1}
\tkzKiviatLineFromFile[thick,
                       color      = mh4,
                       mark size  = 4pt,
                       fill       = mh4!40]{Figures/tableae.dat}{2} 
\tkzKiviatLineFromFile[thick,
                       color      = mh1,
                       mark size  = 4pt,
                       fill       = mh1!30]{Figures/tableae.dat}{3}
\end{tikzpicture}}
    \caption{ \textbf{Comparison of \textcolor{mh1}{Radar}, \textcolor{mh4}{LiDAR}, and \textcolor{mh7}{Camera} across different criteria}, based on the surveys \cite{vargasOverviewAutonomousVehicles2021,liCompositionApplicationCurrent2021,martiReviewSensorTechnologies2019}. A higher rank along an axis indicates a more advantageous performance of a sensor with respect to that criterion. Since specifications may vary across application domains, we limit the comparison to the Autonomous Driving domain to ensure fairness. }
    \label{fig:sensors_spider_criteria}
\end{figure}

\textit{Indoor and confined space LiDAR applications} include floor-cleaning robots, indoor delivery robots and other service systems operating across floors.
\textsl{Ouster} offers the 35 m OS0 \cite{ousterOusterOS0}, suited for applications like lawnmowing robots and warehouse management.
The Low-Cost Robot \textsl{LoCoBot} is 
equipped with a 12 m RPLIDAR A2 \cite{slamtecRPLIDARA2Solid}, a single-beam 2D LiDAR used for localization and mapping.
\textsl{Intel RealSense L515} \cite{realsenseaiRealSenseLiDAR} is a compact 9 m MEMS LiDAR widely used in indoor robotics.
\textsl{Unitree} employs an in-house NRS pattern in its L1 (20 m) \cite{unitreeLidarApplication} and L2 (30 m) \cite{unitreeUnitreeLiDAR} LiDARs, which use rotating mirrors similar to MEMS.

\begin{table*}[t]
\centering
\tabcolsep 4pt

\newcommand{\M}{\textcolor{cyan}{\textbf{M}}}
\newcommand{\D}{\textcolor{lime}{\textbf{D}}}
\newcommand{\depth}{\textcolor{pink}{\textbf{D}}}
\newcommand{\RI}{\textcolor{violet}{\textbf{RI}}}
\newcommand{\B}{\textcolor{orange}{\textbf{B}}}
\renewcommand{\V}{\textcolor{teal}{\textbf{V}}}
\newcommand{\IBI}{\textcolor{pink}{\textbf{IbI}}}
\newcommand{\WS}{\textcolor{purple}{\textbf{WS}}}

\caption{Comparative Summary of 3D HPE and HMR methods.}

\rowcolors{3}{gray!10}{} 

\resizebox{1.0\textwidth}{!}{
\begin{tabular}{lcccccccccc}

\toprule

\multirow{2}{*}{Paper}
& \multirow{2}{*}{Modality}
& Aux.
& \multirow{2}{*}{Temporal}
& \multirow{2}{*}{Supervision}
& \multirow{2}{*}{Task}
& \multirow{2}{*}{Data}
& Paper
& \multirow{2}{*}{LiDAR} 
& Real-
& \multirow{2}{*}{Sensor type} 
\\

&  & Modality & & & & & type & & time &    \\
 \midrule

\rowcolor{white} 
\textbf{\textit{HPE Methods}} &  & &  &  &  &  & &   &  &  \\

\rowcolor{gray!10}
DAPT~\cite{anPretrainingDensityAwarePose2025} &  LiDAR &  \V & \xmark & supervised & multi & Real+Synth & \M & single  & ? & RMB + NRS \\

\rowcolor{white} 
FusionPose~\cite{congWeaklySupervised3D2023} &  CAM+LiDAR &   & \cmark & weakly-supervised & single & Real & \M+\D & single  & ? & RMB  \\

\rowcolor{gray!10}
GC-KPL~\cite{weng3DHumanKeypoints2023} & LiDAR &   & \cmark & unsupervised  & multi & Real+Synth & \M & single  & ? & RMB  \\

\rowcolor{white} 
HPERL~\cite{furstHPERL3DHuman2021} &  CAM+LiDAR & \B & \xmark & weakly-supervised & single & Real & \M & single & ? & RMB \\

\rowcolor{gray!10}
HUM3DIL~\cite{zanfirHUM3DILSemisupervisedMultimodal2022} &  CAM+LiDAR & \depth & \xmark & weakly-supervised & single & Real & \M & single & ? & RMB  \\

\rowcolor{white} 
LiCamPose~\cite{panLiCamPoseCombiningMultiView2025} &  CAM+LiDAR & \V & \xmark & weakly-supervised & single & Real+Synth & \M & multi  & ? & NRS \\

\rowcolor{gray!10}
LiDAR-HMP~\cite{hanPracticalHumanMotion2024} &  LiDAR &   & \cmark & supervised & multi & Real & \M & single  & \cmark & RMB  \\

\rowcolor{white} 
LidPose~\cite{kovacsLidPoseRealTime3D2024} &  LiDAR  & \RI &  \xmark & supervised &  single & Real & \M+\D & single &  \cmark & NRS  \\

\rowcolor{gray!10}
LPFormer~\cite{yeLPFormerLiDARPose2024} &  LiDAR & \B+\V & \xmark & supervised & multi & Real & \M & single  & ? & RMB \\

\rowcolor{white} 
MMVP~\cite{fanHumanM3MultiviewMultimodal2023}   & CAM+LiDAR & \V & \xmark & supervised & single & Real & \M+\D & multi &  ? & NRS  \\

\rowcolor{gray!10}
UniPVU-Human~\cite{xuUnifiedFrameworkHumancentric2024} &  LiDAR &   & \cmark & supervised & multi & Real+Synth & \M & single  & ? & RMB  \\

\rowcolor{white} 
VoxelKP~\cite{shiVoxelKPVoxelbasedNetwork2023} &  LiDAR & \B+\V & \xmark & supervised & single & Real & \M & single  & \xmark & RMB \\

\rowcolor{gray!10}
WS-HPE~\cite{zhengMultimodal3DHuman2022} &  CAM+LiDAR  &   & \xmark & weakly-supervised & multi & Real & \M & single  & ? & RMB  \\

\rowcolor{white} 
WS-Fusion~\cite{bauerWeaklySupervisedMultiModal2023} &  CAM+LiDAR  &   & \xmark & weakly-supervised & single & Real & \M & single & \cmark & RMB \\

\midrule

\rowcolor{white}
\textbf{\textit{HPE \& HMR Methods}} &  &  &  &  &  &  & &   &  &  \\

\rowcolor{gray!10} 
LiDARCap~\cite{liLiDARCapLongrangeMarkerless2022} &  LiDAR &   & \cmark & supervised & multi & Real & \M+\D & single  & ? & RMB \\

\rowcolor{white}
LiDARCapV2~\cite{zhangLiDARCapV23DHuman2024} &  LiDAR &   & \cmark & supervised & multi & Real & \M+\D & single  & ? & RMB  \\

\rowcolor{gray!10} 
LiDAR-HMR~\cite{fanLiDARHMR3DHuman2025} &  LiDAR &   & \xmark & supervised & multi & Real & \M & single &  ? & RMB + NRS \\

\rowcolor{white}
LIP~\cite{renLiDARaidInertialPoser2023} &  IMU+LiDAR &   & \cmark & supervised & multi & Real & \M+\D & single  & ? & RMB  \\

\midrule

\rowcolor{white} 
\textbf{\textit{HMR Methods}} &  & &  &  &  &  & &   &  &  \\

\rowcolor{gray!10}
CIMI4D~\cite{yanCIMI4DLargeMultimodal2023}* &  CAM+IMU+LiDAR &   & \cmark & weakly-supervised & single & Real & \M+\D & single  & ? & RMB  \\

\rowcolor{white}
FreeCap~\cite{xueFreeCapHybridCalibrationFree2025} & CAM+LiDAR &   & \cmark & supervised & single & Real & \M & single & \xmark  & RMB + NRS \\

\rowcolor{gray!10}
HSC4D~\cite{daiHSC4DHumancentered4D2022}*  &  IMU+LiDAR &   & \cmark & weakly-supervised & single & Real & \M+\D & single &  ? & RMB  \\

\rowcolor{white} 
Human-M3~\cite{fanHumanM3MultiviewMultimodal2023}* & CAM+LiDAR &   & \cmark & weakly-supervised & single & Real & \M+\D & multi & ? & NRS  \\

\rowcolor{gray!10}
LiveHPS~\cite{renLiveHPSLiDARbasedScenelevel2024} &  LiDAR &   & \cmark & supervised & multi & Real & \M+\D & single &  \cmark & RMB  \\

\rowcolor{white} 
LiveHPS++~\cite{renLiveHPSRobustCoherent2024} &  LiDAR &   & \cmark & supervised & multi & Real+Synth & \M & single & \cmark & RMB\\

\rowcolor{gray!10}
NE-3D-HPE~\cite{zhangNeighborhoodEnhanced3DHuman2024} &  LiDAR &   & \cmark & supervised & multi & Real & \M & single &  ? & RMB \\

\rowcolor{white} 
PEAR-Proj~\cite{linHmPEARDatasetHuman2024} &  CAM+LiDAR &   & \cmark & supervised & multi & Real & \M+\D & single &  ? & RMB  \\

\rowcolor{gray!10}
ReMP~\cite{jangReMPReusableMotion2025} &  IMU/LiDAR &   & \cmark & weakly-supervised & multi & Real+Synth & \M & single  & ? & RMB  \\

\rowcolor{white} 
SLOPER4D~\cite{daiSLOPER4DSceneAwareDataset2023}* &  CAM+IMU+LiDAR &   & \cmark & weakly-supervised & single & Real & \M+\D & single  & ? & RMB  \\

\rowcolor{gray!10}
SMPLify-3D~\cite{dumontImprovingImagebased3D} &  CAM+LiDAR &   & \xmark & supervised & single & Real+Synth & \M & single &  ? & RMB + NRS  \\

\bottomrule
\end{tabular}}
\label{tab:rw}
\begin{tablenotes} 
    \item 
    \footnotesize Modality: CAM stands for \textit{Camera.} + is joint use of modalities, / is either modality. Auxiliary Modality: \RI\xspace stands for \textit{Range Image}, \B\xspace stands for \textit{BEV}, \V\xspace stands for \textit{Voxel}, \depth\xspace stands for \textit{Depth}. Paper type: \M\xspace stands for \textit{Method} and \D\xspace stands for \textit{Dataset}. Sensor Type: \textbf{NRS} is \textit{Non-Repetitive Scanning}  and \textbf{RMB} is \textit{Rotating Multi-Beam}. * indicates that the proposed HMR methods leverage weak supervision for 3D data annotation in their proposed datasets.
\end{tablenotes}
\end{table*}

\begin{table*}[t]
\tabcolsep 4pt

\caption{Comparative Analysis of 3D HPE and HMR Methods Based on Network Architectures.}

\newcolumntype{s}{>{\columncolor{gray!10}} c}
\newcolumntype{t}{>{\columncolor{gray!5}} c}

\rowcolors{4}{gray!10}{}

\resizebox{1.0\textwidth}{!}{
\begin{tabular}{lstststststp{0.35\textwidth}}

\toprule

\multirow{2}{*}{Method}
& 3D Sp. 
& Atten. 
& \multirow{-1}{*}{CNN}
& \multirow{-1}{*}{GCN}
& \multirow{-1}{*}{GRU}
& \multirow{-1}{*}{MLP}
& Optim. 
& \multirow{-1}{*}{PointNet}
& Point
& \multirow{-1}{*}{Trans.}
& \multirow{-1}{*}{Off-the-shelf Component} 
\\

&  Conv. & Mech. & & & & & Phase & &  Trans. &  & \\
\midrule

\rowcolor{white} 
\textbf{\textit{HPE Methods}} &  &  &  &  &    & &   &  &  & & \\

DAPT ~\cite{anPretrainingDensityAwarePose2025} &  & \cmark &   &   &  & \cmark  &  &  &  \cmark & & \\ 

\shortstack{FusionPose~\cite{congWeaklySupervised3D2023}}  &  & \cmark &   &   & \cmark  & \cmark  &  & \cmark &   & & 2D detector (YOLO v5)~\cite{yolov5}, 3D detector (STCrowd)~\cite{stcrowd}, 2D pose estimator OpenPose~\cite{cao2019openpose}, 2D feature extractior HRNet~\cite{hrnet} \\ 

GC-KPL~\cite{weng3DHumanKeypoints2023} &  &  &  &   &  & \cmark &  &  &   & \cmark & 3D flow estimator Scene Flow Prior~\cite{li2021neural} \\

HPERL~\cite{furstHPERL3DHuman2021} &  &  & \cmark &   &  &  \cmark &  &  & & & 2D feature extractor VGG-16~\cite{vgg16} \\ 

HUM3DIL~\cite{zanfirHUM3DILSemisupervisedMultimodal2022} &  &  & \cmark  &   &  &   &  &  &   & \cmark & \\ 

\shortstack{LiCamPose~\cite{panLiCamPoseCombiningMultiView2025}} &  &  & \cmark  &   &  &   &  &  &   & &  3D detector PointPillars~\cite{lang2019pointpillars}, 2D pose estimators~\cite{vitpose,fang2022alphapose,hrnet} \\ 

LiDAR-HMP ~\cite{hanPracticalHumanMotion2024} & & \cmark  & &   &  & \cmark  &  & \cmark &   & \cmark &  \\

LidPose~\cite{kovacsLidPoseRealTime3D2024} &  &  &   &   &  &   &  &  &  & \cmark &  \\

\shortstack{LPFormer~\cite{yeLPFormerLiDARPose2024}} & \cmark$\dagger$  & &  &   &  &   &  &  &  & \cmark & 3D Feature extractor LidarMultiNet~\cite{yeLidarMultiNetUnifiedMultiTask2023}. $\dagger$ from LidarMultiNet  \\ 

MMVP \cite{fanHumanM3MultiviewMultimodal2023} &  & & \cmark    &   &  &   &  &  &  & & 2D keypoint feature extractor OpenPIFPAF~\cite{kreiss2021openpifpaf}  \\ 
UniPVU-Human~\cite{xuUnifiedFrameworkHumancentric2024} &  & &     &   &  &   &  &  \cmark &  &  \cmark &  \\ 

VoxelKP~\cite{shiVoxelKPVoxelbasedNetwork2023}   &  \cmark  & \cmark &  &   &  &  \cmark &  &  &  &   & \\ 

WS-HPE~\cite{zhengMultimodal3DHuman2022} &   & &  \cmark  &   &  &   &  & \cmark  &  &  &  \\

WS-Fusion~\cite{bauerWeaklySupervisedMultiModal2023} &   & &   &   &  &  \cmark &  & \cmark  &  &  & 2D pose estimator AlphaPose \cite{fang2022alphapose} \\ 
\midrule

\rowcolor{white} 
\textbf{\textit{HPE \& HMR Methods}} &  &  &   &  &  & &   &  &  &  &\\

LiDARCap~\cite{liLiDARCapLongrangeMarkerless2022} &   &  & & \cmark  & \cmark & \cmark &  & \cmark  &  &  & SMPL module \\ 

LiDARCapV2~\cite{zhangLiDARCapV23DHuman2024} &   &  &   &  \cmark & \cmark & \cmark  &  & \cmark  &  & & SMPL module \\ 

LiDAR-HMR~\cite{fanLiDARHMR3DHuman2025} &   & \cmark &    &   \cmark &  &  \cmark &  &   & \cmark & & \\ 

LIP \cite{renLiDARaidInertialPoser2023} &   &  &  \cmark  &    &  \cmark &  \cmark  &  & \cmark  &  & &  SMPL module \\ 

\midrule

\rowcolor{white} 
\textbf{\textit{HMR Methods}} & & & & & & & & & & &  \\

CIMI4D\cite{daiSLOPER4DSceneAwareDataset2023} &   &  & & &  & & \cmark &  &  & &  \\ 

\shortstack{FreeCap \cite{xueFreeCapHybridCalibrationFree2025}} &   & \cmark   &    &    &  & \cmark  & \cmark &   &  & & 2D pose estimator RTMPose~\cite{jiang2023rtmpose}, image-based SMPL pose estimator WHAM~\cite{wham:cvpr:2024}, 3D pose estimator LiveHPS~\cite{renLiveHPSLiDARbasedScenelevel2024} \\ 

\shortstack{HSC4D \cite{daiHSC4DHumancentered4D2022}} &   &  &   &    &  &   & \cmark &   &  & & Ground points extractor CSF~\cite{zhang2016easy}, plane fitter RANSAC~\cite{schnabel2007efficient} \\ 

\shortstack{Human-M3 \cite{fanHumanM3MultiviewMultimodal2023} } &   &  &    &    &  &   & \cmark &   &  & & 3D detector PointPillars~\cite{lang2019pointpillars}, tracker AB3DMOT \cite{weng2020ab3dmot}, 2D Pose estimator ViTPose \cite{vitpose}, VPoser \cite{pavlakos2019expressive} \\

LiveHPS \cite{renLiveHPSLiDARbasedScenelevel2024} &   & \cmark &    &    & \cmark &  \cmark &  & \cmark  &  & &  \\

LiveHPS++ \cite{renLiveHPSRobustCoherent2024} &   & \cmark &   &    & \cmark & \cmark  &  & \cmark  &  & &   \\

NE-3D-HPE \cite{zhangNeighborhoodEnhanced3DHuman2024} &   & \cmark &  &  \cmark   &  &  \cmark &  & \cmark  &  & & SMPL module \\ 

PEAR-Proj \cite{linHmPEARDatasetHuman2024} &   &  & \cmark &  \cmark   & \cmark &   &  & \cmark  &  & & 2D feature extractor HRNet~\cite{hrnet} \\ 

ReMP \cite{jangReMPReusableMotion2025} &   & \cmark &   &    &  & \cmark  &  &  & &  \cmark & off-the-shelf SMPL \\ 

\shortstack{SLOPER4D \cite{daiSLOPER4DSceneAwareDataset2023}} &   &  &    &    &  &    & \cmark &  &  & & 2D detector Detectron \cite{wu2019detectron2}, tracker Deepsort \cite{wojke2017simple} \\ 

\shortstack{SMPLify-3D~\cite{dumontImprovingImagebased3D}} &   &  &    &    &  &    & \cmark &  &  &  & Image-based SMPL pose estimator CLIFF~\cite{li2022cliff}, SMPL module\\ 

\bottomrule
\end{tabular}}
\label{tab:rm_arch2}
\begin{tablenotes} 
    \item 
    \footnotesize Optim. stands for \textit{Optimization}, Trans. stands for \textit{Transformer}, Atten. Mech. stands for \textit{Attention Mechanism}. Sp. Conv. stands for \textit{Sparse Convolution}.
As attention is inherent to Transformers, we report it only when used independently.
\end{tablenotes}
\end{table*}

\paragraph{\textbf{Limitations of Sensors in Data Acquisition}\label{challenges_and_issues_lidar_radar}}

When deployed in the real world, LiDARs are particularly vulnerable to adverse weather such as fog, rain, snow, and even dust, which are collectively referred to as \textit{weather-induced} noise in the literature~\cite{dreissigSurveyLiDARPerception2023}. Under these conditions, the emitted laser beams experience back-scattering and signal attenuation, resulting in sparser point clouds and significant noise. Beyond weather, another limitation arises from glass surfaces, which contradict the reflectivity principle central to LiDAR operation and thus pose additional challenges.

RADARs are generally more robust than LiDARs in adverse weather due to their use of longer wavelengths. However, their performance strongly depends on the target material and reflectivity, and they offer limited resolution, making them unreliable for distinguishing between different obstacle types at long ranges \cite{martiReviewSensorTechnologies2019}. For instance, metal surfaces can amplify RADAR signals, whereas materials such as wood and glass may remain nearly invisible. 

Cameras provide high-resolution imagery but suffer from a restricted horizontal field of view and extreme sensitivity to lighting conditions such as glare or over/under-exposure. Figure \ref{fig:sensors_spider_criteria} summarizes these sensor-specific shortcomings in a comparative visual form.

Most modern AV systems integrate multiple onboard proprioceptive sensors to achieve a more accurate understanding of the surrounding environment. This process, known as multi-sensor fusion \cite{wangMultiSensorFusionAutomated2020}, aggregates data from ultrasound sensors, cameras, LiDARs, RADARs, and others to compensate for the limitations of individual sensors.

For a comprehensive overview of sensing technologies and their challenges, especially in AV/ADAS systems, refer to the following surveys \cite{martiReviewSensorTechnologies2019,liLidarAutonomousDriving2020,guesmiNavigatingThreatsSurvey2024}.

\section{Overview}\label{sec:overview}

In Table~\ref{tab:rw}, we provide a comprehensive summary of existing methods for 3D HPE and HMR in in-the-wild LiDAR point clouds. The comparison is structured along several key dimensions: input modality; auxiliary representations of point clouds e.g., bird’s-eye view (BEV), depth maps; usage of temporal information; type of supervision employed; type of learning, indicating whether the model is trained for a single task or multiple tasks simultaneously; inclusion of synthetic data; whether the contribution is methodological or dataset-oriented; the configuration of LiDAR sensors (single or multi-view); and, where available, real-time inference capabilities. We also include the type of LiDAR sensor used, distinguishing mainly between Non-Repetitive Scanning (NRS) and Rotating Multi-Beam (RMB) sensors.

In Table~\ref{tab:rm_arch2}, we outline the architectural designs in the reviewed methods. Notably, PointNet and its variants are frequently used due to their natural ability to process unordered point sets. 
Several recent methods move beyond convolutional models to adopt transformers, leveraging their capacity to model complex spatial relationships, an essential trait for unstructured LiDAR data. We observe a diverse spectrum of transformer usage: from transformer-centric architectures to hybrid systems where transformers handle decoding or motion reasoning.
In addition, several methods integrate off-the-shelf components for targeted sub-tasks, such as 2D/3D person detection, 2D/3D feature extraction, and 2D pose estimation.

\begin{figure*}[ht]

\tikzset{
basic/.style  = {draw, text width=6cm, drop shadow, font=\large, rectangle},
basic1/.style  = {draw,  drop shadow, font=\sffamily, rectangle},
root/.style   = {basic, rounded corners=2pt, thin, align=center, fill=mh1!20,text width=4.5cm},
secnode/.style = {basic, thin, rounded corners=2pt, align=center, fill=mh1!20,text width=4.5cm,},
subsecnode/.style = {basic, thin, rounded corners=2pt, align=center, fill=mh4!20,text width=4.5cm,},
parnode/.style = {basic, thin, align=center, fill=mh7!20},
edge from parent/.style={draw=black, edge from parent fork right}
}

\centering
\resizebox{\textwidth}{!}{
\begin{forest}
    multiple directions={minimum height=4ex, anchor=center, forked edge}
  [\large {\shortstack{LiDAR-based \\ 3D HPE/HMR}}, root, grow'=south
    [, grow' subtree=east
    [3D Human Mesh Recovery\\from LiDAR Point Clouds~\ref{sec:hmr}, secnode
      [LiDAR-only Human\\Mesh Recovery~\ref{subsec:lidar_only_hmr}, subsecnode
        [Sparse-to-Dense\\Reconstruction~\ref{par:sparse_to_dense_recon}, parnode]
        [Spatio-temporal Modeling\\for Robustness~\ref{par:st_modeling_robustness}, parnode]
        [Distillation-based\\ Latent Prior Learning~\ref{par:ws_latent_prior}, parnode]
      ]
      [Fusing LiDAR and\\Other Modalities for\\Human Mesh Recovery~\ref{subsec:fusing_lidar_and_other_modalities_hmr}, subsecnode
        [Calibration-free\\Fusion~\ref{par:calib_free_fusion}, parnode]
        [Fusion from LiDAR\\and IMU/RGB~\ref{par:fusion_lidar_imu}, parnode]
        [Multi-modal Optimization\\Pipelines~\ref{par:multi-modal_optim}, parnode]
      ]
    ]]
    [, grow subtree=west
    [3D Human Pose Estimation\\from LiDAR Point Clouds~\ref{sec:hpe}, secnode,
      [Supervised Human\\Pose Estimation~\ref{subsec:supervised_hpe}, subsecnode
        [Sparsity-conscious\\Design~\ref{par:spasity_conscious_design}, parnode]
        [Transformers as the\\Backbone~\ref{par:transformers_backbone}, parnode]
        [Supervision Beyond\\Pose~\ref{par:supervision_beyond_pose}, parnode]
        [Learning from Synthetic\\Data~\ref{par:learning_from_synth_data}, parnode]
      ]
      [Weakly-supervised\\Human Pose Estimation~\ref{subsec:ws_hpe}, subsecnode
        [Bridging the Annotation\\Gap via Weak Supervision~\ref{par:briging_annotation_gap}, parnode]
        [Bridging Modalities via\\Fusion~\ref{par:bridging_modality}, parnode]
      ]
      [Unsupervised Human\\Pose Estimation~\ref{subsec:unsup_hpe}, subsecnode]
    ]]
  ]
\end{forest}}
\caption{\textbf{Systematic categorization of 3D HPE and HMR methods from LiDAR point clouds.} We note that a single method may fall into multiple categories.}
\label{fig:taxonomy}
\end{figure*}
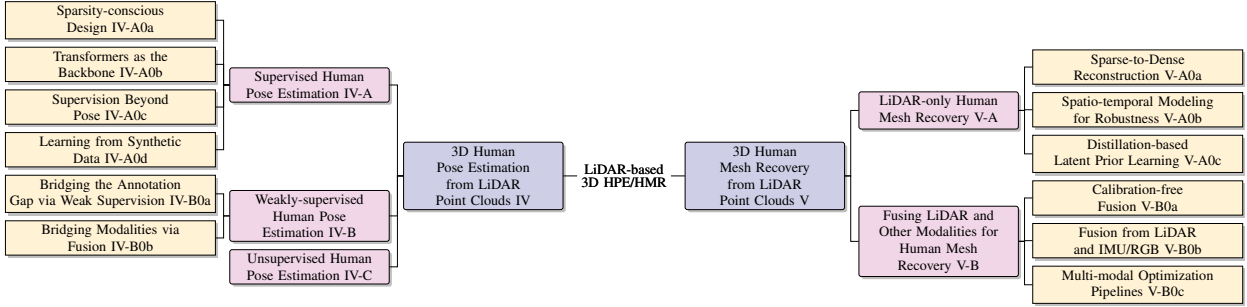

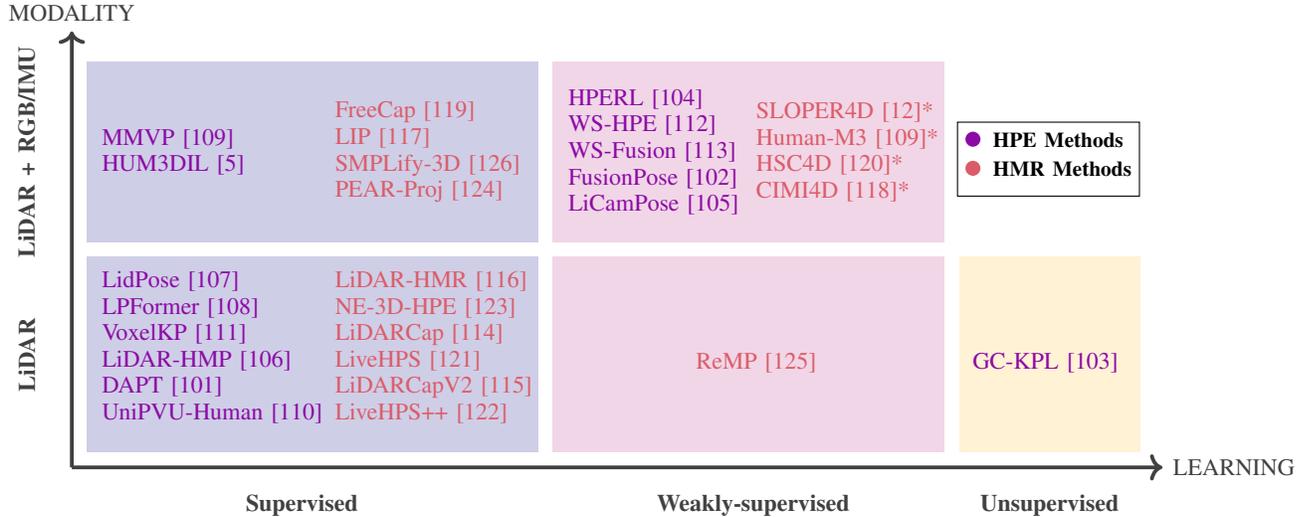
\begin{figure*}[h]
\centering
\begin{tikzpicture}[
  font=\small,
  yellow/.style={text=orange!70!black, align=left},
  red/.style={text=red!70!black, align=left},
  blue/.style={text=blue!70!black, align=left}
]

\draw[->, line width=0.5mm, color=darkgray] (0,0) -- (14.5,0) node[right] {LEARNING};
\draw[->, line width=0.5mm, color=darkgray] (0,0) -- (0,5.8) node[above] {MODALITY};

\node at (3, -0.5) {\textbf{ \textcolor{darkgray} {Supervised}}};
\node at (9, -0.5) {\textbf{ \textcolor{darkgray} {Weakly-supervised}}};
\node at (13, -0.5) {\textbf{\textcolor{darkgray} {Unsupervised}}};

\fill[mh1!20] (0.2,0.2) rectangle (6.2, 2.8);  
\fill[mh1!20] (0.2,3) rectangle (6.2, 5.4);  

\fill[mh4!20] (6.4, 0.2) rectangle (11.6,2.8);
\fill[mh4!20] (6.4, 3) rectangle (11.6,5.4);

\fill[mh7!20] (11.8,0.2) rectangle (14.2 , 2.8);

\node[rotate=90] at (-0.6, 1.5) {\textbf{\textcolor{darkgray} {LiDAR}}};
\node[rotate=90] at (-0.6, 4.2) {\textbf{\textcolor{darkgray} {LiDAR + RGB/IMU}}};

\node[color=mh3, align=left, text width=4cm] at (2.4, 1.6) {
   LidPose \cite{kovacsLidPoseRealTime3D2024} \\ 
  LPFormer \cite{yeLPFormerLiDARPose2024} \\ VoxelKP \cite{shiVoxelKPVoxelbasedNetwork2023}\\
  LiDAR-HMP \cite{hanPracticalHumanMotion2024} \\ DAPT \cite{anPretrainingDensityAwarePose2025} \\ UniPVU-Human \cite{xuUnifiedFrameworkHumancentric2024}
};
\node[color=mh3, align=left, text width=4cm] at (2.4, 4.2) {
   MMVP \cite{fanHumanM3MultiviewMultimodal2023} \\
   HUM3DIL \cite{zanfirHUM3DILSemisupervisedMultimodal2022} 
};

\node[color=mh5, align=left, text width=4cm] at (5.5, 4.2) {
   FreeCap \cite{xueFreeCapHybridCalibrationFree2025} \\ LIP \cite{renLiDARaidInertialPoser2023}\\ SMPLify-3D \cite{dumontImprovingImagebased3D}  \\ PEAR-Proj \cite{linHmPEARDatasetHuman2024} 
};
\node[color=mh5, align=left, text width=4cm] at (5.5, 1.6) {
   LiDAR-HMR \cite{fanLiDARHMR3DHuman2025} \\ NE-3D-HPE \cite{zhangNeighborhoodEnhanced3DHuman2024} \\
  LiDARCap \cite{liLiDARCapLongrangeMarkerless2022} \\ LiveHPS \cite{renLiveHPSLiDARbasedScenelevel2024} \\
 LiDARCapV2 \cite{zhangLiDARCapV23DHuman2024} \\ LiveHPS++ \cite{renLiveHPSRobustCoherent2024}
};

\node[color=mh3, align=left,  text width=4cm] at (8.6, 4.2) {
    HPERL \cite{furstHPERL3DHuman2021} \\
    WS-HPE \cite{zhengMultimodal3DHuman2022} \\
    WS-Fusion \cite{bauerWeaklySupervisedMultiModal2023} \\
    FusionPose \cite{congWeaklySupervised3D2023}\\
    LiCamPose \cite{panLiCamPoseCombiningMultiView2025}
};
\node[color=mh5, align=left, text width=4cm] at (11.1, 4.2) {
    SLOPER4D \cite{daiSLOPER4DSceneAwareDataset2023}* \\ Human-M3 \cite{fanHumanM3MultiviewMultimodal2023}* \\ HSC4D \cite{daiHSC4DHumancentered4D2022}* \\ CIMI4D~\cite{yanCIMI4DLargeMultimodal2023}*
};

\node[color=mh5, align=left,text width=4cm] at (10.3, 1.4) {ReMP \cite{jangReMPReusableMotion2025} };

\node[color=mh3, align=left] at (13, 1.4) {GC-KPL \cite{weng3DHumanKeypoints2023} };

\node[draw, inner sep=3pt, font=\footnotesize, anchor=north east ] at (14.2, 4.6){

    \begin{tabular}{@{}cl@{}}
 \tikz\fill[mh3] (0,0) circle (0.1); &  \hspace{-10pt} \textbf{HPE Methods}\\[2pt]
 \tikz\fill[mh5](0,0) circle (0.1); & \hspace{-10pt} \textbf{HMR Methods}\\[2pt]
    \end{tabular}
};
\end{tikzpicture}
\caption{\textbf{Distribution of 3D HPE and HMR methods across input modalities and supervision levels.} We observe that supervised HPE methods rely on LiDAR, whereas weakly supervised approaches tend to incorporate additional modalities alongside LiDAR. Notably, the use of reduced supervision remains largely unexplored in HMR. * indicates that the proposed HMR methods leverage weak supervision for 3D data annotation in their proposed datasets.}
\label{fig:modality_vs_learning}
\end{figure*}

Finally, Figure~\ref{fig:taxonomy} presents our taxonomy, which we derived from common trends across existing methods. Complementing this, Figure~\ref{fig:modality_vs_learning} details how input modalities interact with learning paradigms, offering insights for potential future research directions.

\section{3D Human Pose Estimation from LiDAR}\label{sec:hpe}

We categorize \hpes methods into three main groups based on their learning paradigms: supervised, weakly supervised, and unsupervised. Supervised HPE methods rely on full 3D supervision, i.e., they are trained with ground-truth 3D pose annotations. 
Weakly supervised methods, in contrast, relax this requirement by relying only on 2D keypoint annotations, which are much easier to collect, either through manual labeling or by leveraging off-the-shelf 2D pose detectors. Finally, unsupervised methods avoid any human pose annotations altogether during training, instead exploiting alternative supervisory signals such as self-consistency, geometric priors, and synthetic data.

\subsection{Supervised Human Pose Estimation}\label{subsec:supervised_hpe}

As the demand for robust and privacy-aware human perception systems grows, LiDAR-based \hpes has emerged as a promising alternative to traditional image-based or multi-modal approaches. Unlike cameras, LiDAR sensors offer rich depth information and are less intrusive in privacy-sensitive environments. Although leveraging LiDAR point clouds alone for accurate HPE is challenging due to their inherent sparsity, non-uniformity, and often incompleteness, recent works have tackled the task with supervised learning, new architectures, training strategies, and data representations.

\paragraph{Sparsity-conscious Design}\label{par:spasity_conscious_design}
These approaches confront the sparsity and irregularity of LiDAR point clouds, which dramatically affect robustness.

LidPose~\cite{kovacsLidPoseRealTime3D2024}  targets Non-Repetitive Scanning (NRS) LiDARs, which produce non-uniform, sparse, motion-artifact-ridden frames. LidPose maps the scene point cloud into a 2D range image and uses a Mixture-of-Gaussians model to separate foreground from background. The foreground is processed to extract individual human instances, which are then projected onto the ground plane to estimate bounding boxes.
Then, it applies a ViTPose-inspired transformer~\cite{vitpose} on an instance-level structured multi-channel (global coordinates, depth, intensity) representation. 

LPFormer~\cite{yeLPFormerLiDARPose2024} has a two-stage top-down design, starting with detection and segmentation via LiDARMultiNet~\cite{yeLidarMultiNetUnifiedMultiTask2023}, which captures the whole scene both globally, with BEV, and at the point level, with sparse-3D-voxels. This combination of features helps overcome nonuniform density. Keypoint prediction follows by conditioning a transformer on features inside predicted 3D human boxes. 

In contrast, VoxelKP~\cite{shiVoxelKPVoxelbasedNetwork2023} embraces the sparsity by voxelizing the whole scene into a high-resolution volumetric point cloud and learning with a hierarchy of sparse 3D convolutions. It integrates different attention mechanisms to gather fine-grained spatial context without dense supervision or intermediate image-like projections. Sparse-Selective-Kernel attention gathers spatial context by aggregating multi-scale features, while Sparse-Box attention captures dependencies between voxels belonging to non-overlapping scene partitions. Hybrid MLP and convolution features learn the fine-to-coarse fusion.
VoxelKP forgoes prior bounding box detection.

LiDAR-HMP~\cite{hanPracticalHumanMotion2024} focuses on end-to-end motion forecasting using raw human instance point cloud sequences. 
It encodes the human body structure using global and part-aware features obtained from point-based networks, and then attends to these features with motion queries in a transformer. The resulting spatio-temporal features provide a coarse joint estimation, which is progressively refined by spatio-temporal transformer layers.
It is tailored to learn from unfiltered, sparse, and partial inputs, relying on auxiliary point-wise body-part segmentation and feature fusion to build robust motion representations.

MMVP~\cite{fanHumanM3MultiviewMultimodal2023} and HUM3DIL~\cite{zanfirHUM3DILSemisupervisedMultimodal2022} address LiDAR sparsity with modality fusion.  
Human-M3~\cite{fanHumanM3MultiviewMultimodal2023} introduces MMVP as a simple baseline for \hpe on their multi-modal benchmark. 
MMVP builds a structured volumetric representation, incorporating RGB data to compensate for incomplete 3D information to detect human instances. It starts by extracting 2D keypoint heatmaps from RGB images using OpenPIFPAF \cite{kreiss2021openpifpaf}, which it projects onto the 3D space using camera-LiDAR calibration, effectively lifting dense image-based pose cues to 3D. Meanwhile, it incorporates regularity and spatial context by voxelizing the LiDAR into an occupancy volume.
The 3D heatmaps and voxelized map are fused by combining the spatial completeness of RGB-derived cues with the geometric fidelity of LiDAR to obtain proposals for human instances. These fused representations are then processed by a 3D CNN to regress the final joint heatmaps.
Similarly, HUM3DIL~\cite{zanfirHUM3DILSemisupervisedMultimodal2022} fuses 2D and 3D features, enriching LiDAR points with visual context. After combining the RGB image with a depth representation of the point cloud, HUM3DIL generates a dense feature map from the RGB-D input using a U-Net \cite{unet}. 
This feature map is queried by the point cloud via 2D projection to obtain pixel-aligned features.
Simultaneously, LiDAR point features are extracted using a Random Fourier Encoding. These pixel-aligned features, per-point features and camera intrinsics are combined and used as input tokens to a transformer that predicts the 3D pose.

UniPVU-Human~\cite{xuUnifiedFrameworkHumancentric2024} and DAPT~\cite{anPretrainingDensityAwarePose2025} first learn a generalizable representation of humans with powerful features, which are subsequently transferred to downstream tasks such as human pose estimation.
UniPVU-Human addresses sparsity with multiple stages, first extracting structured human-specific priors before estimating the poses. The initial Prior Knowledge Extraction trains body-part segmentation and motion flow estimation on synthetic SMPL-based point clouds, thus supporting representation learning with human-centric priors. 
Then, a semantic-guided spatio-temporal learning is performed, where temporal and spatial masking are applied to part-segmented human sequences, which are subsequently embedded into part-level features using a Mini PointNet~\cite{qi2017pointnet}.
These features are leveraged for human point cloud sequence reconstruction, enabling the model to address input sparsity.

On the other hand, DAPT~\cite{anPretrainingDensityAwarePose2025} addresses irregularity with a Density-Aware Pose Transformer and sparsity with Multi-Density Exchange, which enables learnable joint-specific embeddings to interact with human point features across varying density. Instead of numeric coordinates, DAPT takes human instance point clouds and outputs 1D heatmaps along each spatial axis, thus avoiding coordinate regression, which is unstable under sparsity. Its design tolerates severe point dropout, occlusions, and varying density.

Collectively, these methods illustrate a spectrum of sparsity-handling strategies: 
structured projection (LidPose~\cite{kovacsLidPoseRealTime3D2024}), 
hybrid BEV + voxel fusion (LPFormer~\cite{yeLPFormerLiDARPose2024}), 
fully sparse voxel processing (VoxelKP~\cite{shiVoxelKPVoxelbasedNetwork2023}), 
direct sequence modeling (LiDAR-HMP\cite{hanPracticalHumanMotion2024}), 
RGB-LiDAR modality fusion (MMVP~\cite{fanHumanM3MultiviewMultimodal2023}, HUM3DIL \cite{zanfirHUM3DILSemisupervisedMultimodal2022}),
self-supervised masked representation learning (UniPVU-Human~\cite{xuUnifiedFrameworkHumancentric2024}) and
density-aware joint-attentive transformers (DAPT~\cite{anPretrainingDensityAwarePose2025}).

\paragraph{Transformers as the Backbone}\label{par:transformers_backbone}

LiDAR-only \hpes has followed computer vision's trend of employing transformers for modeling long-range dependencies and irregular input structures. Previously mentioned challenges of LiDAR point clouds motivate the use of transformers, which offer a global receptive field and attention-based information fusion. 

In~\cite{kovacsLidPoseRealTime3D2024, yeLPFormerLiDARPose2024, zanfirHUM3DILSemisupervisedMultimodal2022}, transformers refine per-instance or per-box predictions using attention over structured local features.
LidPose~\cite{kovacsLidPoseRealTime3D2024} relies on a ViT-inspired architecture to process per-instance multi-channel representations of humans and predict 3D skeletons, thus leveraging the transformer's ability to model intra-person joint relations even under partial or complete absence of some joints. 
LPFormer~\cite{yeLPFormerLiDARPose2024} first detects human bounding boxes and extracts voxel-level and BEV features. In the second stage, its Keypoint Transformer  takes these localized spatial features along with keypoint queries, fuses them using multi-head attention, and predicts joint positions and visibilities.
While \cite{kovacsLidPoseRealTime3D2024, yeLPFormerLiDARPose2024} apply transformers only to LiDAR features, HUM3DIL~\cite{zanfirHUM3DILSemisupervisedMultimodal2022} incorporates pixel-aligned RGB-D features and camera intrinsics alongside LiDAR point features to predict 3D human pose.

For global end-to-end modeling, transformers drive the entire pipeline, including feature encoding and joint estimation, with attention adapted to varying density. More specifically, DAPT~\cite{anPretrainingDensityAwarePose2025} focuses on a fully transformer-based encoder-decoder. It uses a U-Net-style transformer architecture (based on PTv3~\cite{wu2024point}) where both encoder and decoder are composed of point transformers operating on sparse LiDAR point sets. 
It introduces learnable joint anchors to stabilize learning across regions with varying point densities. 
Here, the transformer goes beyond passively aggregating features by actively learning where to look based on point cloud sparsity, thus enabling robustness in scenes with occluded joints.

Similarly, UniPVU-Human~\cite{xuUnifiedFrameworkHumancentric2024} also employs a fully transformer-based encoder-decoder. Transformer-based spatio-temporal self-attention modules are used in the semantic-guided representation self-learning stage. The model masks tokens representing specific body parts across time and space, then learns to reconstruct them.

For human motion forecasting~\cite{hanPracticalHumanMotion2024}, transformers are employed for both spatial decoding and temporal modeling.
LiDAR-HMP~\cite{hanPracticalHumanMotion2024} extracts a structure-aware body descriptor from segmented human point clouds (via PointNet and part clustering), and applies spatial and temporal transformers alternatively to forecast future poses and point clouds.
STFormer (spatial transformer) captures intra-frame dependencies (e.g., how arms and legs move in relation). TTFormer (temporal transformer) models the dynamics across frames, predicting plausible future motion from observed segments.
Here, the transformer’s attention mechanism becomes essential for temporal consistency and fine-grained motion forecasting, especially when historical frames are sparse or partially occluded.

Even without explicit transformers~\cite{shiVoxelKPVoxelbasedNetwork2023}, attention-like modules help capture contextual relationships crucial for keypoint estimation.
VoxelKP~\cite{shiVoxelKPVoxelbasedNetwork2023} mirrors the philosophy of attention through modules like Sparse Box Attention, which captures localized voxel dependencies in a structured grid, and Sparse Selective Kernel modules, which selectively aggregate features at different receptive fields. 
While not transformer-labeled, they show that attention mechanisms, whether explicit or architectural, can deal with LiDAR point cloud irregularities.

\paragraph{Supervision Beyond 3D Pose}\label{par:supervision_beyond_pose}

Many works use auxiliary tasks (e.g., body part segmentation, human motion forecasting or keypoint prediction) to guide learning towards more structured and reusable features.

LPFormer~\cite{yeLPFormerLiDARPose2024} benefits from the auxiliary task of body part segmentation of the human body. By jointly learning to identify human body regions and predict keypoints, it becomes more robust to occlusions and partial views.

LiDAR-HMP~\cite{hanPracticalHumanMotion2024} takes this idea further by forecasting the future point cloud sequence for auxiliary supervision. The final output includes predicted joints and future human point cloud frames.

HUM3DIL~\cite{zanfirHUM3DILSemisupervisedMultimodal2022} leverages 2D annotations as auxiliary signals to reinforce 3D supervision at the human-instance level.
It projects its 3D predictions onto the image plane using known camera parameters and supervises them using 2D annotations.

Works such as~\cite{anPretrainingDensityAwarePose2025,xuUnifiedFrameworkHumancentric2024} use auxiliary supervision for warm-up.
UniPVU-Human~\cite{xuUnifiedFrameworkHumancentric2024} makes use of Prior Knowledge Extraction to train two auxiliary modules on a synthetic dataset before learning the main representation. The Human Body Segmentation (HBSeg) module segments each frame into nine anatomically distinct body parts using PointNeXt-L~\cite{qian2022pointnext}. The Human Motion Flow Estimation (HMFlow) module captures temporal dynamics by estimating inter-frame motion and learning soft point-wise correspondences using FLOT-based optimal transport~\cite{flot}. 
Similarly, DAPT~\cite{anPretrainingDensityAwarePose2025} extracts a human prior using keypoint regression and body part segmentation tasks from synthetically generated human point clouds.
These pretrained modules enrich the token representation and inject explicit geometric and temporal priors into the representation learning.

\paragraph{Learning from Synthetic Data}\label{par:learning_from_synth_data}

LiDAR-only HPE faces the key challenge of lacking large-scale annotated datasets, which is mainly solved by synthetic data generation and aggressive augmentation.
DAPT~\cite{anPretrainingDensityAwarePose2025} and  UniPVU-Human~\cite{xuUnifiedFrameworkHumancentric2024} have a synthetic pipeline based on SMPL parametric models.  
DAPT~\cite{anPretrainingDensityAwarePose2025} places human meshes in randomized poses and environments, then scans them with a simulated LiDAR via ray casting. Their body and shape parameters are sampled from real-world SMPLs provided by LiDARHuman26M \cite{liLiDARCapLongrangeMarkerless2022}, which are annotated using IMUs. This process generates realistic and diverse point clouds. Additionally, SMPL’s face label map is used to assign point-wise body part labels, while occlusion is simulated through laser masking.

UniPVU-Human~\cite{xuUnifiedFrameworkHumancentric2024} adopts a similar pipeline for ray casting and point-wise part labeling of LiDAR point clouds, where body and shape parameters are sampled from the AMASS motion dataset~\cite{AMASS:ICCV:2019}. Furthermore, it leverages SMPL vertices across consecutive frames as proxies to estimate the motion flow of the generated point clouds. Both methods significantly improve performance by pretraining on the synthetic data and fine-tuning on real data. 

\begin{figure}[ht]
    \centering
    \includegraphics[width=0.95\linewidth]{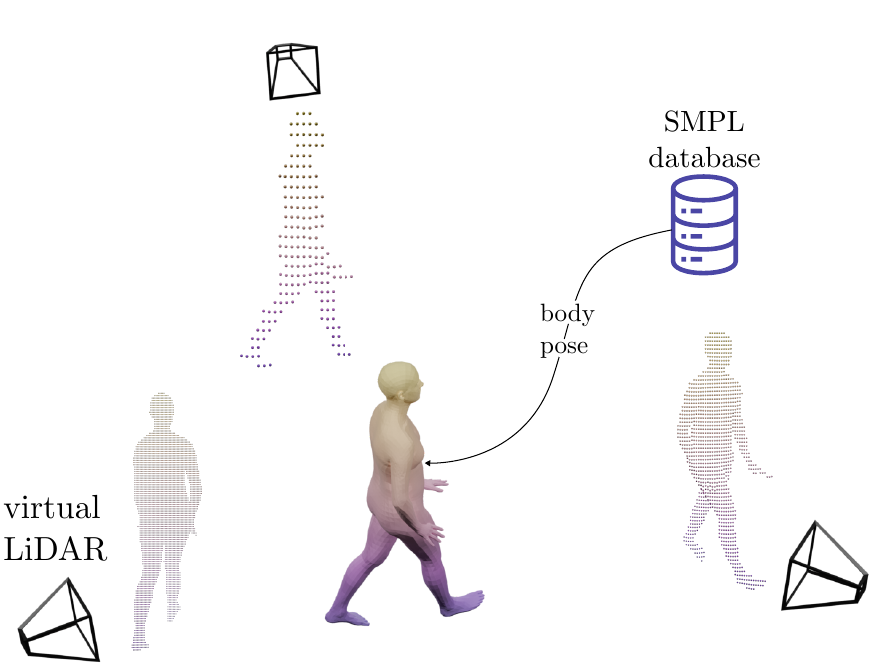}
    \caption{\textbf{Overview of a synthetic LiDAR point cloud generation pipeline.} SMPL meshes are sampled from a database such as AMASS~\cite{AMASS:ICCV:2019} and scanned using a simulated LiDAR via ray casting. This pipeline is commonly adopted by 3D HPE and HMR methods covered in this survey that leverage synthetic data for training.} 
    \label{fig:synt-data-gen}
\end{figure}

Figure \ref{fig:synt-data-gen} illustrates the key steps involved in generating realistic synthetic LiDAR point cloud datasets. This pipeline is commonly employed in HPE and HMR methods that utilize synthetic data as part of their overall training strategy. 

Meanwhile, VoxelKP~\cite{shiVoxelKPVoxelbasedNetwork2023}, LPFormer~\cite{yeLPFormerLiDARPose2024} and LidPose~\cite{kovacsLidPoseRealTime3D2024} rely on data augmentations such as frustum dropout, mixup of frames, random upper/lower body cropping and random transformations to increase sample diversity. They focus on perturbing real LiDAR scenes to simulate motion, occlusion, and noise. 

\begin{figure*}[t]
    \centering

    \includegraphics[width=0.8\linewidth]{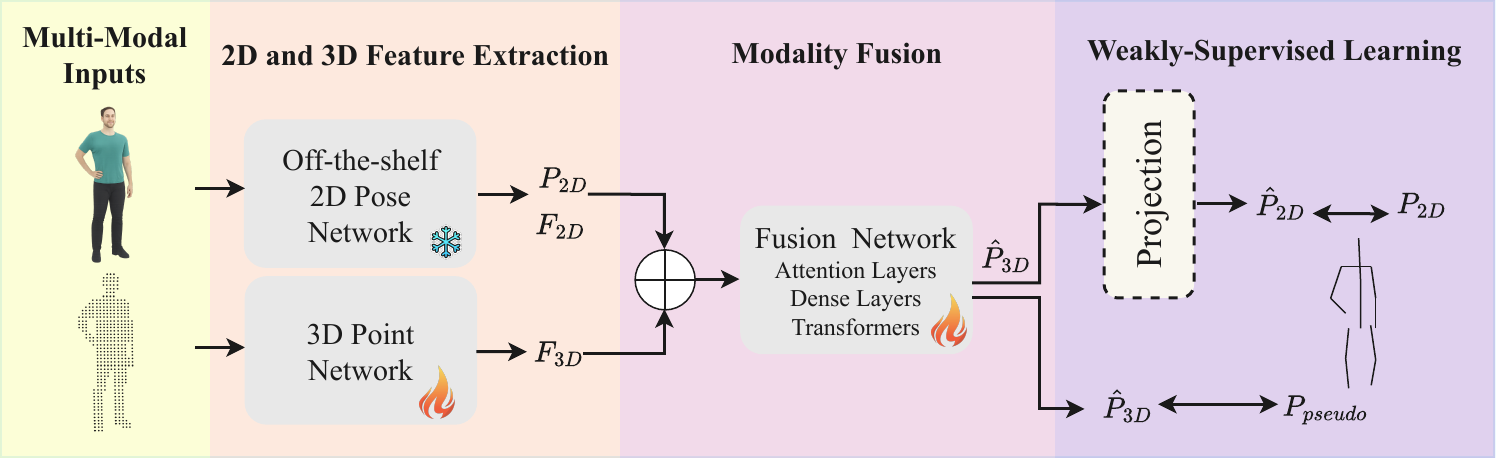}
    \caption{\textbf{Overview of key components in weakly-supervised methods.} An essential component in these pipelines is the use of 2D pose information. This may take the form of 2D poses ($P_{2D}$), either from ground-truth annotations or obtained via an off-the-shelf detector, or 2D features ($F_{2D}$) extracted by an off-the-shelf 2D pose estimator. These 2D cues are then fused with 3D features ($F_{3D}$) through a dedicated fusion network, which can be implemented as an attention mechanism, a dense layer, or a transformer-based module to predict 3D poses $\hat{P}_{3D}$. Dashed lines indicate optional blocks.} 
    \label{fig:ws_pipeline}
\end{figure*}

\subsection{Weakly-supervised Human Pose Estimation}\label{subsec:ws_hpe}

In autonomous systems, leveraging multi-sensor fusion between LiDAR and RGB data opens new possibilities for \hpes. 
Yet, the cost of dense 3D annotations remains a major bottleneck.
Recent works focus on weak supervision by utilizing 2D annotations to bypass the need for extensive 3D labels. These 2D annotations may be ground-truth labels or pseudo-labels obtained from pretrained off-the-shelf networks. Modality fusion and temporal cues integration are key strategies for the weak supervision of \hpes from LiDAR data. We present a typical pipeline for weakly supervised methods, with common components illustrated in Figure~\ref{fig:ws_pipeline}.

\paragraph{Bridging the Annotation Gap via Weak Supervision}\label{par:briging_annotation_gap}

As shown in Figure~\ref{fig:modality_vs_learning}, all weakly supervised methods share the trait of relying on multi-modal inputs.
Pseudo-labeling, projection consistency, auxiliary tasks as well as the design of loss functions are all strategies that mitigate the annotation bottleneck.

HPERL~\cite{furstHPERL3DHuman2021} outputs 2D and 3D predictions supervised entirely by 2D annotations. The method uses a subset of pre-defined 2D and 3D pose anchors from LCRNet~\cite{rogez2017lcr} 
and combines them with regions of interest extracted from both the whole LiDAR scene and image input. Next, it refines the anchor poses with model-predicted pose deltas. The weak supervision combines 2D as well as projection-based 3D losses, applying a smooth L1 loss against the 2D keypoints.

In contrast, WS-HPE~\cite{zhengMultimodal3DHuman2022} operates at human instance level and directly generates 3D pseudo-labels from 2D keypoints through projection. The method requires reliable camera intrinsics and extrinsics to allow projecting LiDAR points onto the image plane and then lifts each 2D ground-truth keypoint to 3D by associating it with the nearest projected LiDAR points.
WS-HPE incorporates a part-segmentation task based on pseudo point-wise 3D labels generated through a pipeline similar to the one described above.
The pose estimation branch is trained with a regression Huber loss using the generated 3D labels, while the segmentation branch is trained with cross-entropy. The auxiliary segmentation branch introduces a dense learning signal that compensates for the sparsity of pose supervision.

Also operating at human instance level, WS-Fusion~\cite{bauerWeaklySupervisedMultiModal2023} takes weak supervision further by training without any ground-truth labels in the target dataset.
The only supervision signal comes from 2D keypoints predicted on the input images. These predictions are obtained using the off-the-shelf extractor AlphaPose~\cite{fang2022alphapose} 
and lifted to 3D pseudo-labels by projecting the human LiDAR points onto image space and computing a weighted average of 3D coordinates near each 2D keypoint annotation. In contrast to WS-HPE~\cite{zhengMultimodal3DHuman2022}, WS-Fusion proposes a weighting directly in the point cloud space instead of the image plane, arguing that the 3D spatial structure should guide the pseudo-label construction. The resulting 3D keypoints supervise the network through an MPJPE loss. 

FusionPose~\cite{congWeaklySupervised3D2023} applies weak supervision in a temporally structured setting. It forgoes 3D pose annotations in favor of 2D multi-person pseudo-labels generated by OpenPose~\cite{cao2019openpose}. These 2D predictions guide training with three losses supervising the keypoint motion map, the consistency between the projection of the predicted 3D pose and the 2D detections, 
and enforcing stability of the predicted joints by bounding them to an average value computed over a time window.
Additionally, the method reinforces the geometric coherence of the output with the Chamfer Distance between the predicted 3D keypoints and the observed human point cloud.

LiCamPose~\cite{panLiCamPoseCombiningMultiView2025} adopts a hybrid approach combining weak supervision and domain adaptation. It first performs supervised pretraining on synthetic data, where human avatar skeletons are animated using 3D annotations from standard pose datasets. A LiDAR point cloud is then simulated via NRS-based ray casting. Finally, the model is fine-tuned on real-world data using unsupervised losses. 
The key idea is an entropy-based filtering strategy on 3D heatmaps to select confident predictions as 3D pseudo-labels for the next epoch. It supervises learning through symmetry losses, bone-length and plausible angle constraints and a projection-based loss using 2D pseudo-labels from the RGB inputs extracted via ViT-Pose~\cite{vitpose} or AlphaPose \cite{fang2022alphapose}. 

Building on LiCamPose~\cite{panLiCamPoseCombiningMultiView2025}, the SA-VR system~\cite{guoSportsAnalysisVR2024} augments the pipeline with multi-view tracking to support scene-wide sports analysis and VR visualization. It adopts a tracking-by-detection paradigm with two branches. First, a point cloud branch synchronizes and merges multi-view LiDAR scans, rasterizes the cloud into BEV representation, and processes it with PointPillars~\cite{lang2019pointpillars}. Second, an image branch processes synchronized multi-view images through a CNN, projecting them onto a shared BEV space.
These BEV-aligned features are jointly used to detect 3D bounding boxes, enabling consistent player localization across frames. Appearance (RGB) and geometric (LiDAR) features are then fused for multi-modal data association and temporal tracking. This system exemplifies how robust \hpes, as enabled by LiCamPose, can be scaled up and deployed in real-world applications such as sports analytics and immersive VR viewing.

Unlike HPERL \cite{furstHPERL3DHuman2021} and WS-HPE~\cite{zhengMultimodal3DHuman2022}, which depend on ground-truth 2D annotations, WS-Fusion~\cite{bauerWeaklySupervisedMultiModal2023}, FusionPose~\cite{congWeaklySupervised3D2023} and LiCamPose~\cite{panLiCamPoseCombiningMultiView2025} push weak supervision to the extreme, relying entirely on 2D pseudo-labels obtained from off-the-shelf models and geometric projection to bootstrap 3D learning. Notably, these off-the-shelf (e.g. ViT-Pose~\cite{vitpose}, AlphaPose \cite{fang2022alphapose}, OpenPose \cite{cao2019openpose}) models are trained in a \textit{supervised manner} using 2D ground-truth annotations.

\textbf{Temporal Reasoning.} One of the biggest challenges in LiDAR-based HPE is data sparsity. While spatial fusion helps, adding temporal coherence proves more powerful. FusionPose~\cite{congWeaklySupervised3D2023} tackles large-scale multi-person HPE with a temporal pipeline, which fuses RGB and LiDAR features using an image-to-point cross attention module and a GRU encoder for sequences of frames, thus extracting time-consistent joint features. The final pose prediction refines preliminary 3D joints using motion and consistency constraints. FusionPose showcases motion as supervision, with temporal consistency helping to learn robust representations even with sparse or noisy data.

\paragraph{Bridging Modalities via Fusion}\label{par:bridging_modality}

Combining 2D and 3D features is at the heart of multi-modal human pose estimation, as shown by all weakly supervised methods being multi-modal in Figure~\ref{fig:modality_vs_learning}. Most strategies vary in how early or late in the pipeline they merge the modalities. A few methods align features across modalities implicitly with learned attention. 

HPERL~\cite{furstHPERL3DHuman2021} fuses 2D and 3D features later in the pipeline. It encodes LiDAR scene point clouds in BEV maps and processes images with a standard convolutional backbone. The resulting feature maps for each modality undergo two steps. First, a predefined anchor grid is used to extract regions from maps via RoI Align, followed by cross-modal feature averaging. A Region Proposal Network then predicts objectness scores and refines regions. Secondly, the updated regions are RoI Aligned to re-extract feature maps, followed by cross-modal feature concatenation. The fused features are used to classify anchor poses and refine them using an LCRNet-inspired architecture.
HPERL operates at the object level, combining high-level representations extracted from corresponding spatial regions in each modality.

WS-HPE~\cite{zhengMultimodal3DHuman2022} performs low-level feature fusion by directly integrating 2D keypoint heatmaps with LiDAR points. First, a 2D pose-estimation network processes RGB images into keypoint heatmaps, which are Gaussian-smoothed and queried at the projected locations of 3D LiDAR points. This enriches geometric LiDAR points with texture information. WS-HPE extracts a K-dimensional vector for each point (K being the number of joint keypoints) representing the point's likelihood of being near each joint and concatenates these vectors with the original 3D coordinates. A PointNet-based network uses the complete (K+3)-d features for 3D pose regression and the auxiliary task of body segmentation. The early-stage fusion results in pointwise features that combine geometric structure with pose-specific semantic cues.

WS-Fusion~\cite{bauerWeaklySupervisedMultiModal2023} processes 2D and 3D features separately and performs a late fusion just before predicting the 3D keypoints. An independent branch processes LiDAR data into 3D joint predictions using a point network, while another branch lifts 2D keypoints obtained from a pretrained pose estimator into 3D. The final pose prediction combines the two 3D outputs through a dense layer that weights the contribution of each modality. This high-level fusion strategy allows the network to remain modular and flexible.

FusionPose~\cite{congWeaklySupervised3D2023} is the most dynamic and flexible approach to multi-modal feature combination, introducing cross attention to fuse image and point features in a \textit{calibration-free manner}. The architecture includes a dedicated Image-to-Point Attention Fusion module called IPAFusion, which learns correspondences between image-derived features and LiDAR point features, without relying on explicit projection. A PointNet ~\cite{qi2017pointnet} processes point clouds and  HRNet~\cite{hrnet} processes the images. Self-attention layers use the resulting features to capture global context and cross-attention between point queries and image keys/values produces fused descriptors for each point. This approach enables learning how and where to transfer information from image to point space, adaptively fusing local geometry with semantic content. A temporal module further refines the fusion output.

LiCamPose~\cite{panLiCamPoseCombiningMultiView2025} implements a volumetric-fusion architecture where features from both modalities are processed in parallel branches. RGB heatmaps (obtained from multi-view 2D pose detectors and then back-projected) and segmented human LiDAR point clouds are each voxelized and encoded using V2V-Net~\cite{v2v}. These volumetric features are then concatenated and passed through a fusion module to predict heatmaps of joint positions. By operating in a shared 3D voxel space, LiCamPose enables fusion even when the alignment between modalities is imperfect. The final joint positions are inferred using soft-argmax over the fused heatmap. This volumetric fusion strategy, while computationally expensive, proves robust in handling occlusions and scanning pattern-induced sparsity. LiCamPose fuses not only multi-modal data (RGB + LiDAR) but also multi-sensor LiDAR inputs, aggregating information from multiple views to construct a more complete volumetric representation of the scene. This multi-LiDAR fusion design proves particularly effective in scenarios where a single LiDAR sensor might miss parts of the human body due to self-occlusion or limited field-of-view.

\begin{figure}[t]
    \centering
    \includegraphics[width=\linewidth]{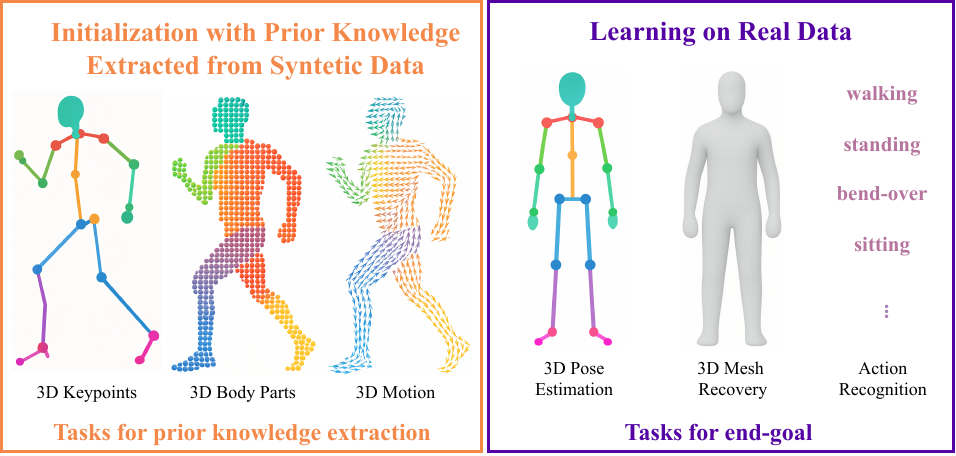}
    \caption{\textbf{Overview of synthetic pretraining} used in GC-KPL \cite{weng3DHumanKeypoints2023}, ReMP \cite{jangReMPReusableMotion2025}, and UniPVU-Human \cite{xuUnifiedFrameworkHumancentric2024}.
These approaches typically follow a two-stage training strategy. In the first stage, the models are pretrained on synthetic datasets to learn generalizable features through prior task learning. In the second stage, the pretrained weights are fine-tuned on real LiDAR datasets, with the objective of adapting the models to downstream tasks such as 3D HPE, 3D HMR, and action recognition.}
    \label{fig:pretraining}
\end{figure}

In summary, these methods differ not only in when they fuse 2D and 3D features, early-stage (WS-HPE~\cite{zhengMultimodal3DHuman2022}), middle-stage (HPERL~\cite{furstHPERL3DHuman2021}), or late-stage (WS-Fusion~\cite{bauerWeaklySupervisedMultiModal2023}, LiCamPose~\cite{panLiCamPoseCombiningMultiView2025}) but also in how explicitly they model the alignment. Some rely on precise geometric projection, others on object-level regions, and some like FusionPose~\cite{congWeaklySupervised3D2023} learn correspondences through attention without direct projection. Together, they represent a diverse set of strategies for integrating sparse depth with dense semantics in a unified learning pipeline.

\textbf{Neural architecture choices} vary widely from CNNs and PointNets to Transformers and attention-based modules, depending on where and how the modalities are fused. WS-HPE~\cite{zhengMultimodal3DHuman2022}, WS-Fusion~\cite{bauerWeaklySupervisedMultiModal2023} and FusionPose~\cite{congWeaklySupervised3D2023} adopt a PointNet~\cite{qi2017pointnet, ni2020pointnet++} backbone for LiDAR, but FusionPose integrates a cross-attention mechanism and a GRU temporal module to capture both semantic alignment and motion consistency. The relatively older HPERL relies on VGG-16 \cite{vgg16} encoders for feature extraction from images and LiDAR BEV-maps, fusing the results by RoI-alignment.
WS-Fusion adopts dual-branch networks that independently predict 3D pose from 3D-lifted 2D joints and LiDAR, with fusion at the final embedding level via a simple dense layer.

Collectively, these works highlight a few recurring themes for pseudo-labeling to generate reliable training signals: 3D to 2D projection, 2D to 3D lifting, and camera-LiDAR calibration.
Auxiliary tasks (e.g. segmentation) help reinforce learning in sparse supervision regimes. Pretrained 2D networks act as teachers, enabling 3D reasoning without costly labels. 
Transformers and attention-based architectures model long-range dependencies across modalities and time.

\subsection{Unsupervised Human Pose Estimation}\label{subsec:unsup_hpe}

In contrast to supervised and weakly supervised approaches, unsupervised learning 
remains relatively under-explored, with only GC-KPL~\cite{weng3DHumanKeypoints2023} explicitly addressing it by carefully designing unsupervised losses.

GC-KPL~\cite{weng3DHumanKeypoints2023} adopts a two-stage pipeline, first pretraining on supervised synthetic data, then refining by self-supervision on unlabeled real-world LiDAR sequences (Figure \ref{fig:pretraining}), which learns human-body structure with geometric priors rather than annotations. A flow loss encourages temporal consistency by ensuring that the surface points of a limb preserve radial and axial coordinates across frames. A point-to-limb loss minimizes the distance between each point in a body part segment and its corresponding predicted limb axis, weighted by the segmentation confidence. A symmetry loss further regularizes the limb predictions by requiring surface points with the same axial coordinates (along the limb) to have similar radial distances. Finally, a joint-to-part loss encourages predicted keypoints to remain near the geometric center of their associated surface points, even though real joint positions may not lie exactly at the center, an unconventional constraint that improves robustness in the absence of labels. A weighted combination of these losses forms the learning objective, which guides the network to develop a meaningful, anatomically consistent internal representation of human pose using only unlabeled data.

For downstream evaluation, GC-KPL~\cite{weng3DHumanKeypoints2023} demonstrates strong performance in both unsupervised settings and data-efficient learning scenarios. For unsupervised evaluation on HPE, they report MPJPE on matched keypoints after establishing correspondences between predicted and annotated keypoints through Hungarian-Matching.
To demonstrate the effectiveness of its pretrained network in data-efficient training, GC-KPL fine-tunes its pretrained model with 1\% or 10\% of labeled data, showing that self-supervised pretraining significantly improves performance, especially in low-data regimes.


\section{3D Human Mesh Recovery from LiDAR}\label{sec:hmr}
\subsection{LiDAR-only Human Mesh Recovery}\label{subsec:lidar_only_hmr}

Despite challenges posed by sparse, noisy, and occluded point clouds, single-modality LiDAR-based human mesh recovery has shown promising results.

\paragraph{Sparse-to-Dense Reconstruction}\label{par:sparse_to_dense_recon}
A foundational challenge in LiDAR-based mesh recovery is transforming sparse and incomplete point clouds into full-body dense 3D meshes. LiDAR-HMR~\cite{fanLiDARHMR3DHuman2025} proposes a two-stage cascaded architecture that incrementally reconstructs dense meshes from raw-LiDAR sparse pose estimation. Pose Regression Network (PRN) leverages a PointTransformer-v2~\cite{wu2022pointv2} backbone combined with a voting module to estimate a sparse set of human keypoints, which are refined and sent to a Mesh Reconstruction Network, using Graphormer~\cite{lin2021mesh} to iteratively densify the mesh. Intermediate supervision with down-sampled ground-truth meshes enhances stability and quality.

\begin{figure}[]
    \centering
    \includegraphics[width=1.0\linewidth]{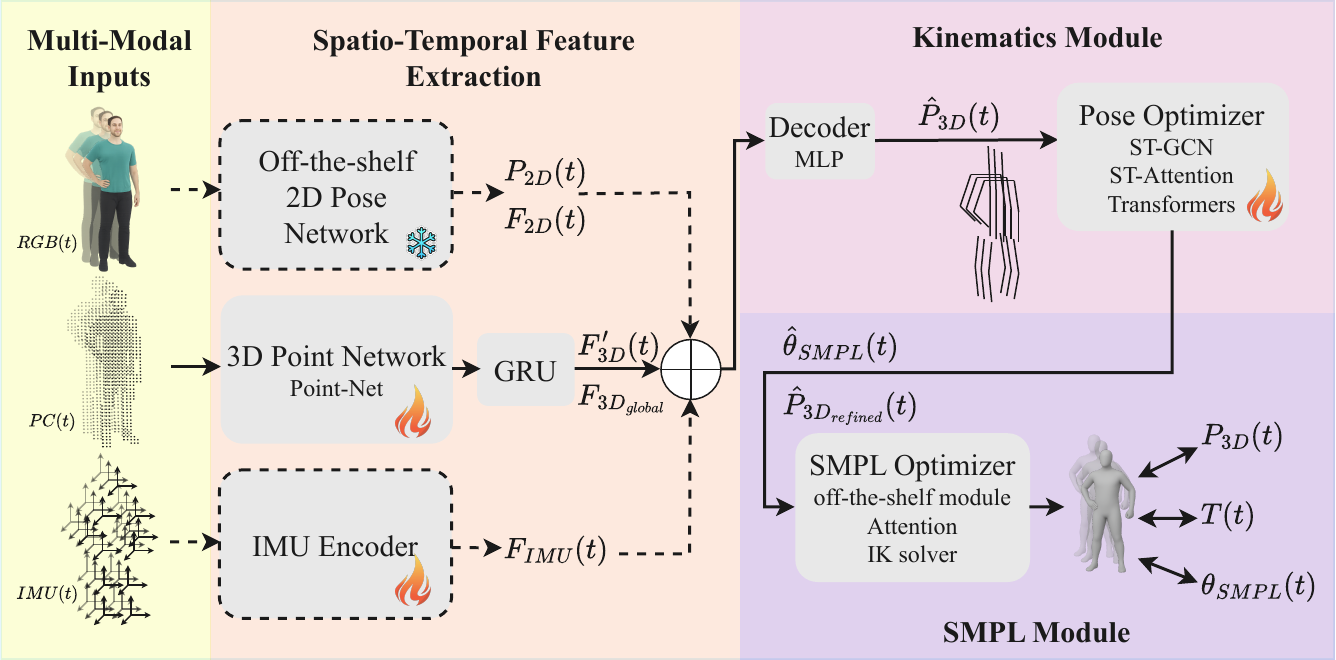}
    \caption{\textbf{Common Training Pipeline of a LiDAR-based Mesh Recovery.} The HMR methods typically leverage temporal information and, in many cases, incorporate additional modalities such as images or IMUs. The pipeline generally begins with the extraction of spatio-temporal features from the input data. For LiDAR point clouds, PointNet and variants~\cite{qi2017pointnet,ni2020pointnet++} are widely used to obtain spatial representations $F_{3D}(t)$, while bi-GRUs capture temporal dependencies, producing enhanced features $F'_{3D}(t)$ and global descriptors $F_{3D_{global}}$. When multiple modalities are available, a fusion step combines their representations before proceeding to preliminary 3D keypoint estimation. These keypoints, $\hat{P}_{3D}(t)$, are predicted via an MLP decoder applied to the spatio-temporal features. The estimates are then refined by a Pose Optimizer, which employs either ST-GCNs~\cite{stgcn} or spatio-temporal attention mechanisms to produce refined predictions $\hat{P}_{3D_{refined}(t)}$. Finally, the refined keypoints are mapped to SMPL parameters through a dedicated module. This module can vary across approaches, ranging from off-the-shelf SMPL regression networks to attention-based designs or inverse-kinematics solvers, ultimately yielding the reconstructed human mesh.}
    \label{fig:hmr_pipeline}
\end{figure}

\paragraph{Spatio-temporal Modeling for Robustness}\label{par:st_modeling_robustness}

To push LiDAR-based HMR closer to real-world deployment, recent works have introduced spatio-temporal modeling to leverage context reasoning and to tolerate occlusion and human-object interaction. These methods employ both local and global features over time, incorporate contextual and temporal cues, and reconstruct meshes from noisy and incomplete raw LiDAR data. The common pipeline is shown in Figure~\ref{fig:hmr_pipeline}.

NE-3D-HPE~\cite{zhangNeighborhoodEnhanced3DHuman2024} explicitly models context by constructing two spatial structures: 3D Scanning Neighbors (3SN) and 3D Background Neighbors (3BN). For each segmented human point cloud, 3BN collects surrounding points within a 2-meter Euclidean distance from the human center, while 3SN converts the human point cloud to polar coordinates and selects points within a given angular window and radial range. 
The full input consists of a temporal sequence of human points and their associated 3BN and 3SN contexts, expressed in local coordinates. Features are extracted independently for each modality using three PointNet++~\cite{ni2020pointnet++} encoders and then fused by a Coherence-Fuse module with self-attention on 3SN, cross-attention across modalities, and global self-attention. The fused features are used to regress 3D keypoints, which, along with a global descriptor, allow a Spatio-Temporal Conditional Graph Network (ST-CGN)~\cite{stgcn} to predict joint rotations. The network is trained end-to-end using intermediate losses on keypoints, SMPL regressed joints, and body pose.  

LiDARCap~\cite{liLiDARCapLongrangeMarkerless2022} approaches motion capture through temporal graph modeling. Its three-stage architecture begins with a Temporal Encoder that processes sequences of segmented human LiDAR frames using PointNet++~\cite{ni2020pointnet++} to extract per-frame features, followed by a bi-directional GRU (bi-GRU) that fuses temporal information. The encoded sequence is decoded using an MLP to produce sparse keypoint predictions. These predictions and the temporal embeddings are fed to an Inverse Kinematics Solver, which constructs a graph to capture intra- and inter-frame keypoint relations. An ST-GCN~\cite{stgcn} regresses joint rotations, which are subsequently passed on to an off-the-shelf SMPL model to regress the mesh. Loss functions are again applied on intermediate keypoints, SMPL regressed joints and body pose.

In realistic outdoor settings, occlusion and interactions with objects degrade LiDAR point clouds and challenge existing methods. To improve robustness under occlusion and human-object interaction, LiDARCapV2~\cite{zhangLiDARCapV23DHuman2024} builds on LiDARCap~\cite{liLiDARCapLongrangeMarkerless2022} by incorporating a noise-resilient noise-augmented segmentation (AgNoiseSegment) module. This component injects progressively decaying Gaussian noise during training to help the segmentation network distinguish humans from nearby objects. A bi-GRU skeleton extractor and ST-GCN-based inverse kinematics module refine pose estimation under partial visibility. ST-GCN is followed by an SMPL module, as in LiDARCap~\cite{liLiDARCapLongrangeMarkerless2022} and NE-3D-HPE~\cite{zhangNeighborhoodEnhanced3DHuman2024}, using the same loss signals. 

LiveHPS~\cite{renLiveHPSLiDARbasedScenelevel2024}  targets practical deployment in unconstrained environments by combining motion optimization and scene-level SMPL estimation in a lightweight pipeline. The method starts with a PointNet-GRU-based (inspired by LiP~\cite{renLiDARaidInertialPoser2023}) body tracker trained to predict human joints from unstructured point clouds. To inject regularity into LiDAR scenes tainted by occlusion and a varying number of points across frames, it introduces Vertex-guided Adaptive Distillation: ground-truth mesh vertices are aligned with LiDAR points using k-NN matching, and vertex features are distilled into point features via KL divergence and joint MSE loss.
A Consecutive Pose Optimizer is introduced to improve temporal and spatial coherence. For each frame, 
every joint learns to attend to its neighboring joints in the skeleton as well as itself across a temporal window. The final stage, a multi-head attention-based SMPL solver, regresses shape parameters, body pose, and global translation. The entire model is supervised by MSE losses on joints, vertices, SMPL parameters, along with a unidirectional Chamfer distance from point clouds to mesh vertices. 

LiveHPS++~\cite{renLiveHPSRobustCoherent2024} extends LiveHPS~\cite{renLiveHPSLiDARbasedScenelevel2024} for motion resilience under extreme noise and dynamic interactions. It introduces a Trajectory-guided Body Tracker, where human trajectory and human point cloud sequences are jointly encoded. A frozen pretrained feature encoder from ground-truth mesh vertices and trajectories distills structured features into the main model using Vertex-Trajectory-guided Adaptive Distillation. 
To prevent kinematic inconsistency due to noisy parent joint predictions, a noise-insensitive model predicts the velocity vectors for the predicted keypoints and global translation features using cross-attention between them and raw points.
Additionally, a Kinematic-aware Pose Optimizer generates multiple joint candidates from prior time steps and refines them. {The final SMPL solver stage is identical to LiveHPS.}

\paragraph{Distillation-based Latent Prior Learning}\label{par:ws_latent_prior}

Unlike the aforementioned methods, which rely on dense supervision and real-world annotated LiDAR scans, ReMP~\cite{jangReMPReusableMotion2025} introduces weak supervision based on a reusable motion prior trained across diverse motion capture sources. It leverages a transformer-based variational autoencoder to learn a spatio-temporal latent space of human motion sequences using SMPL parameters from the AMASS \cite{AMASS:ICCV:2019} dataset. This motion prior, once trained, remains frozen and generalizes across modalities, including LiDAR. 
In a downstream HMR task using LiDAR data, ReMP maps raw point cloud sequences to a reusable latent motion space through a distillation pipeline. A PointNet-based encoder extracts high-level spatial features from each LiDAR input frame. A transformer-based distillation module aligns the features with the latent distribution learned by the frozen transformer-VAE motion prior. This alignment is trained using synthetic LiDAR data generated from motion capture datasets like CMU~\cite{cmuDataset}, where LiDAR-style point clouds are rendered from known SMPL mesh sequences (Figure~\ref{fig:synt-data-gen}). Because the synthetic data comes with ground-truth pose, translation, and shape parameters, it enables self-annotated supervised training of the point encoder and latent mapper. During training, the model is optimized using reconstruction losses on body pose, angular velocity, root translation and speed as well as SMPL vertices and joints, enforcing consistency with the motion prior's latent space using KL divergence. Once trained on synthetic data, the model can be directly tested on real LiDAR datasets, showcasing ReMP’s strong generalization capabilities from synthetic to real data. 
We note that ReMP can also accept IMU or depth maps and distill learned priors into these modalities, demonstrating that SMPL-based priors can be easily transferred to different modalities, giving the flexibility to work with IMU or depth alone at inference.

\subsection{Fusing LiDAR and Other Modalities for Human Mesh Recovery}\label{subsec:fusing_lidar_and_other_modalities_hmr}

To address long-standing challenges in motion capture, including occlusions, calibration dependencies, and global localization in large-scale environments, integrating LiDAR with other sensor modalities such as RGB cameras and IMUs is now popular. These methods can be broadly categorized into three methodological paradigms: calibration-free fusion in dynamic scenes, hierarchical sensor fusion with inertial signals, and scene-aware optimization pipelines for large-scale annotation.

\paragraph{Calibration-free Fusion}\label{par:calib_free_fusion} Calibration-free systems utilize RGB and LiDAR for HMR in open and unconstrained environments. 

FreeCap~\cite{xueFreeCapHybridCalibrationFree2025} begins with a Pose-Aware Cross-Sensor Matching algorithm to associate human instances between LiDAR point clouds and multi-view images. It uses 2D SMPL body pose parameters and keypoints from WHAM~\cite{wham:cvpr:2024} and RTMPose~\cite{jiang2023rtmpose}, together with 3D counterparts from LiveHPS~\cite{renLiveHPSLiDARbasedScenelevel2024}. Matching is initialized via Hungarian Matching using cosine similarity of local 2D and 3D poses, then refined based on the 3D-to-2D reprojection error of local and global body pose as the objective function using Perspective-n-Point to obtain camera pose. 
The aligned sequences are passed to the Sensor-Expandable Pose Optimizer, where MLPs encode 2D motion, image, 3D motion, and point cloud features. Self-attention models temporal interactions within each modality, while bidirectional cross-attention captures cross-modal relations. Finally, enriched 2D/3D features are decoded into 2D/3D joints, and SMPL parameters are estimated using a temporal-attention network.

SMPLify-3D~\cite{dumontImprovingImagebased3D} complements FreeCap~\cite{xueFreeCapHybridCalibrationFree2025} by a calibration-free framework that refines image-based predictions, such as those from CLIFF~\cite{li2022cliff}, using LiDAR cues.
It applies a visibility filter inspired by the backface culling algorithm and aligns visible mesh faces with sparse LIDAR points via Iterative Closest Point. Later, it further refines alignment through optimization guided by priors and visibility-aware losses. This design enables robust performance under occlusion, improving over image-only approaches.

\paragraph{Fusion from LiDAR and IMU/RGB}\label{par:fusion_lidar_imu} These methods explore hierarchical fusion of those modalities, enabling robust motion capture across time and space, particularly in long-range or occluded scenarios. LiDAR-aid Inertial Poser (LiP)~\cite{renLiDARaidInertialPoser2023} proposes a lightweight and scalable motion capture pipeline using a single LiDAR sensor and four sparse IMUs. The method starts with a global temporal pose-prior distillation step, where point clouds over time are encoded using PointNet~\cite{qi2017pointnet} and bi-GRU to infer coarse joint positions and root orientation. These are further refined using a hierarchical inverse kinematics network that integrates IMU readings to predict accurate joint rotations. To solve the drifting and localization limitations of IMUs, LiP introduces a trajectory-guided translation correction mechanism, predicting the discrepancy between the estimated root and the global position of the human instance using temporal LiDAR sequences. LiP is supported by the introduction of the LIPD dataset, a long-range LiDAR-IMU dataset. 

HmPEAR~\cite{linHmPEARDatasetHuman2024} presents a large-scale multi-modal dataset for both 3D Human Pose Estimation and Human Action Recognition. To demonstrate the utility of this dataset, the authors introduce PEAR-Proj, a model that jointly learns both pose and action representations from a sequence of RGB images and point clouds. PEAR-Proj uses HRNet~\cite{hrnet} and PointNet++~\cite{ni2020pointnet++} for multi-modal RGB and point cloud features, and a temporal bi-GRU encoder that operates on the multi-modal features to extract both frame-wise and sequence-wise features. 
The frame-wise features are mapped into 3D keypoint predictions, then an ST-GCN~\cite{stgcn} uses the concatenation of these features and keypoints to produce per-frame action labels. On the other hand, sequence-wise features are decoded into shape and, combined with the 3D predicted keypoints, are fed into HybrIK~\cite{li2021hybrik} to obtain SMPL meshes. Training alternates between pose- and action-labeled clips, optimizing pose and cross-entropy losses jointly. 
They show that joint training enhances both pose and action recognition accuracy. 

\begin{figure}[]
    \centering
    \includegraphics[width=1.0\linewidth]{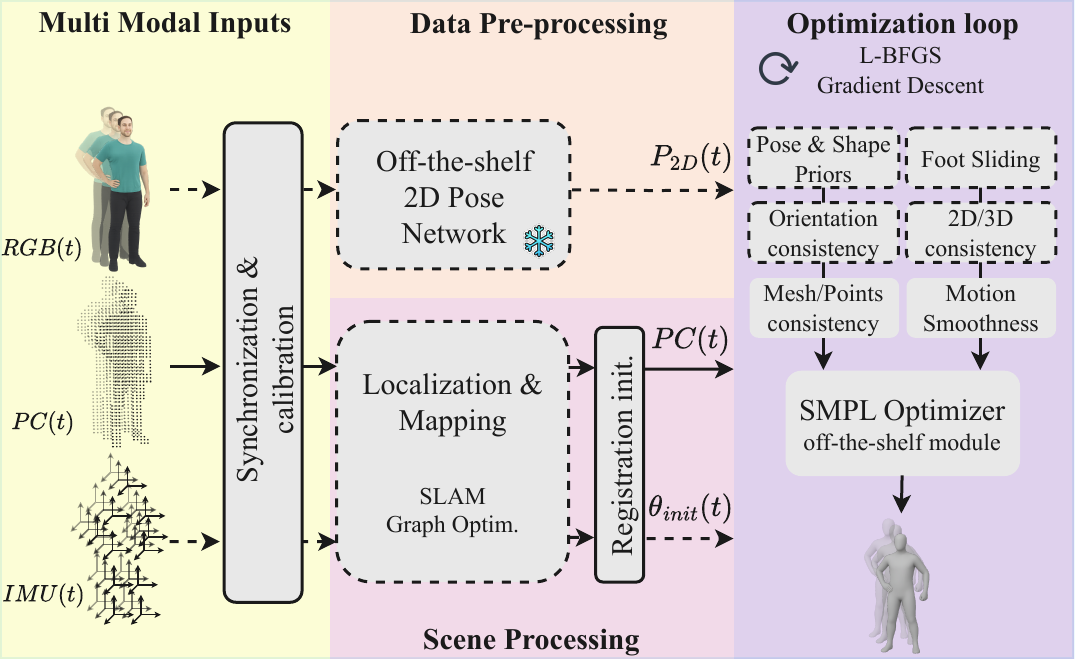}  
    \caption{\textbf{The common multi-model optimization pipeline for HMR methods.} Optimization-based HMR is mostly used in dataset papers for their annotation pipeline, which mainly relies on a temporal optimization loop. The pipeline begins by capturing multiple modalities through well-synchronized and calibrated sensors. These inputs are then processed according to their type. For instance, if image data is available, a 2D pose sequence $P_{2D}(t)$ is extracted using an off-the-shelf pose estimator. When IMU data is present, a MoCap-based solution provides an initial estimate of the SMPL mesh parameters $\theta_{init}(t)$. In scenarios where scene-level mesh reconstruction is required, LiDAR SLAM or graph optimization techniques are applied to jointly process LiDAR and IMU inputs. Finally, an optimization loop refines the SMPL parameters by integrating cues from the 2D poses, LiDAR point clouds, and the initial SMPL estimates. This process is guided by task-specific losses and priors, and typically relies on optimization methods such as gradient descent or L-BFGS \cite{lbfgs}.}
    \label{fig:optim_hmr_pipeline}
\end{figure}

\paragraph{Multi-modal Optimization Pipelines}\label{par:multi-modal_optim} These pipelines are scene-aware and usually aimed at dataset generation or fine-grained annotation. They often rely on a combination of LiDAR, RGB, and IMU data to ensure global consistency, smooth motion, and plausible human-scene interaction. Their common pipeline is shown in Figure~\ref{fig:optim_hmr_pipeline}.

SLOPER4D~\cite{daiSLOPER4DSceneAwareDataset2023} and HSC4D~\cite{daiHSC4DHumancentered4D2022}  share a common optimization pipeline. 
HSC4D captures both human motion and 3D scenes using wearable LiDAR and IMUs. It is a data annotation framework for the proposed dataset of the same name. It builds a complete 3D scene map via LiDAR-based SLAM and fuses IMU and LiDAR trajectories through a graph-based optimization process. The nodes of the graph represent IMU poses, while edges encode relative transformations. After aligning these trajectories, a joint optimization procedure iteratively refines the entire motion sequence, guided by physics-inspired constraints such as foot-ground contact, orientation smoothness, and reduced foot sliding. The final result includes temporally coherent SMPL-mesh sequences aligned to a consistent 3D scene, producing high-quality annotations of human-scene interactions.

\begin{figure*}[t!]
\centering
\includegraphics[width=1.0\linewidth]{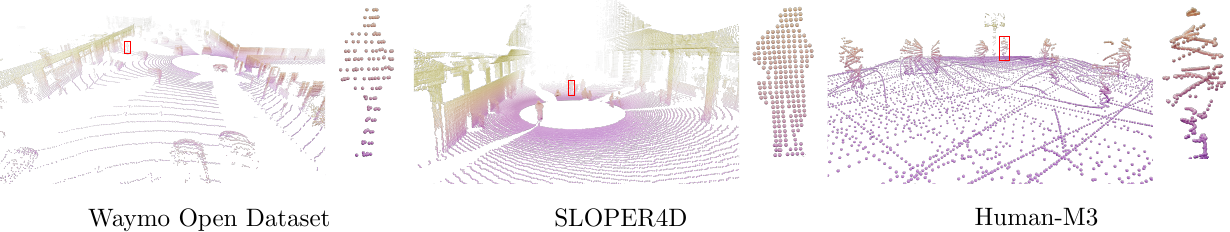}
\caption{\textbf{Comparison of LiDAR point cloud datasets for 3D HPE and HMR}. For each dataset, we show an example scene and a person instance, which is depicted with a red bounding box in the scene.
The figure shows differences in scanning patterns across datasets. For example, while both WOD~\cite{sunScalabilityPerceptionAutonomous2020} and SLOPER4D~\cite{daiSLOPER4DSceneAwareDataset2023} are captured using RMB pattern LiDARs, SLOPER4D achieves a denser scene due to its sensor having twice as many beams. In contrast, Human-M3~\cite{fanHumanM3MultiviewMultimodal2023} produces the sparsest point clouds, as its NRS pattern requires longer integration times for higher density. Self-occlusion is visible in WOD (e.g., cars) and SLOPER4D (e.g., humans), where only surfaces facing the sensor are captured. Human-M3, however, shows no self-occlusion because it fuses four LiDAR views in a post-processing step, covering all sides of the scene. Note that the humans shown in each example are at different distances from their respective sensors, which partly affects point density. This figure should be used primarily to compare scanning patterns and general scene characteristics; for fair comparisons of human point cloud density, refer to Section~\ref{sec:datasets_stats} and particularly Table~\ref{tab:dataset_intrinsic_stats}.} 
\label{fig:datasets_figures} 
\end{figure*}

SLOPER4D~\cite{daiSLOPER4DSceneAwareDataset2023} extends HSC4D~\cite{daiHSC4DHumancentered4D2022} data annotation framework to annotate their newly proposed SLOPER4D dataset. To this end, they additionally leverage RGB data. 
Images are processed using Detectron~\cite{wu2019detectron2} and DeepSort~\cite{wojke2017simple} for human detection, 2D keypoints and tracking, while a fused LiDAR scene is built using VDB-Fusion \cite{vizzoVDBFusionFlexibleEfficient2022}, ensuring removal of moving artifacts.
MoCap estimates of SMPL parameters are then refined using similar losses to HSC4D with additional mesh-to-points alignment and camera reprojection losses. 
This mesh-to-points alignment enables visibility-aware registration, leading to improved accuracy under occlusion.

CIMI4D~\cite{yanCIMI4DLargeMultimodal2023} is another extension of HSC4D~\cite{daiHSC4DHumancentered4D2022} that captures human subjects engaging in a climbing activity. Unlike SLAM-based pipelines, the static nature of the background enables straightforward scene mesh reconstruction and point cloud sequence registration. Similar to SLOPER4D~\cite{daiSLOPER4DSceneAwareDataset2023}, it further extends the modality by incorporating RGB data. Moreover, CIMI4D synchronizes modalities at the frame level, in contrast to the sequence-level synchronization used in HSC4D and SLOPER4D.

Inspired by the annotation pipeline of SLOPER4D~\cite{daiSLOPER4DSceneAwareDataset2023}, 
SMPLify-3D~\cite{dumontImprovingImagebased3D} builds on image-based mesh recovery models like CLIFF~\cite{li2022cliff} by refining predictions through LiDAR-informed optimization. 
The framework first uses CLIFF to generate initial SMPL mesh predictions. Inspired by the backface culling algorithm, it then applies a visibility filter
guided by available 2D confidence scores and SMPL body-part-to-face mapping. This ensures that invisible mesh faces are not considered during optimization. It employs Iterative Closest Point to align visible mesh faces with sparse LiDAR points. This alignment is then refined through an extended optimization process that deforms the SMPL mesh by optimizing its parameters to better fit the human point cloud. It incorporates several loss terms, including joint reprojection, pose and shape priors, unnatural joint bending penalties, and a visibility-aware Chamfer distance for 3D alignment. 

Another dataset-focused work is HmPEAR~\cite{linHmPEARDatasetHuman2024}, where the dataset is annotated with a pipeline inspired by SLOPER4D~\cite{daiSLOPER4DSceneAwareDataset2023}, incorporating similar loss terms.

Human-M3~\cite{fanHumanM3MultiviewMultimodal2023} introduces a multi-view, multi-LiDAR dataset and proposes a dedicated HMR-based annotation pipeline to generate 3D pose and mesh labels. The process begins by manually matching LiDAR and camera to compute accurate extrinsic calibration parameters between modalities. Using the calibrated setup, pedestrians are detected directly in the LiDAR point clouds via the PointPillars~\cite{lang2019pointpillars} detector and tracked over time using the AB3DMOT~\cite{weng2020ab3dmot} multi-object tracking framework. These 3D detections are then projected into corresponding RGB images. Within the projected image regions, ViTPose~\cite{vitpose} is applied to estimate high-resolution 2D human keypoints, forming the basis for the subsequent optimization stage.
To recover full-body pose and shape, Human-M3 employs a temporal optimization strategy that predicts and refines SMPL parameters for each individual across a sequence. This optimization aims to jointly solve for the root translation, body shape, and pose parameters by minimizing a composite objective. The loss function includes: (1) a 2D keypoint reprojection loss that enforces alignment between the projected 3D joints and detected 2D keypoints across all views; (2) a Chamfer distance loss that measures geometric consistency between the mesh surface and observed human point cloud segments; (3) shape and pose priors from VPoser~\cite{pavlakos2019expressive} to regularize anthropomorphically valid body configurations; and (4) a temporal smoothness term that penalizes abrupt pose changes across frames. The complete objective is optimized using the L-BFGS algorithm \cite{lbfgs}, ensuring temporally coherent and geometrically grounded 3D annotations. To ensure annotation quality, the results of the automatic annotation pipeline are manually reviewed and corrected.

All these methods follow a common pipeline: they begin by integrating and synchronizing multiple input modalities, then refine predictions through an optimization loop that leverages tailored losses, priors and scene-specific constraints. This process yields more accurate and robust human mesh reconstruction.

\section{Datasets}\label{sec:datasets}

In this section, we provide detailed descriptions of the Waymo Open Dataset~\cite{sunScalabilityPerceptionAutonomous2020} (Section \ref{sec:datasets_wod}), SLOPER4D~\cite{daiSLOPER4DSceneAwareDataset2023} (Section \ref{sec:datasets_sloper}), and Human-M3~\cite{fanHumanM3MultiviewMultimodal2023} (Section \ref{sec:datasets_humanm3}) datasets. These datasets were selected for their popularity, relevance to autonomous and outdoor scenarios, high-quality annotations, and inclusion of diverse LiDAR scanning patterns. Figure~\ref{fig:datasets_figures} presents a visual comparison at both the scene level and the human instance level, highlighting differences in the captured environments as well as LiDAR characteristics, including point density and scanning patterns.
Next, in Section \ref{sec:datasets_stats}, we present a dedicated comparative study in which we extract and analyze dataset characteristics, intrinsic properties, and diversity.  
Finally, Section \ref{sec:datasets_others} provides brief overviews of other publicly available LiDAR point cloud datasets. As shown in Table~\ref{tab:rw}, although numerous datasets have been introduced in the literature, not all of them are publicly accessible. 

For more information about the datasets, please refer to our GitHub page \footnote{\fontsize{6pt}{8pt}\selectfont\url{https://github.com/valeoai/3D-Human-Pose-Shape-Estimation-from-LiDAR}}.

\subsection{Waymo Open Dataset} %
\label{sec:datasets_wod}

The Waymo Open Dataset (WOD)~\cite{sunScalabilityPerceptionAutonomous2020} is a large-scale resource for autonomous driving research, comprising two main components: the Motion dataset and the Perception dataset, with this survey focusing on the latter. The Perception dataset offers rich multimodal annotations, supporting a wide range of 2D and 3D computer vision tasks, including human-centric applications such as 2D/3D human pose estimation and segmentation.

\textbf{Data Collection} 
\begin{table}[ht]
\centering
\caption{Extrinsic characteristics of lidar point cloud datasets.}
\resizebox{\linewidth}{!}{%
\begin{tabular}{llll}
\toprule
Datasets & WOD \cite{sunScalabilityPerceptionAutonomous2020} & SLOPER4D \cite{daiSLOPER4DSceneAwareDataset2023} & Human-M3 \cite{fanHumanM3MultiviewMultimodal2023}\\
\midrule
Area coverage ($m^2$) & 76M & 2-13k & \cellcolor{mh6!40}111.5k \\
Sequence length (s) & 20 & \cellcolor{mh6!40}102-441 & \cellcolor{mh6!40}12-45 \\
\# scenes & 998 & 6 & 4 \\
\# subjects & 23.6k & 12 & \cellcolor{mh6!40}237 \\
\# 3D human instances & \cellcolor{mh6!40}9.9k & \cellcolor{mh6!40}33k & \cellcolor{mh6!40}89k \\
\# LiDAR frames & 230k & \cellcolor{mh6!40}42.3k & \cellcolor{mh6!40}12.2k \\

\bottomrule 
\end{tabular}%
}
\label{tab:dataset_extrinsic_stats}

\begin{tablenotes} 
    \item  
    \textcolor{mh6}{Computed by us} and no color indicates information reported in the original paper. For SLOPER4D, only 6 of the 15 sequences were publicly released; accordingly, our statistics are based solely on these, resulting in 42.3k lidar frames rather than the 100k reported in the original paper.
\end{tablenotes}
\end{table}
The Perception dataset was collected using LiDAR sensors and high-resolution RGB cameras mounted on Waymo vehicles. Each vehicle is equipped with five in-house LiDAR units: one mid-range roof sensor (70 meters range, 25.2\textdegree $\times$ 120\textdegree\xspace FOV) and four short-range sensors at the front, rear, and sides (20 meters range, 25.2\textdegree $\times$ 20\textdegree\xspace FOV). Only the first two returns from each LiDAR beam are retained to ensure quality and consistency.

In addition to LiDAR, five RGB cameras capture images in the front, front-left, front-right, side-left, and side-right directions. Stored in JPEG format, these images include calibration data for LiDAR–camera fusion. All sensors operate at 10 Hz.

Released in 2019 and updated through April 2024~\cite{sunScalabilityPerceptionAutonomous2020}, the dataset encompasses day and night recordings from San Francisco, Phoenix, and Mountain View, providing geographically diverse urban and suburban driving scenarios.

\textbf{Data Format and Structure} 
The dataset is organized into folders by annotation type, including human keypoints, bounding boxes, and camera labels. Annotations are stored in Apache Parquet format for efficient handling of large-scale data. The data is divided into 20-second sequences sampled at 10 Hz, supporting synchronized multi-sensor processing and temporal analysis.

\subsection{SLOPER4D}
\label{sec:datasets_sloper}

SLOPER4D~\cite{daiSLOPER4DSceneAwareDataset2023} is a large-scale dataset designed for 3D human pose estimation in urban environments. It is the first to provide calibrated and synchronized multi-modal data, combining IMUs, LiDAR point clouds, and RGB images for each subject. The dataset includes rich 3D annotations such as SMPL parametric body models, precise subject locations in a global coordinate system, and a complete 3D mesh of the surrounding scene. 
Owing to its diverse sensory inputs and detailed annotations, SLOPER4D supports a wide range of human-centric 2D and 3D perception tasks, including human tracking, segmentation, and pose estimation.

\textbf{Data Collection} 
Data was collected using a mobile setup in which an operator, equipped with a head-mounted LiDAR-camera system, follows a subject performing everyday actions. The subject wears multiple IMUs for precise body motion capture. The LiDAR is an Ouster OS-1 mid-range model (360\textdegree $\times$ 45\textdegree\xspace FOV) mounted at a 45$^{\circ}$ angle to 
optimally capture the performer’s full body in motion. All sensor modalities are synchronized and subsampled to the LiDAR’s 20 Hz frame rate, ensuring high-quality spatial and temporal alignment.

Released in 2023, the dataset was collected in an unspecified Chinese city. The urban setting introduces real-world challenges such as occlusions, dynamic backgrounds, and varied lighting, making it highly valuable for 3D HPE and HRM research.

\textbf{Data Format and Structure} 
SLOPER4D is organized into folders, each containing a complete sequence with LiDAR point clouds (PCD format), tracking trajectories, raw trajectory data, MoCap recordings (BVH format), and a video of the sequence. Each folder also includes a JSON metadata file with contextual and calibration details, facilitating reproducibility and integration into machine learning pipelines.

\subsection{Human-M3}
\label{sec:datasets_humanm3}
Human-M3~\cite{fanHumanM3MultiviewMultimodal2023} is an outdoor, multi-modal, multi-person, and multi-view dataset designed as a benchmark for 3D HPE and HMR. 
It captures multi-person interaction scenes using four diagonally opposed camera–LiDAR pairs, providing fused and post-processed scene point clouds (from all four LiDARs) alongside their corresponding camera views. 
The dataset includes sparse 2D keypoints and pseudo-3D SMPL annotations generated through an optimization process similar to SMPLify-X \cite{pavlakos2019expressive}. 
Unlike the vehicle-centric WOD \cite{sunScalabilityPerceptionAutonomous2020} or the human-centric SLOPER4D \cite{daiSLOPER4DSceneAwareDataset2023}, Human-M3 \cite{fanHumanM3MultiviewMultimodal2023} employs a fixed sensor setup for the entire capture, enabling participants to move and interact freely within the scene. This design offers rich human-human and human-scene interactions, as well as a distinct scanning pattern that presents unique challenges for evaluating HPE/HMR algorithms.

\textbf{Data Collection} 
Data collection follows the method in~\cite{zhangFlexibleMultiviewMultimodal2022} using four camera-LiDAR units, with each camera mounted directly above its corresponding LiDAR to align their fields of view. Point clouds are captured from four diagonally opposed angles using Livox MID-100~\cite{livoxtechMid40Lidar} sensors (100\textdegree $\times$ 40\textdegree\xspace FOV) operating with a NRS pattern. All sensor streams are synchronized and sampled at 10 Hz, and the test set is manually annotated to ensure reliability.

\begin{table}[ht]
\centering

\caption{Acquisition-related characteristics of lidar point cloud datasets.} 
\resizebox{\linewidth}{!}{%
\begin{tabular}{llll}
\toprule
Datasets & WOD \cite{sunScalabilityPerceptionAutonomous2020} & SLOPER4D \cite{daiSLOPER4DSceneAwareDataset2023} & Human-M3 \cite{fanHumanM3MultiviewMultimodal2023} \\
\midrule
\# beams & \cellcolor{mh6!40}64 & \cellcolor{mh1!40}128 & 3$\dagger$ \\
PC resolution & \cellcolor{mh6!40}169600 & 131072 & \cellcolor{mh6!40}80928 \\
Range (m) & 20/75 & \cellcolor{mh1!40}90 & \cellcolor{mh1!40}90 \\
Framerate (Hz) & 10 & 20 & 10 \\
FOV (hfov $\times$ vfov) & 25.2\textdegree $\times$ 120/20\textdegree & \cellcolor{mh1!40}360\textdegree $\times$ 42.2\textdegree & \cellcolor{mh1!40}98.4\textdegree $\times$ 38.4\textdegree \\
\bottomrule 
\end{tabular}%
}
\label{tab:aquisition_stats}
\begin{tablenotes} 
    \item  
\textcolor{mh6}{Computed by us}, \textcolor{mh1}{from datasheets/specifications}, and no color indicates information reported in the original paper. 
$\dagger$ Sensor uses an NRS pattern with three laser beams, scanning at 300,000 points/s. We report the Range @10\% reflectivity for SLOPER4D and Human-M3. For WOD, we report the restricted range of their in-house sensors.
\end{tablenotes}
\end{table}

Based on the setup in~\cite{zhangFlexibleMultiviewMultimodal2022}, we infer that data collection took place in 2022 in a Chinese city, with annotations released in 2024. The dataset comprises four scenes: two tracking players on a basketball court, one at a major urban intersection, and one in a plaza.

\textbf{Data Format and Structure}
The data is organized into folders first by split (90\% training, 10\% testing) and then by sequence. Each sequence folder contains point clouds in PCD format for each frame, along with corresponding RGB images, camera calibration files, 3D sparse keypoint annotations, and SMPL labels, all provided in JSON format.

\subsection{Comparative Analysis of Dataset Properties}
\label{sec:datasets_stats}
In this section, we present a detailed analysis of the datasets along with a range of statistical summaries. We begin by examining the extrinsic characteristics of each dataset, including the \textit{area coverage}, which refers to the physical coverage during data capture; the \textit{sequence length} in seconds, reported either as a single value when the duration is fixed or as a range between the shortest and longest sequences recorded; the \textit{number of scenes}, representing the total captured sequences and, in some cases, implying distinct locations; the \textit{number of subjects}, corresponding to the unique tracked performers, actors, or individuals across all sequences; the \textit{number of 3D human instances}, which indicates the total labeled 3D human annotations; and the total number of \textit{LiDAR frames}. These extrinsic dataset properties are summarized in Table~\ref{tab:dataset_extrinsic_stats}.
We also report acquisition-related statistics, including the \textit{point cloud resolution} in terms of beam count and range (meters), the capture \textit{framerate}, and the \textit{field of view} (FOV).
These values are summarized in Table~\ref{tab:aquisition_stats}. The statistics are obtained from a combination of sources: the original dataset papers, sensor datasheets, and related setup or pipeline descriptions cited by the dataset authors. We verify these numbers through manual inspection and automated scripts, and, when relevant for consistency and fair comparison, derive additional statistics not reported in the original publications.

The second set of statistics, summarized in Table~\ref{tab:data_stats}, captures the intrinsic properties and diversity of the datasets. These include  the \textit{number of annotated human keypoints}, indicating the labeling standard adopted by the dataset; the \textit{average number of LiDAR points per human instance}, representing the mean point density on captured instances; the \textit{average bounding box size of human instances}, which is closely related to the image resolution; and the \textit{average human-to-sensor distance}. We also evaluate \textit{3D pose diversity}, which measures the variability of human poses within each dataset. To compute this, we align each posed SMPL model with a template model in T-pose, then calculate the mean squared error (MSE) between corresponding joints. 

Formally, let $N_{jts}$ and $N_{verts}$ represent the number of joints and number of vertices, respectively. $\hat{J} \in \mathbb{R}^{N_{jts} \times 3}$ and $J^{T} \in \mathbb{R}^{N_{jts} \times 3}$ denote the predicted and template joints, respectively. 
The posed joints are obtained as $\hat{J} = J_{reg} \cdot \hat{M}$, 
where $\hat{M} \in \mathbb{R}^{N_{verts} \times 3}$ is the predicted SMPL mesh generated using only the pose parameters, and 
the template joints as $J^{T} = J_{reg} \cdot M^{T}$, where $\hat{M}^{T} \in \mathbb{R}^{N_{verts} \times 3}$ is the SMPL template mesh in T-pose. 
Here, $J_{reg} \in \mathbb{R}^{N_{jts} \times N_{verts}}$ is the SMPL joint regressor matrix provided by the official SMPL model. 
The 3D pose diversity is then computed as:

\begin{equation}
   \text{dist($\hat{J}$, $J^{T}$)} = \frac{1}{N_{jts}} \sum_{i=1}^{N_{jts}} \left\| \hat{J_i} - J^{T}_i \right\|_2^2
\end{equation}

We select these characteristics to provide deeper insights for benchmarking 3D HPE and HMR methods, enabling a more informed interpretation of performance differences across datasets. 
In our GitHub repository, we provide detailed descriptions of the procedures used to extract these statistics for each dataset, along with the corresponding code and per-keypoint diversity metrics.

\begin{table}[]
\caption{Intrinsic characteristics and diversity of the lidar point cloud datasets.} 
\label{tab:data_stats}
\centering
\setlength{\tabcolsep}{5pt}
\resizebox{\linewidth}{!}{
\begin{tabular}{llll}
\toprule
Datasets & WOD \cite{sunScalabilityPerceptionAutonomous2020} & SLOPER4D \cite{daiSLOPER4DSceneAwareDataset2023} & Human-M3 \cite{fanHumanM3MultiviewMultimodal2023} \\
\midrule
\# Human keypoints & 14 & 21 & 15 \\
Avg. points per instance & 384.1 & 967.8 & 369.1 \\
Avg. bounding box size (px) & 10340.7 & 37471.1 & \na \\
Avg. human-sensor dist (m) & 14.5 & 2.8 & \na \\
3D pose diversity (cm) & 22.0 & 22.9 & 22.3 \\
\bottomrule 
\end{tabular}
}
\label{tab:dataset_intrinsic_stats}
\begin{tablenotes}
    \item 
   Note that statistics for Human-M3 cannot include average bounding box size, as no 2D information is provided. Similarly, average human-sensor distance cannot be computed because the point cloud scenes are generated by fusing four LiDAR views.
\end{tablenotes}
\end{table}

\subsection{Overview of Additional Public Datasets}
\label{sec:datasets_others}

Below, we provide a summary of other publicly available datasets.

\textbf{ The LiDARHuman26M dataset  \cite{liLiDARCapLongrangeMarkerless2022}} 
contains 184,048 frames of synchronized LiDAR point clouds, RGB images, and 3D human motion data from IMU-based systems. Each frame includes full-body 3D joints and SMPL parameters (pose, shape, global translation) for mesh recovery. It features 13 subjects performing 20 daily activities at 12-28 meters range, with preprocessing to remove background points and correct IMU localization errors.

\textbf{The HmPEAR dataset \cite{linHmPEARDatasetHuman2024} } 
is a large-scale, multi-modal dataset for 3D HPE and Human Action Recognition (HAR) in complex outdoor settings, containing 300K+ synchronized frames from a 128-beam LiDAR and three RGB cameras. Over 250K frames have refined 3D poses from motion capture, and 6K clips cover 40 daily actions by 25 subjects in 10 scenes under varying lighting. Annotations include actions, bounding boxes, SMPL meshes, and manually aligned pose-action data. 

\textbf{The HSC4D dataset \cite{daiHSC4DHumancentered4D2022}} 
provides 250K IMU frames (100 Hz) with SMPL parameters (global translation, 24-joint rotations, 10D shape) and 50K LiDAR frames (20 Hz) with dense point clouds and SLAM-based trajectories. All data is synchronized, calibrated, and refined via joint optimization for globally accurate, smooth human motion in world coordinates. Additional annotations include foot contact states and colorized 3D maps from a Trimble TX5 scanner for enhanced scene visualization.

\textbf{The PedX dataset \cite{kimPedXBenchmarkDataset2019}} 
is a large-scale multimodal benchmark for 3D pedestrian pose estimation in complex urban settings. Captured from a stationary vehicle with four synchronized RGB cameras and four roof-mounted LiDAR sensors, it offers high-resolution image triplets, stereo-merged LiDAR point clouds, and annotations including 2D/3D keypoints, SMPL body models, and instance segmentations. 
The dataset contains 5k pairs of stereo images and 2.5k 3D LiDAR frames, along with 14k human instances, each annotated with 2D and 3D pose labels.

\section{Benchmarks}\label{sec:benchmark}

In this section, we first introduce the metrics used for 3D HPE and HMR from LiDAR data. We then present benchmark tables summarizing all surveyed papers on the WOD \cite{sunScalabilityPerceptionAutonomous2020}, SLOPER4D \cite{daiSLOPER4DSceneAwareDataset2023}, and Human-M3 \cite{fanHumanM3MultiviewMultimodal2023} datasets for both HPE and HMR tasks. 

\subsection{Metrics}

We curate both widely used and recently introduced metrics in the field of LiDAR-based 3D human pose estimation and mesh recovery. To ensure consistency, we compile metrics from the surveyed papers and unify their definitions. We begin by introducing preliminary notations in Table~\ref{tab:notation}, followed by the formal metric equations. We indicate whether lower or higher values are better using arrows next to each metric title: $\downarrow$ for lower-is-better, and $\uparrow$ for higher-is-better.

\begin{table}[ht]
\small
\caption{Preliminary notations used in the metrics. \label{tab:notation}}

    \begin{tabular}{|>{\centering\arraybackslash}m{.12\linewidth}|>{\centering\arraybackslash}m{.76\linewidth}|}
         \hline
         \textbf{Symbol} & \textbf{Denotes} \\ \hline
         $P$ & Ground-truth keypoints \\ \hline
         $\hat{P}$ & Predicted keypoints \\ \hline         
         $p_i$ & Ground-truth positions of the $i$-th keypoint \\ \hline         
         $\hat{p}_i$ & Predicted positions of the $i$-th keypoint \\ \hline               
         $V$ & Ground-truth mesh vertices \\ \hline   
         $\hat{V}$ & Predicted mesh vertices \\ \hline            
         $v_i$ & Ground-truth positions of the $i$-th vertex \\ \hline
         $\hat{v}_i$ & Predicted positions of the $i$-th vertex \\ \hline
         $e_i$ & Ground-truth length of $i$-th edge in mesh \\ \hline  
         $\hat{e_i}$ & Predicted length of $i$-th edge in mesh \\ \hline  

         $l_i$ & Ground-truth length of $i$-th limb in skeleton \\ \hline  
         $\hat{l_i}$ & Predicted length of $i$-th limb in skeleton \\ \hline 
         
         $\delta$ & A threshold distance or size \\ \hline
         $\mathbb{I}(\cdot)$ & Indicator function, which equals 1 if the condition inside holds true, and 0 otherwise  \\ \hline
         $Tr(\cdot)$ & A transformation applied to the points \\ \hline     
         $A$& Total number of angles in the skeleton \\ \hline
         $E$ & Total number of edges in the mesh \\ \hline
         $N$ & Total number of keypoints \\ \hline
         $M$ & Total number of vertices in the mesh  \\ \hline
         $T$ &Total number of frames in the motion sequence \\ \hline

    \end{tabular}
\end{table}
\begin{table*}[h]

\caption{
Benchmark results for 3D HPE on the WOD \cite{sunScalabilityPerceptionAutonomous2020} validation split.}

\resizebox{1.0\textwidth}{!}{
\begin{tabular}{p{0.12\textwidth}p{0.02\textwidth}p{0.137\textwidth}p{0.448\textwidth}cccccc}
\toprule

\multirow{2}{*}{\textbf{Paper}}
& \multirow{2}{*}{\textbf{M}}
& \multirow{2}{*}{\textbf{Reference Paper}}
& \multirow{2}{*}{\textbf{Training/Evaluation Setup}}
& \multicolumn{3}{c}{\textbf{Data} \%}
& \multirow{2}{*}{\textbf{PEM}}
& \multirow{2}{*}{\textbf{OKS} \%} 
& \multicolumn{1}{c}{\multirow{2}{*}{\textbf{MPJPE}}} \\
& &  & & \textbf{1} & \textbf{10} & \textbf{100} & & & \\
\midrule

LPFormer~\cite{yeLPFormerLiDARPose2024} & L & LPFormer~\cite{yeLPFormerLiDARPose2024}  & \textbf{WOD v.1.4.2} - It is trained on the combined training and validation splits, and results are reported on the \textbf{test} set. 
&   &   & \cmark & 0.1524 & -  & 5.94    \\
\midrule

LPFormer~\cite{yeLPFormerLiDARPose2024} & L & DAPT~\cite{anPretrainingDensityAwarePose2025}  & \textbf{WOD v.1.4.2} - Results are reported on the full val set.
&   &   & \cmark & 0.0976  & -  & 6.16    \\
\midrule

WS-HPE~\cite{zhengMultimodal3DHuman2022} & L & WS-HPE~\cite{zhengMultimodal3DHuman2022} & 
\textbf{WOD v. ?} - LiDAR Branch Evaluation. Trained on 155,182 objects from internal data using 13 keypoints for 2D ground truth labeling. Evaluated on a WOD subset with 986 pedestrians manually annotated in 3D using the same 13 keypoints. Paired 2D/3D ground truth boxes were used for evaluation. &   &   & \cmark & - & 59.58 & 10.80

\\
\midrule

WS-HPE~\cite{zhengMultimodal3DHuman2022}  & CL &  WS-HPE~\cite{zhengMultimodal3DHuman2022} & Same as above &   &   & \cmark & - &  63.14  & 10.32   \\
\midrule

HUM3DIL~\cite{zanfirHUM3DILSemisupervisedMultimodal2022} & CL & HUM3DIL~\cite{zanfirHUM3DILSemisupervisedMultimodal2022} &  \textbf{WOD v.1.3.2} - 
Randomly split the WOD-val subset into 50\% for validation and 50\% for testing. Ground-truth camera and LiDAR bounding boxes are used for both training and evaluation.
&   &   & \cmark & -  & - & 6.72   \\
\midrule

HUM3DIL~\cite{zanfirHUM3DILSemisupervisedMultimodal2022} & CL & HUM3DIL~\cite{zanfirHUM3DILSemisupervisedMultimodal2022} &  \textbf{WOD v.1.3.2} - 
Evaluation of the fully supervised variant by removing the $\lambda$ term corresponding to weak 2D supervision.
&   &   & \cmark & - & - & 8.62   \\
\midrule

\multirow{3}{=}{HUM3DIL~\cite{zanfirHUM3DILSemisupervisedMultimodal2022}} & \multirow{3}{=}{L} & \multirow{3}{=}{GC-KPL~\cite{weng3DHumanKeypoints2023}} &  
\multirow{3}{=}{\textbf{WOD v.?} - For a fair comparison with GC-KPL~\cite{weng3DHumanKeypoints2023}, the model uses only the LiDAR branch with a randomly initialized backbone.}
& \cmark &   &   & -  & -  & 19.57 \\

\\

 &  &  &  &   & \cmark &   & -  & -  & 16.36 \\

\\

 &  &  &  &   &   & \cmark & -  & -  & 12.21 \\

\midrule

\multirow{3}{=}{GC-KPL~\cite{weng3DHumanKeypoints2023}} & \multirow{3}{=}{L} & \multirow{3}{=}{GC-KPL~\cite{weng3DHumanKeypoints2023}} & 
\multirow{3}{=}{\textbf{WOD v.?} - The model is pretrained on synthetic data only and fine-tuned on (cf. Data) \% of the ground-truth training set. For evaluation, 50\% of the WOD-val set is randomly selected for validation, and the remaining 50\% is used as the test split for benchmarking.}
& \cmark &   &   & -  & -  & 18.52

\\

\\

 &  &  &  &   & \cmark &   & -  & -  & 15.10 \\

\\

 &  &  &  &   &   & \cmark & -  & -  & 11.27 \\

\midrule

\multirow{3}{=}{GC-KPL~\cite{weng3DHumanKeypoints2023}} & \multirow{3}{=}{L} & \multirow{3}{=}{GC-KPL~\cite{weng3DHumanKeypoints2023}} & 
\multirow{3}{=}{\textbf{WOD v.?} -The model is pretrained on synthetic data and 5,000 samples from WOD-train, then fine-tuned on (cf. Data) \% of the ground-truth training set. For evaluation, 50\% of the WOD-val set is randomly selected for validation, with the remaining 50\% used as the test split for benchmarking.}

& \cmark &   &   & -  & -  & 17.87 \\

\\

 &  &  &  &   & \cmark &   & -  & -  & 14.51 \\

\\

 &  &  &  &   &   & \cmark & -  & -  & 10.73 \\

\midrule

\multirow{5}{=}{GC-KPL~\cite{weng3DHumanKeypoints2023}} & \multirow{5}{=}{L} & \multirow{5}{=}{GC-KPL~\cite{weng3DHumanKeypoints2023}} & 
\multirow{5}{=}{\textbf{WOD v.?} -The model is pretrained on synthetic data and 200,000 samples from WOD-train, then fine-tuned on (cf. Data) \% of the ground-truth training set. For evaluation, 50\% of the WOD-val set is randomly selected for validation, and the remaining 50\% is used as the test set for benchmarking.}
& \cmark &   &   & -  & -  & 17.80$\ast$ \\

\\

 &  &  &  &   & \cmark &   & -  & -  & 14.30$\ast$ \\
\\

 &  &  &  &   &   & \cmark & -  & -  & 10.60$\ast$ \\
\\

 &  &  &  & \cmark &   &   & -  & -  & 17.20 \\
\\

 &  &  &  &   & \cmark &   & -  & -  & 13.40 \\
\\

 &  &  & *KMeans clustering is used to label body parts in Stage I. 
 &   &   & \cmark & -  & -  & 10.10 \\
\midrule

GC-KPL~\cite{weng3DHumanKeypoints2023}  & L & GC-KPL~\cite{weng3DHumanKeypoints2023} & 
\textbf{WOD v.?} - Fully unsupervised setup with Stage I trained on solely synthetic data. Predicted keypoints are evaluated using Hungarian matching against ground truth.
&   &   &  & - & - & 17.7  \\
\midrule

GC-KPL~\cite{weng3DHumanKeypoints2023}  & L & GC-KPL~\cite{weng3DHumanKeypoints2023} & 
\textbf{WOD v.?} - Fully unsupervised setup with Stage I trained on 5,000 WOD samples. Predicted keypoints are evaluated using Hungarian matching against ground truth.
&   &   &  & - & - & 14.64  \\
\midrule

GC-KPL~\cite{weng3DHumanKeypoints2023}  & L & GC-KPL~\cite{weng3DHumanKeypoints2023} & 
\textbf{WOD v.?} - Fully unsupervised setup with Stage I trained on 200,000 WOD samples. Predicted keypoints are evaluated using Hungarian matching against ground truth.
&   &   &  & - & - & 13.92  \\
\midrule

DAPT~\cite{anPretrainingDensityAwarePose2025}  & L & DAPT~\cite{anPretrainingDensityAwarePose2025}  & 
\textbf{WOD v.2.0} -Trained and evaluated on WOD following pretraining on LiDARHuman26M~\cite{liLiDARCapLongrangeMarkerless2022}. 
&   &   & \cmark & - & - & 5.16  \\
\midrule

VoxelKP~\cite{shiVoxelKPVoxelbasedNetwork2023}  & L & VoxelKP~\cite{shiVoxelKPVoxelbasedNetwork2023}  & \textbf{WOD v.1.4.2} - 
To address limited training data, ground truth sampling is used to insert target objects from other frames into sampled frames.
&   &   & \cmark & 0.1695 & 73.00  & 8.87  \\
\midrule

PRN~\cite{fanLiDARHMR3DHuman2025} & L & LiDAR-HMR \cite{fanLiDARHMR3DHuman2025}  & \textbf{WOD v.2.0}  &   &   & \cmark & -  &  - & 6.78  \\ 
\midrule

WS-Fusion~\cite{bauerWeaklySupervisedMultiModal2023} & CL & WS-Fusion~\cite{bauerWeaklySupervisedMultiModal2023}  & 
\textbf{WOD 1.4.2} - Evaluation of the fully supervised approach using LiDAR point clouds and image-extracted 2D keypoints. 
&   &   & \cmark & -  & -  & 8.58 \\
\midrule

WS-Fusion~\cite{bauerWeaklySupervisedMultiModal2023} & CL & WS-Fusion~\cite{bauerWeaklySupervisedMultiModal2023}  & 
\textbf{WOD 1.4.2} - Evaluation of the weakly supervised approach using a 2D keypoint predictor to extract weak labels.
&   &   & \cmark & -  & -  & 9.01 \\

\bottomrule
\end{tabular}
}
\label{tab:benchmark_hpe_wod_val}

\begin{tablenotes}[para]
    \footnotesize \textbf{L} denotes \textit{LiDAR}, \textbf{C} denotes \textit{Camera}, and \textbf{M} indicates  the \textit{input modality}. MPJPE is reported in centimeters.
\end{tablenotes}
\end{table*}


\begin{table*}[t]
\centering

\caption{Benchmark results for 3D HMR on the WOD \cite{sunScalabilityPerceptionAutonomous2020} validation split.}

\resizebox{0.9\textwidth}{!}{
\begin{tabular}{p{0.15\textwidth}p{0.05\textwidth}p{0.15\textwidth}p{0.35\textwidth}ccc}

\toprule

\textbf{Paper}
& \textbf{M}
& \textbf{Reference Paper}
& \textbf{Training/Evaluation Setup}
& \textbf{MPJPE}
& \textbf{MPVPE}
& \textbf{MPERE} \\

\midrule

LiDAR-HMR~\cite{fanLiDARHMR3DHuman2025} & L & LiDAR-HMR~\cite{fanLiDARHMR3DHuman2025}  &
\textbf{WOD v.2.0.} They used keypoint annotations and input point clouds to reconstruct pseudo-human mesh labels, following a process similar to SMPLify-X \cite{pavlakos2019expressive}.
& 6.28 &  8.24 & 0.119 \\

\bottomrule
\end{tabular}}
\label{tab:benchmark_hmr_wod_val}
\begin{tablenotes}
    \item 
    \footnotesize \textbf{L} denotes \textit{LiDAR}, and \textbf{M} indicates  the \textit{input modality}. MPJPE, MPVPE and MPERE are reported in centimeters.
\end{tablenotes}
\end{table*}
\begin{table*}[t]
\centering
\tabcolsep 4pt

\caption{Benchmark results for 3D HPE on SLOPER4D \cite{daiSLOPER4DSceneAwareDataset2023}.}

\resizebox{0.85\textwidth}{!}{
\begin{tabular}{p{0.08\textwidth}p{0.03\textwidth}p{0.13\textwidth}p{0.4\textwidth}SS}

\toprule

\multirow{1}{*}{\textbf{Paper}}
& \multirow{1}{*}{\textbf{M}}
& \textbf{Reference Paper}
& \multirow{1}{*}{\textbf{Training/Evaluation Setup}}
& \multicolumn{1}{c}{\multirow{1}{*}{\textbf{MPJPE}}}
& \multicolumn{1}{c}{\multirow{1}{*}{\textbf{PA-MPJPE}}}   \\

 \midrule
PRN~\cite{fanLiDARHMR3DHuman2025}  & L & LiDAR-HMR~\cite{fanLiDARHMR3DHuman2025}  &   
\textbf{SLOPER4D v.1.0} - A data fragment was selected as the test set, containing 8,064 annotated human meshes, while the training set includes 24,936 annotations.
& 5.70 & {-} \\
\hline

PRN~\cite{fanLiDARHMR3DHuman2025}  & L & DAPT~\cite{anPretrainingDensityAwarePose2025}  & 
\textbf{SLOPER4D v.?} - pretrained on LiDARHuman26M~\cite{liLiDARCapLongrangeMarkerless2022} using a custom data synthesis pipeline. For fair comparison, DAPT replaces PRN’s point cloud backbone with PTv3 \cite{wu2024point}. Following NE-3D-HPE~\cite{zhangNeighborhoodEnhanced3DHuman2024}, each sequence is split into 16-frame patches, randomly shuffled, and the dataset is divided into training and testing sets with a 7:3 ratio.
& 4.876 & {-} \\
\hline

DAPT~\cite{anPretrainingDensityAwarePose2025}  & L & DAPT~\cite{anPretrainingDensityAwarePose2025}  & 
\textbf{SLOPER4D v.?} - pretrained on LiDARHuman26M~\cite{liLiDARCapLongrangeMarkerless2022} using a custom data synthesis pipeline. Following NE-3D-HPE~\cite{zhangNeighborhoodEnhanced3DHuman2024}, each sequence is split into 16-frame patches, randomly shuffled, and the dataset is divided into training and testing sets in a 7:3 ratio.
& 2.801 & 2.152 \\

\bottomrule
\end{tabular}}
\label{tab:benchmark_hpe_sloper_val}
\begin{tablenotes}
    \item 
    \footnotesize \textbf{L} denotes \textit{LiDAR}, and \textbf{M} indicates  the \textit{input modality}.  
The publicly released version 1.0 contains only 6 of the 15 sequences described in the paper and does not provide any data splits. MPJPE and PA-MPJPE are reported in centimeters.
\end{tablenotes}
\end{table*}


\begin{table*}[t]
\centering
\tabcolsep 4pt

\caption{Benchmark results for 3D HMR on SLOPER4D \cite{daiSLOPER4DSceneAwareDataset2023}.}
\resizebox{1.0\textwidth}{!}{
\begin{tabular}{p{0.15\textwidth}p{0.05\textwidth}p{0.15\textwidth}p{0.35\textwidth}SSSSccc}

\toprule
\multirow{2}{*}{\textbf{Paper}}
& \multirow{2}{*}{\textbf{M}}
& \multirow{2}{*}{\textbf{Reference Paper}}
& \multirow{2}{*}{\textbf{Training/Evaluation Setup}}
& \multicolumn{1}{c}{\textbf{MPJPE}}
& \multicolumn{1}{c}{\textbf{MPVPE}}
& \multicolumn{1}{c}{\textbf{MPJPE}}
& \multicolumn{1}{c}{\textbf{MPVPE}}
& \multirow{2}{*}{\textbf{MPERE}} 
& \textbf{Ang} 
& \textbf{Accel} \\

& &  & & \multicolumn{1}{c}{\textbf{No Align.}} & \multicolumn{1}{c}{\textbf{No Align.}} & \multicolumn{1}{c}{\textbf{Align.}} & \multicolumn{1}{c}{\textbf{Align.}} & &  \multicolumn{1}{c}{\textbf{Err}}  & \multicolumn{1}{c}{\textbf{Err}} \\
 \midrule

LiDAR-HMR~\cite{fanLiDARHMR3DHuman2025}  & L & LiDAR-HMR~\cite{fanLiDARHMR3DHuman2025}  & 
\textbf{SLOPER4D v.1.0} - They select a data fragment as the test set, with 24,936 annotated human meshes in the train set and 8,064 in the test set.
& 5.103 & 5.189 & {-} & {-} & 0.094 & - & - \\
\hline

SMPLify-3D~\cite{dumontImprovingImagebased3D}  & CL & SMPLify-3D~\cite{dumontImprovingImagebased3D} & 
\textbf{SLOPER4D v.1.0} - They follow the splits proposed by LiDAR-HMR \cite{fanLiDARHMR3DHuman2025}. 
& 5.58 & 6.44 & 4.35 & {-} & 0.06 & {-} & {-} \\
\hline

LiveHPS~\cite{renLiveHPSLiDARbasedScenelevel2024}  & L & LiveHPS~\cite{renLiveHPSLiDARbasedScenelevel2024} & 
\textbf{SLOPER4D v.?} - Global acceleration error is reported. \textit{Accel Err} is taken from LiveHPS++ \cite{renLiveHPSRobustCoherent2024}. LiveHPS and SLOPER4D share authors, suggesting potential but unverified access to the full dataset and official splits.
& 7.773$_t$ & 8.583$_t$ & 4.622$_p$ & 5.672$_p$ & {-} & 12.77 & 564 \\
\hline

LiveHPS~\cite{renLiveHPSLiDARbasedScenelevel2024}  & L & LiveHPS++~\cite{renLiveHPSRobustCoherent2024} & 
\textbf{SLOPER4D v.?} - Same as above.
& 8.835$_t$ & 9.585$_t$ & 5.337$_s$ & 6.315$_s$ & {-} & 13.08 & 588 \\
\hline

LiveHPS++~\cite{renLiveHPSRobustCoherent2024}  & L & LiveHPS++~\cite{renLiveHPSRobustCoherent2024} & 
\textbf{SLOPER4D v.?} - Same as above.
& 7.698$_t$ & 8.167$_t$ & 4.27$_s$ & 5.062$_s$ & {-} & 11.92 & 434 \\
\hline

ReMP~\cite{jangReMPReusableMotion2025} & L & ReMP~\cite{jangReMPReusableMotion2025} & 
\textbf{SLOPER4D v.?} - No data preparation details provided. This model is trained on synthetic AMASS \cite{AMASS:ICCV:2019} data without using motion prior. 
& 87.46 & 96.94 & {-} & {-} & {-} & 9.96 & {-} \\
\hline

ReMP~\cite{jangReMPReusableMotion2025} & L & ReMP~\cite{jangReMPReusableMotion2025} & 
\textbf{SLOPER4D v.?} - No data preparation details provided. 
& 21.66 & 22.03 & {-} & {-} & {-} & 8.58 & {-} \\
\hline

PE-Proj~\cite{linHmPEARDatasetHuman2024} & CL & HmPear~\cite{linHmPEARDatasetHuman2024} & 
\textbf{SLOPER4D official} - Trained on a mix of SLOPER4D and HmPEAR, evaluated on SLOPER4D to explore cross-task synergies. The authors state they use SLOPER4D’s official splits (first 11 sequences for training, remaining 4 for testing) without further clarification. 
& 5.55 & {-} & 3.89 & {-} & {-} & {-} & {-} \\
\hline

Pear-Proj (BestPE)~\cite{linHmPEARDatasetHuman2024} & CL & HmPear~\cite{linHmPEARDatasetHuman2024} & 
\textbf{SLOPER4D official} - Trained on HmPEAR and evaluated on SLOPER4D’s official test split, using only the optimal HPE model with the Action Recognition branch removed.
& 8 & {-} & 4.42 & {-} & {-} & {-} & {-} \\
\hline

PEAR-Proj (BestAR)~\cite{linHmPEARDatasetHuman2024} & CL & HmPear~\cite{linHmPEARDatasetHuman2024} &
\textbf{SLOPER4D official} - Trained on HmPEAR and evaluated on SLOPER4D’s official test split, using only the optimal Action Recognition model with the HPE branch removed.
& 7.96 & {-} & 4.52 & {-} & {-} & {-} & {-} \\
\hline

NE-3D-HPE~\cite{zhangNeighborhoodEnhanced3DHuman2024} & L 
& NE-3D-HPE~\cite{zhangNeighborhoodEnhanced3DHuman2024} & 
\textbf{SLOPER4D v.?} - Each SLOPER4D sequence is segmented into 16-frame patches, which are then randomly shuffled. The data is split into training and testing sets in a 7:3 ratio. Note: NE-3D-HPE and SLOPER4D share authors, indicating possible though unverified access to the full dataset and official splits.
& 9.68 & {-} & 7.67 & {-} & {-} & {-} & 38.55\\
\hline

LiDARCap~\cite{liLiDARCapLongrangeMarkerless2022}  & L & LiveHPS~\cite{renLiveHPSLiDARbasedScenelevel2024} & 
\textbf{SLOPER4D v.?} - Implemented by LiveHPS for comparison, with a shape regression head added, mirroring the architecture of their pose regression head. Note: LiveHPS and SLOPER4D share authors, suggesting possible but unverified access to the full dataset and official splits.
& 17.933$_t$ & 18.539$_t$ & 6.74$_p$ & 8.008$_p$ & {-} & 15.92 & {-} \\
\hline

LiDARCap~\cite{liLiDARCapLongrangeMarkerless2022}  & L & LiveHPS++~\cite{renLiveHPSRobustCoherent2024} & 
\textbf{SLOPER4D v.?} - Global acceleration error is reported. LiveHPS++ and SLOPER4D share authors, suggesting possible but unverified access to the full dataset and official splits.
& 13.871$_t$ & 14.779$_t$ & 7.164$_s$ & 8.423$_s$ & {-} & 13.72 & 616 \\
\hline

LiDARCap~\cite{liLiDARCapLongrangeMarkerless2022}  & L & NE-3D-HPE~\cite{zhangNeighborhoodEnhanced3DHuman2024} \& DAPT~\cite{anPretrainingDensityAwarePose2025} & 
\textbf{SLOPER4D v.?} - Each SLOPER4D sequence is segmented into 16-frame patches, which are
then randomly shuffled. The data is split into training and testing sets in a 7:3 ratio. NE-3D-HPE and SLOPER4D share authors, suggesting possible but unverified access to the full dataset and official splits.
& 10.189 & {-} & 7.893 & {-} & {-} & {-} & 40.09 \\
\hline

LiDARCap~\cite{liLiDARCapLongrangeMarkerless2022}  & L & SLOPER4D~\cite{daiSLOPER4DSceneAwareDataset2023} \& HmPear~\cite{linHmPEARDatasetHuman2024} & 
\textbf{SLOPER4D official} - Reported by SLOPER4D paper. 
& 8.61 & {-} & 6.51 & {-} & {-} & {-} & {-} \\
\hline

LiDARCap~\cite{liLiDARCapLongrangeMarkerless2022}  & L & SMPLify-3D~\cite{dumontImprovingImagebased3D} &  
\textbf{SLOPER4D v.1.0} - They follow LiDAR-HMR's \cite{fanLiDARHMR3DHuman2025} proposed splits.
& 15.83 & {-} & 14.81 & {-} & 0.05 & {-} & {-} \\
\hline

\hline

LIP~\cite{renLiDARaidInertialPoser2023}  & IL & LiveHPS~\cite{renLiveHPSLiDARbasedScenelevel2024} & 
\textbf{SLOPER4D v.?} - Implemented by LiveHPS for comparison, using only the LiDAR branch of LIP and adding a shape regression head mirroring the pose head architecture. LiveHPS and SLOPER4D share authors, suggesting possible but unverified access to the full dataset and official splits. 
& 9.481$_t$ & 10.634$_t$ & 6.011$_p$ & 7.49$_p$ & {-} & 19.95 & {-} \\
\hline

LIP~\cite{renLiDARaidInertialPoser2023}  & IL & LiveHPS++~\cite{renLiveHPSRobustCoherent2024} & 

\textbf{SLOPER4D v.?} - Global acceleration error is reported. LiveHPS++ and SLOPER4D share authors, suggesting possible but unverified access to the full dataset and official splits. 
& 13.469$_t$ & 14.69$_t$ & 7.438$_s$ & 9.189$_s$ & {-} & 20.53 & 659 \\

\bottomrule
\end{tabular}}
\label{tab:benchmark_hmr_sloper_val}

\begin{tablenotes}
    \item 
    \footnotesize   \textbf{L} denotes \textit{LiDAR}, \textbf{C} denotes \textit{Camera}, \textbf{I} denotes \textit{IMU}, and \textbf{M} indicates  the \textit{input modality}. 
The official SLOPER4D split includes 11 training and 4 testing sequences. 
The publicly released version 1.0 contains only 6 of the 15 sequences described in the paper and does not provide any data splits.
For papers that report \texttt{J/V Err(P/PS/PST)} using SMPL-regressed joints, we treat \texttt{J Err(PST)} as \textbf{MPJPE} and \texttt{V Err(PST)} as \textbf{MPVPE}. When available, we use \texttt{J Err(P)} and \texttt{V Err(P)} as \textbf{MPJPE Align.} and \textbf{MPVPE Align.}, respectively. If \texttt{J Err(P)} and \texttt{V Err(P)} are not available, we use \texttt{J Err(PS)} and \texttt{V Err(PS)} instead. 
Subscripts \textbf{p}, \textbf{s}, and \textbf{t} indicate \texttt{J/V Err} computed with \texttt{P}, \texttt{PS}, and \texttt{PST} joints, as reported in the original papers. 
MPJPE, MPVPE, and their aligned variants are reported in centimeters. Accel Err is reported in $cm/s^2$.
\end{tablenotes}
\end{table*} 
\begin{table*}[t]
\centering
\tabcolsep 4pt

\caption{Benchmark results for 3D HPE on Human-M3 \cite{fanHumanM3MultiviewMultimodal2023}.}

\resizebox{0.88\textwidth}{!}{
\begin{tabular}{p{0.15\textwidth}p{0.05\textwidth}p{0.15\textwidth}p{0.35\textwidth}S}

\toprule

\multirow{1}{*}{\textbf{Paper}}
& \multirow{1}{*}{\textbf{M}}
& \textbf{Reference Paper}
& \multirow{1}{*}{\textbf{Training/Evaluation Setup}}
& \multicolumn{1}{c}{\multirow{1}{*}{\textbf{MPJPE}}}   \\
\midrule

PRN~\cite{fanLiDARHMR3DHuman2025} & L & LiDAR-HMR~\cite{fanLiDARHMR3DHuman2025}  & The training set contains 80,103 annotated human keypoints, and the test set contains 8,951. This appears to follow the official 90–10\% split described in the Human-M3 paper. 
& 8.22  \\
\hline

SMPLify-3D~\cite{dumontImprovingImagebased3D} & CL & SMPLify-3D~\cite{dumontImprovingImagebased3D}  & 
SMPLify-3D appears to follow the LiDAR-HMR \cite{fanLiDARHMR3DHuman2025} protocol, based on the reported test set size. This likely corresponds to the official 90–10\% split described in the Human-M3 paper. 
& 8.39   \\
\hline

LiDARCap~\cite{liLiDARCapLongrangeMarkerless2022} & L & SMPLify-3D~\cite{dumontImprovingImagebased3D}  & 
Same as above. & 17.58   \\
\hline

DAPT ~\cite{anPretrainingDensityAwarePose2025}  & L & DAPT~\cite{anPretrainingDensityAwarePose2025}  & 
Given the large dataset size and minimal variation between consecutive frames, only 20\% of the data is used for training. 
& 5.976   \\
\hline

LPFormer~\cite{yeLPFormerLiDARPose2024} & L & DAPT ~\cite{anPretrainingDensityAwarePose2025}  & 
Same as above.
& 8.369   \\
\hline

PRN~\cite{fanLiDARHMR3DHuman2025} & L & DAPT~\cite{anPretrainingDensityAwarePose2025}  & 
Same as above.
& 7.056  \\
\hline

MMVP \cite{fanHumanM3MultiviewMultimodal2023} & CL & Human-M3 ~\cite{fanHumanM3MultiviewMultimodal2023}  & - & 7.9  \\

\bottomrule
\end{tabular}}
\label{tab:benchmark_hpe_humanm3_val}
\begin{tablenotes}
    \item 
    \footnotesize  \textbf{L} denotes \textit{LiDAR}, \textbf{C} denotes \textit{Camera}, and \textbf{M} indicates  the \textit{input modality}. MPJPE is reported in centimeters.
\end{tablenotes}
\end{table*}


\begin{table*}[t]
\centering
\tabcolsep 4pt

\caption{Benchmark results for 3D HMR on Human-M3 \cite{fanHumanM3MultiviewMultimodal2023}. }
\resizebox{1.0\textwidth}{!}{
\begin{tabular}{p{0.15\textwidth}p{0.05\textwidth}p{0.15\textwidth}p{0.35\textwidth}SSSSccc}

\toprule
\multirow{2}{*}{\textbf{Paper}}
& \multirow{2}{*}{\textbf{M}}
& \multirow{2}{*}{\textbf{Reference Paper}}
& \multirow{2}{*}{\textbf{Training/Evaluation Setup}}
& \multicolumn{1}{c}{\textbf{MPJPE}}
& \multicolumn{1}{c}{\textbf{MPVPE}}
& \multicolumn{1}{c}{\textbf{MPJPE}}
& \multicolumn{1}{c}{\textbf{MPVPE}}
& \multirow{2}{*}{\textbf{MPERE}} 
& \multicolumn{1}{c}{\textbf{Ang}} 
& \multicolumn{1}{c}{\textbf{Accel}}\\

& &  & & \multicolumn{1}{c}{\textbf{No Align.}} & \multicolumn{1}{c}{\textbf{No Align.}} & \multicolumn{1}{c}{\textbf{Align.}} & \multicolumn{1}{c}{\textbf{Align.}} & & \multicolumn{1}{c}{\textbf{Err}} & \multicolumn{1}{c}{\textbf{Err}} \\
 \midrule

LiveHPS~\cite{renLiveHPSLiDARbasedScenelevel2024} & L & FreeCap ~\cite{xueFreeCapHybridCalibrationFree2025}  & 
They use a single camera. No details are provided regarding dataset splits. 
& 9.711$_t$ & 10.31$_t$ & 5.781$_s$ & 7.127$_s$ & - & 10.44 & 1258 \\
\hline

FreeCap ~\cite{xueFreeCapHybridCalibrationFree2025} & CL & FreeCap ~\cite{xueFreeCapHybridCalibrationFree2025} & 
Same as above.
& 9.647$_t$ & 10.267$_t$ & 5.545$_s$ & 6.852$_s$ & - & 9.14 & 960 \\
\hline

LiDAR-HMR~\cite{fanLiDARHMR3DHuman2025}  & L & LiDAR-HMR~\cite{fanLiDARHMR3DHuman2025}  & 
The training set contains 80,103 annotated human keypoints, and the test set contains 8,951. This appears to follow the official 90–10\% split described in the Human-M3 paper. 
They use keypoint annotations and point clouds to reconstruct pseudo ground-truth human meshes, following a process similar to SMPLify-X \cite{pavlakos2019expressive}.
& 7.76 & 8.95 & {-} & {-} & 0.088 & - & - \\

\bottomrule
\end{tabular}}
\label{tab:benchmark_hmr_humanm3_val}
\begin{tablenotes}
    \item 
    \footnotesize  \textbf{L} denotes \textit{LiDAR}, \textbf{C} denotes \textit{Camera}, and \textbf{M} indicates  the \textit{input modality}. 
For papers that report \texttt{J/V Err(P/PS/PST)} using SMPL-regressed joints, we treat \texttt{J Err(PST)} as \textbf{MPJPE} and \texttt{V Err(PST)} as \textbf{MPVPE}. When available, we use \texttt{J Err(P)} and \texttt{V Err(P)} as \textbf{MPJPE Align.} and \textbf{MPVPE Align.}, respectively. If \texttt{J Err(P)} and \texttt{V Err(P)} are not available, we use \texttt{J Err(PS)} and \texttt{V Err(PS)} instead.
Subscripts \textbf{p}, \textbf{s}, and \textbf{t} indicate \texttt{J/V Err} computed with \texttt{P}, \texttt{PS}, and \texttt{PST} joints, as reported in the original papers. 
MPJPE, MPVPE, and their aligned variants are reported in centimeters. Accel Err is reported in $cm/s^2$.
\end{tablenotes}
\end{table*}

\paragraph{Mean Per-Joint Position Error (MPJPE)$\downarrow$} 
MPJPE measures the mean Euclidean distance between predicted and ground-truth joint positions across all corresponding joints in the skeleton. For temporal methods, the error is additionally averaged over all frames. In the Global Human Pose Estimation paradigm, this metric is referred to as G-MPJPE, where joint positions are evaluated in global coordinates. The error is typically reported in millimeters.
\begin{equation}
    \text{MPJPE($P,\hat{P}$)} = \frac{1}{N} \sum_{i=1}^{N} \| \hat{p}_i - p_i \|_2
\end{equation}

In some works, this metric is referred to as \texttt{J Err(PST)}, where \texttt{PST} indicates that joint positions are derived from the SMPL model using pose, shape, and translation parameters.

\paragraph{Procrustes Aligned Mean Per-Joint Position Error (PA-MPJPE)$\downarrow$}

PA-MPJPE measures the MPJPE after applying a transformation $Tr$ (typically including translation, rotation, and scaling) to align the predicted pose with the ground truth. This ensures the error reflects only pose prediction accuracy, not misalignment. Procrustes Alignment (PA) used in this metric may be restricted to rigid transformations (translation and rotation only), excluding scaling. The error is typically reported in millimeters.
\begin{equation}
    \text{PA-MPJPE($P,\hat{P}$)} = \frac{1}{N} \sum_{i=1}^{N} \| Tr(\hat{p}_i) - p_i \|_2
\end{equation}

In some works, PA-MPJPE is referred to as \texttt{J Err(P/PS)}, where \texttt{P} and \texttt{PS} indicate that joint positions are derived from the SMPL model using pose only, or pose and shape parameters, respectively.

\paragraph{Percentage of Correct Keypoints (PCK/PCK$@\delta$)$\uparrow$}
This metric measures the percentage of predicted joints that lie within a specified error threshold 
$\delta$ from their corresponding ground-truth positions. Common absolute thresholds include $30$ or $50$ millimeters (or centimeters). Alternatively, the threshold can be normalized and expressed as a percentage of the object size. For temporal methods, the percentage is averaged over all frames.
\begin{equation}
    \text{PCK$@\delta(P,\hat{P}$)} = \frac{100}{N} \sum_{i=1}^{N} \mathbb{I} \left( \| \hat{p}_i - p_i \|_2 \leq \delta \right)
\end{equation}
\paragraph{Pose Estimation Metric (PEM)$\downarrow$} 
PEM was developed specifically for the WOD estimation challenge. It measures keypoint matching accuracy using the sum of MPJPE for matched keypoints, along with a penalty term for unmatched keypoints.
\begin{equation}
    \text{PEM($P,\hat{P}$)} = \frac{\sum_{i \in X} \| \hat{p}_i - p_i \|_2 + C|Y|}{|X| + |Y|}
\end{equation}
where $X$ and $Y$ are the sets of matched and unmatched keypoints, and $C$ is a constant. Note that $Y$ includes both the ground truth keypoints without matching predicted keypoints and the predicted keypoints without matching ground truth keypoints. 
\paragraph{Mean Per-Vertex Position Error (MPVPE/PA-MPVPE)$\downarrow$}
MPVPE is the mesh-based counterpart to MPJPE in Human Mesh Recovery. It measures the mean Euclidean distance between predicted and ground-truth mesh vertices, and is averaged across all frames in temporal methods. This metric is sometimes denoted as \texttt{V Err(PST)}, where \texttt{PST} indicates that vertex positions are obtained from the SMPL model using pose, shape, and translation parameters.
\begin{equation}
    \text{MPVPE($V,\hat{V}$)} = \frac{1}{M} \sum_{i=1}^{M} \| \hat{v}_i - v_i \|_2
\end{equation}

The PA-aligned variant, PA-MPVPE, is sometimes denoted as \texttt{V Err(P/PS)}, where \texttt{P} and \texttt{PS} indicate that vertex positions were extracted from the SMPL model using pose only, or pose and shape parameters, respectively.
\paragraph{Mean Per-Edge Relative Error (MPERE)$\downarrow$} 
MPERE is particularly useful for evaluating mesh reconstruction quality in non-parametric methods, where vertex connectivity and spatial relationships are not fixed. It is especially effective for assessing the accuracy of short edges in densely connected regions of the mesh. MPERE is computed as the ratio of the Manhattan distance between predicted and ground-truth edge lengths to the ground-truth edge length, averaged across all mesh edges. In temporal methods, the metric is also averaged across all frames.
\begin{equation}
    \text{MPERE} = \frac{1}{E} \sum_{i=1}^{E} \frac{\left| \hat{e}_i - e_i \right|}{e_i}
\end{equation}
\paragraph{Angular Error (Ang Err)$\downarrow$} 
This metric evaluates local pose accuracy by measuring the average angular difference between the predicted and ground-truth joint rotations. It is typically computed as the mean geodesic distance between the predicted rotation matrices $\hat{R}_i$ and the ground-truth rotation matrices $R_i$ for all joints: 
\begin{equation}
\text{Ang Err} = \frac{1}{N} \sum_{i=1}^{N} \cos^{-1} \left( \frac{\mathrm{tr}(R_i^{\top} \hat{R}_i) - 1}{2} \right)
\end{equation}
where $\mathrm{tr}(\cdot)$ denotes the trace of a matrix. The result is typically expressed in degrees or radians.

\paragraph{Object Keypoint Similarity (OKS)$\uparrow$}

OKS assesses the similarity between predicted and visible ground-truth keypoints by accounting for the distance between keypoints, the overall object size (e.g., human body), and the visibility of each keypoint.

\begin{equation}
    \text{OKS} = \frac{1}{N} \sum_{i=1}^{N} \exp \left( -\frac{\left\| \hat{p}_i - p_i \right\|_2^2}{2 s^2 k_i^2} \right)
\end{equation}

Here, $k_i$ is a per-joint constant defined by COCO, which accounts for keypoint visibility and localization uncertainty, while $s$ represents the scale of the detected person.
Note that two OKS variants are commonly used in LiDAR-based HPE/HMR literature: OKS-AP, which computes average precision over multiple OKS thresholds, and OKS-Acc, which reports the percentage of keypoints above a fixed OKS threshold.
\paragraph{Average Displacement Error (ADE)$\downarrow$}

ADE measures the average Euclidean distance between predicted and ground-truth trajectories over time. It is particularly relevant for evaluating temporal consistency and dynamic motion accuracy:

\begin{equation}
\text{ADE} = \frac{1}{T \cdot N} \sum_{t=1}^{T} \sum_{i=1}^{N} \left\| \hat{p}_{t,i} - p_{t,i} \right\|_2
\end{equation}
where $\hat{p}_{t,i}$ is the predicted 3D position of keypoint $i$ at time $t$, and $p_{t,i}$ is the ground-truth 3D position of keypoint $i$ at time $t$.

\paragraph{Limb Angle Error (LAE)$\downarrow$} 
LAE quantifies the average angular difference between the predicted limb orientations $\hat{\theta}$ and the ground-truth orientations $\theta$, for corresponding limbs such as the arms and legs \cite{kovacsLidPoseRealTime3D2024}.

\begin{equation}
    \text{LAE} = \frac{1}{A} \sum_{i=1}^{A} \left| \hat{\theta}_i - \theta_i \right|
\end{equation}
\paragraph{Limb Length Error (LLE)$\downarrow$}
LLE evaluates how accurately the network predicts the overall lengths of limbs, such as arms and legs. It does not penalize joint-level inaccuracies (e.g., at the elbow or knee) as long as the total limb length is estimated correctly \cite{kovacsLidPoseRealTime3D2024}. 

\begin{equation}
    \text{LLE} = \frac{1}{L} \sum_{i=1}^{L} \left| \hat{l}_i - l_i \right|
\end{equation}
\paragraph{Acceleration Error (Accel Err)$\downarrow$}
Acceleration error, expressed in meters per second squared ($m/s^2$) or centimeters per second squared ($cm/s^2$), is reported as an evaluation metric for sequential data. It reflects the temporal smoothness and physical plausibility of predicted motion.
\begin{equation}
    \text{Accel Err} = \frac{1}{T} \sum_{t=1}^{T} \left| \hat{a}(t) - a(t) \right|
\end{equation}
Here, $\hat{a(t)}$ and $a(t)$ denote the predicted and ground-truth accelerations, respectively, computed from keypoint or vertex positions at time $t$.
\paragraph{Chamfer-Distance (CD/SU-CD)$\downarrow$}
CD measures the similarity between two sets of 3D points by computing the average distance from each point in set $R$ to its nearest neighbor in set $Q$, and vice versa. It is commonly used to evaluate the distance between predicted mesh vertices or 3D keypoints and a 3D human point cloud input. The metric is typically reported in millimeters. 
\begin{equation}
\begin{aligned}
    \begin{split}
    \text{CD} = & \frac{1}{|R|} \sum_{r_i \in R} \min_{q_j \in Q} \| r_i - q_j \|^2 \\ 
    & + \frac{1}{|Q|} \sum_{q_j \in Q} \min_{r_i \in R} \| q_j - r_i \|^2
    \end{split}
    \end{aligned}
\end{equation}
As the name suggests, Scene-level Unidirectional Chamfer Distance (SU-CD) is a unidirectional variant of Chamfer Distance, introduced in LiveHPS \cite{renLiveHPSLiDARbasedScenelevel2024}. It is particularly useful for evaluating global translation error. In LiveHPS, it was used to measure the displacement from the raw input point cloud to the predicted mesh. 
\begin{equation}
    \text{SU-CD} = \frac{1}{|R|} \sum_{r_i \in R} \min_{q_j \in Q} \| r_i - q_j \|^2
\end{equation}

\subsection{3D Human Pose Estimation and Mesh Recovery Benchmarks}

In this section, we present detailed benchmark tables for the WOD~\cite{sunScalabilityPerceptionAutonomous2020}, SLOPER4D~\cite{daiSLOPER4DSceneAwareDataset2023}, and Human-M3~\cite{fanHumanM3MultiviewMultimodal2023}, covering both human pose estimation and human mesh recovery tasks. We present benchmark tables for the WOD in Table~\ref{tab:benchmark_hpe_wod_val} for HPE and Table~\ref{tab:benchmark_hmr_wod_val} for HMR. The SLOPER4D benchmarks are presented in Table~\ref{tab:benchmark_hpe_sloper_val} for HPE and Table~\ref{tab:benchmark_hmr_sloper_val} for HMR. Finally, for Human-M3, the HPE benchmark is shown in Table~\ref{tab:benchmark_hpe_humanm3_val}, and the HMR benchmark in Table~\ref{tab:benchmark_hmr_humanm3_val}. 

Alongside various evaluation metrics, we include \textit{Reference Paper} and \textit{Training/Evaluation Setup} columns. The \textit{Reference Paper} section highlights cases where the same method yields different results on a benchmark due to variations in training setups or evaluation schemes used by the original reference paper. The \textit{Training/Evaluation Setup} column summarizes key details of the training and testing procedures as reported by the authors. This includes dataset-specific practices such as data splitting, mixing, or cleaning, as well as the use of pseudo labels or any modifications to evaluation metrics. Where available, dataset versions are specified, and when not clearly stated by the authors, we make an effort to indicate potential access to the official data.

It is important to note that for SLOPER4D and Human-M3 (as shown in Tables~\ref{tab:benchmark_hmr_sloper_val} and~\ref{tab:benchmark_hmr_humanm3_val}), some works evaluate MPJPE, MPVPE, and PA-like metrics using joints regressed from the SMPL(-X) model. 

While compiling results for HMR tasks on SLOPER4D and Human-M3, we observed that there is no standard method for aligning predicted and ground-truth meshes when evaluating local pose. To avoid confusion, we use the broader terms MPJPE Align. and MPVPE Align. to cover three alignment conventions: the widely used PA-MPJPE/PA-MPVPE, and two others denoted as \texttt{J/V Err(P)} and \texttt{J/V Err(PS)} in relevant papers.

For the papers using the \texttt{J/V Err(P)}, \texttt{J/V Err(PS)}, as well as \texttt{J/V Err(PST)} conventions: \texttt{J} refers to joints, \texttt{V} to mesh vertices, and  \texttt{Err} to the computed error. The letters \texttt{P}, \texttt{PS}, and \texttt{PST} indicate the SMPL parameters used to generate the predicted outputs: \texttt{P} for pose only, \texttt{PS} for pose and shape, and \texttt{PST} for pose, shape, and translation.
\texttt{J/V Err(PST)} thus corresponds directly to MPJPE and MPVPE, as it compares predicted joints and vertices to ground truth in global coordinates. 
For aligned metrics, we use \texttt{J/V Err(P)} when available; otherwise, we fall back on \texttt{J/V Err(PS)}. Subscripts $p$, $s$, and $t$ indicate whether the error was computed using \texttt{P}, \texttt{PS}, or \texttt{PST}, respectively, following the conventions of the original papers.

WOD has approximately 800 scenes for training and 200 scenes for validation. For SLOPER4D, the original paper introduces an official training and testing split with 11 sequences for training and 4 for testing, but this split is not included in the publicly available data. The released version, \textbf{v1.0}, contains only 6 of the 15 sequences mentioned in the paper. For Human-M3, the official split allocates 90\% of each sequence for training and reserves the final 10\% of frames for testing.

We believe that these benchmark tables will serve as a common reference point, helping to connect researchers, and driving progress in the field under fair and transparent conditions.
 
\section{Future Directions}\label{sec:future}

In this section, we identify and discuss key challenges and potential research directions for 3D human understanding from in-the-wild LiDAR data.

\subsection{Addressing Data Scarcity }

A central challenge in 3D HPE and HMR from LiDAR data is the scarcity of annotated datasets. To address this, researchers have begun exploring weakly-supervised learning strategies. However, most existing weakly-supervised methods still rely on auxiliary data modalities such as RGB images or IMU signals.
We believe a promising research direction lies in reducing dependency on these extra modalities by investigating how weakly-supervised HPE and HMR can be achieved using only LiDAR data.

\paragraph{Leveraging Existing LiDAR Datasets Without Pose Annotations}
To minimize reliance on additional modalities, existing LiDAR point cloud datasets that include human (without 2D/3D pose labels) can be repurposed. These datasets can support weak supervision strategies through pseudo-labeling, self-training, or contrastive learning paradigms.

\paragraph{Incorporating Temporal Information}
Temporal coherence offers a rich source of motion dynamics. Weakly-supervised methods can benefit from modeling sequences of LiDAR frames, extracting temporal cues to enhance pose estimation accuracy without requiring extra supervision.

\paragraph{Toward Realistic Synthetic Data Generation}
Another direction is the generation of high-quality, realistic synthetic data. Current synthetic pipelines simulate human LiDAR point clouds using AMASS \cite{AMASS:ICCV:2019} via raycasting. However, this poses two key issues. The first issue is domain mismatch, AMASS primarily contains controlled indoor poses, whereas datasets like WOD~\cite{sunScalabilityPerceptionAutonomous2020} or SLOPER4D~\cite{daiSLOPER4DSceneAwareDataset2023} involve outdoor settings.
The second issue is the realism gap, raycasting may not fully capture the noise, sparsity, and viewpoint characteristics of real LiDAR sensors.
To overcome these issues, learning to generate synthetic LiDAR data directly from real-world distributions could be more effective. With the advances in generative modeling, particularly diffusion models, this is a promising area for exploration.

\paragraph{Data-Efficient 3D HPE and HMR}
Currently, few methods (e.g., GC-KPL \cite{weng3DHumanKeypoints2023}) address data efficiency in LiDAR-based pose estimation. Inspired by approaches in 2D/3D vision, one promising avenue is to pretrain models in a self-supervised fashion, then fine-tune them for downstream tasks (3D HPE/HMR) under minimal supervision.

\paragraph{Weakly-Supervised 3D HMR}
While 3D HPE has seen weak supervision attempts, all existing 3D HMR methods remain fully supervised (outside of data annotation pipelines). Developing weakly-supervised methods specifically for HMR is a largely unexplored yet promising research direction.

\subsection{Eliminating Camera Parameter Dependency}
Most of the current multi-modal methods using LiDAR and RGB data depend heavily on accurate camera parameters to align 2D-3D correspondences. This introduces significant practical challenges, especially in real-world datasets. A future direction would be to replace this dependency with learnable modules, enabling end-to-end alignment without the need for explicit calibration.

\subsection{Domain Adaptation in LiDAR-based 3D HPE and  HMR}

\paragraph{Bridging the Domain Gap Between Sensors}
One challenge in LiDAR-based HPE and HMR is the domain gap between datasets captured with different LiDAR sensor characteristics. For example, models trained on the WOD~\cite{sunScalabilityPerceptionAutonomous2020} dataset often fail to generalize to datasets like SLOPER4D~\cite{daiSLOPER4DSceneAwareDataset2023} due to variations in sensor characteristics such as point density, range, and noise patterns. A promising research direction is to develop domain adaptation techniques that enable models to generalize across such sensor discrepancies without requiring extensive re-annotation.

\paragraph{Bridging the Domain Gap Between LiDAR Scanning Patterns}
LiDAR sensors employ different scanning patterns, with NRS (e.g. Human-M3~\cite{fanHumanM3MultiviewMultimodal2023}) and RMB (e.g. WOD~\cite{sunScalabilityPerceptionAutonomous2020} and SLOPER4D~\cite{daiSLOPER4DSceneAwareDataset2023}) being the most prevalent. These patterns produce structurally different point cloud distributions. Developing robust architectures or adaptation strategies that can effectively handle both scanning types is important for building generalizable LiDAR-based pose estimation and mesh recovery systems.

\section{Conclusion}\label{sec:conclusion}

In this paper, we presented a comprehensive review of 3D HPE  and HMR methods from in-the-wild LiDAR point clouds. To provide a complete and coherent perspective, we compared existing approaches across multiple dimensions, ranging from sensing modalities to network architectures, and proposed a structured taxonomy of these methods. Following this taxonomy, we discussed each method, highlighting its strengths and limitations.

In addition to an in-depth discussion of 3D HPE and HMR methods, we conducted a quantitative analysis of the three most widely used datasets, highlighting their characteristics. We also compiled and presented unified definitions of the metrics used to evaluate 3D HPE/HMR performance in LiDAR point clouds. Another key contribution of our work was the creation of benchmark tables for both 3D HPE and HMR on these same three datasets, which we believe will facilitate fair comparisons and advance the field.

Finally, we identified and elaborated on several open challenges that are critical to improving performance in 3D human understanding based on LiDAR data.

As a review paper, our work aims to help the research community understand the current state of the field, identify its gaps, and guide the development of better 3D HPE and HMR solutions in LiDAR point clouds.

\clearpage

\newpage
\bibliographystyle{unsrt}
\bibliography{refs}
\end{document}